\PassOptionsToPackage{textwidth=25mm,textsize=tiny}{todonotes}

\documentclass{article}

\usepackage{iclr2023_conference,times}

\usepackage{amsmath,amsfonts,bm}

\newcommand{\NN}{{\ensuremath \operatorname{NN}}}

\def\figref#1{Fig.~\ref{#1}}

\def\secref#1{\S\ref{#1}}

\def\eqref#1{equation~\ref{#1}}

\def\1{\bm{1}}

\DeclareMathAlphabet{\mathsfit}{\encodingdefault}{\sfdefault}{m}{sl}
\SetMathAlphabet{\mathsfit}{bold}{\encodingdefault}{\sfdefault}{bx}{n}

\newcommand{\reals}{\ensuremath{\mathbb{R}}}

\usepackage[utf8]{inputenc} %
\usepackage[T1]{fontenc}    %
\usepackage{url}            %
\usepackage{booktabs}       %
\usepackage{amsfonts}       %
\usepackage{nicefrac}       %
\usepackage{microtype}      %
\usepackage[inline]{enumitem}
\usepackage{caption}
\usepackage{subcaption}
\usepackage{float}
\usepackage{amsmath,amsthm}
\usepackage{mathtools}
\usepackage{multirow}
\usepackage[final]{changes}
\usepackage{wrapfig}

\usepackage{hyperref}       %
\usepackage{xcolor}         %

\usepackage{graphicx}
\usepackage{xspace}

\newcommand{\para}[1]{\noindent{\textbf{#1}}}
\renewcommand{\cite}{\citep}

\newcommand{\Mtl}{{Human-compatible representations}\xspace}
\newcommand{\mtl}{{human-compatible representations}\xspace}
\newcommand{\shortmtl}{\texttt{HC}\xspace}
\newcommand{\mtlfiltered}{\texttt{HC-filtered}\xspace}
\newcommand{\shortresn}{\texttt{MLE}\xspace}
\newcommand{\resn}{{MLE representations}\xspace}
\newcommand{\resnmodel}{{ResNet}\xspace}
\newcommand{\tn}{\texttt{TMLModel}\xspace}
\newcommand{\tnfiltered}{\texttt{TML-filtered}\xspace}
\newcommand{\shorttn}{\texttt{TML}\xspace}
\newcommand{\sameclasstriplets}{same-class triplets\xspace}

\newcommand{\labelderivedtriplets}{label-derived triplets\xspace}

\newcommand{\nino}{neutral decision support\xspace}
\newcommand{\NINO}{Neutral decision support\xspace}
\newcommand{\nifo}{persuasive decision support\xspace}
\newcommand{\NIFO}{Persuasive decision support\xspace}

\theoremstyle{plain}

\newcommand{\f}{\ensuremath{f}}
\newcommand{\inputspace}{\ensuremath{\mathcal{X}}}
\newcommand{\outputspace}{\ensuremath{\mathcal{Y}}}

\newcommand{\tabref}[1]{Table~\ref{#1}}

\iclrfinalcopy %
\title{Learning Human-Compatible Representations for Case-Based Decision Support}

\author{Han Liu, Yizhou Tian, Chacha Chen, Shi Feng, Yuxin Chen \& Chenhao Tan \\
Department of Computer Science, University of Chicago \\
\texttt{\{hanliu,tianh,chacha,shif,chenyuxin,chenhao\}@uchicago.edu} \\
}

\begin{document}
\maketitle

\begin{abstract}
    Algorithmic case-based decision support provides examples to aid people in decision making tasks by providing contexts for a test case. Despite the promising performance of supervised learning, representations learned by supervised models may not align well with human intuitions: what models consider similar examples can be perceived as distinct by humans.
    As a result, they have limited effectiveness in case-based decision support.
    In this work, we incorporate ideas from metric learning with supervised learning to examine the importance of alignment for effective decision support.
    In addition to instance-level labels, we use human-provided triplet judgments to learn human-compatible decision-focused representations.
    Using both synthetic data and human subject experiments in multiple classification tasks, we demonstrate that such representation is better aligned with human perception than representation solely optimized for classification.
    Human-compatible representations identify nearest neighbors that are perceived as more similar by humans and allow humans to make more accurate predictions, leading to substantial improvements in human decision accuracies (17.8\% in butterfly vs. moth classification and 13.2\% in pneumonia classification).
\end{abstract}

\section{Introduction}

Despite the impressive performance of machine learning (ML) models, humans are often the final decision maker in high-stake domains due to ethical and legal concerns \citep{lai+tan:19,green2019principles}, 
so ML models as decision support is preferred over full automation.
In order to provide meaningful information to human decision makers, the model
cannot be illiterate in the underlying problem, e.g., a model for assisting breast cancer radiologists should have a 
high diagnostic accuracy by itself.
However, a model with high \textit{autonomous} performance
may not provide the most effective decision support,
because it could solve the problem in a way that is not comprehensible or even perceptible to humans, e.g., AlphaGo's famous move 37~\citep{silver2016mastering,silver2017mastering,metz2016two}.
Our work studies the relation between these two objectives that effective decision support must balance: achieving high autonomous performance and aligning with human intuitions.

We focus on case-based decision support for classification problems~\citep{kolodneer1991improving,begum2009case,liao2000case,lai+tan:19}.
For each test example, in addition to showing the model's predicted label, case-based decision support shows one or more related examples retrieved from the training set.
These examples can be used to justify the model's prediction, e.g., by showing similar-looking examples with the predicted label, or to help human decision makers calibrate its uncertainty, e.g., by showing similar-looking examples from other classes.
Both use cases require the model to know what 
is similiar-looking to the human decision maker.
In other words, an important consideration in aligning with human intuition is approximating human judgment of similarity.

\begin{figure}[t]
\centering
\includegraphics[width=.85\linewidth]{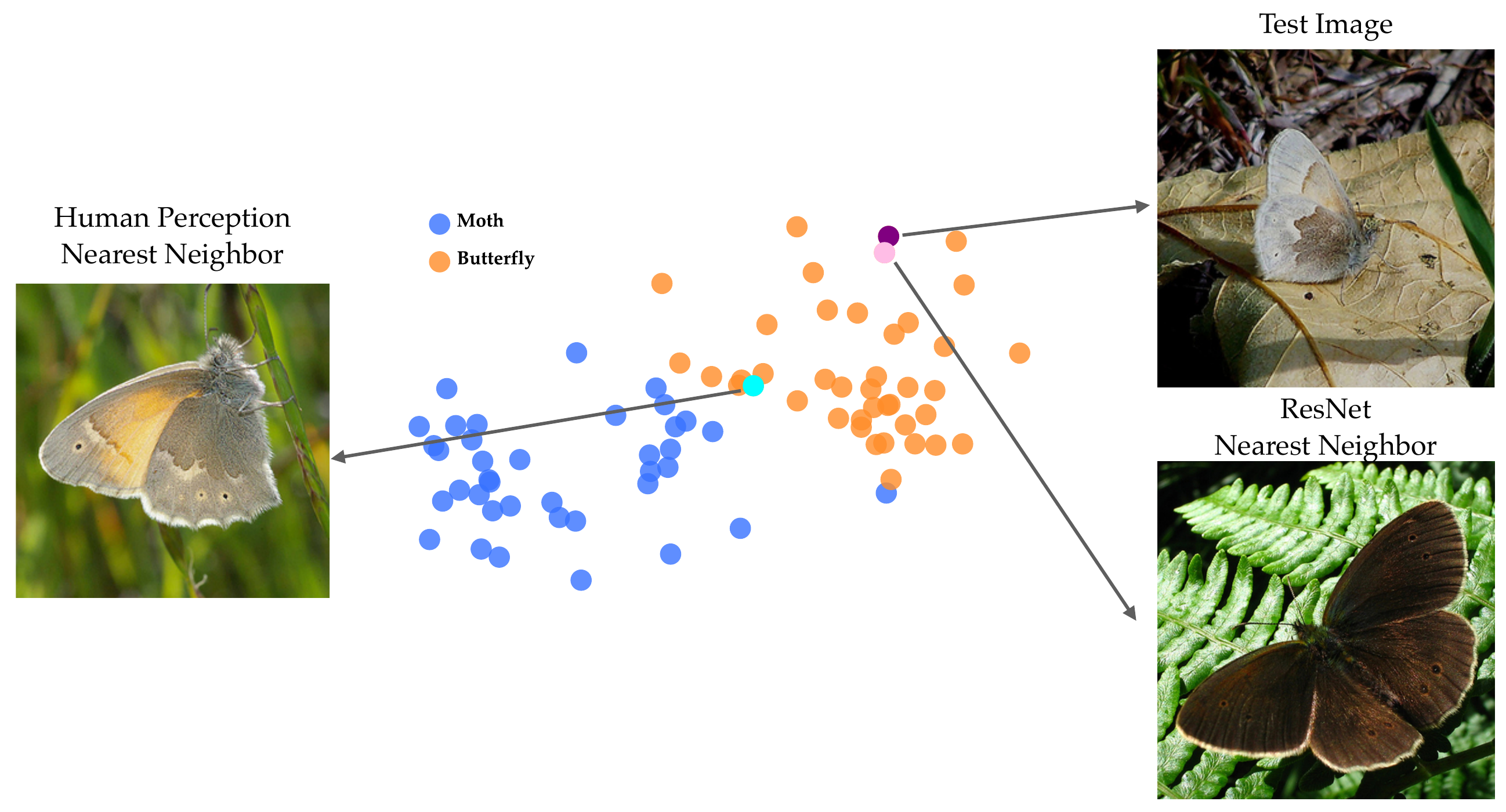}
\caption{Nearest neighbor retrieved by the model representation might not align with human similarity judgment. The \resn (512-dim) are visualized using t-SNE \citep{van2008visualizing}. The purple circle represents a specific
test instance. The nearest neighbor found by \resn (pink circle) is not as visually similar
as the instance in cyan circle found by optimizing a metric learning objective.
}
\label{fig:model_vs_human}
\end{figure}

Figure~\ref{fig:model_vs_human} illustrates the importance of such alignment on a classification problem of distinguishing 
butterfly from moth.
A high-accuracy \resnmodel~\citep{he2016deep} produces a highly linearly-separable representation space, which leads to high classification accuracy.
But the nearest neighbor cannot provide effective justification for model prediction because it looks 
\added{dissimilar to} the test example for humans.
The similarity measured in model representation space does not align with human visual similarity.
If we instead use representations from a second model trained specifically to mimic human visual similarity rather than to 
\added{classify images}, the nearest neighbor would provide strong justification for the model prediction.
However, using the second model for decision support has the risk of misleading or even deceiving the human decision maker because the ``justification'' is generated based on a representation space that is different from the model used to predict the label;
it becomes persuasion rather than justification.

The goal of this work is to learn a \emph{single} representation space that satisfies two properties:
\begin{enumerate*}[label=(\roman*)]
  \item producing easily separable representations for different classes to support accurate classification, and
  \item constituting a metric space that is aligned with human perception of similarity between examples
\end{enumerate*}.
Simultaneously matching the best model on classification accuracy and achieving perfect approximation of human similarity might not be possible, but we hypothesize that a good trade-off between the two would benefit decision support.
We propose a novel multi-task learning
method that combines supervised learning and metric learning.
We supplement the standard maximum likelihood objective with a \added{triplet margin} loss function from \citet{balntas2016learning}.
  Our method learns from human annotations of similarity judgments among data instances in the triplet form.

We validate our approach with both synthetic data and user study.
We show that representations learned from our framework identify nearest neighbors that are perceived as more similar by the synthetic human than that based on supervised classification (henceforth {\em MLE representations}, see \secref{sec:formulation} for details), and are therefore more suitable to provide decision support. 
We further demonstrate that the advantage of \mtl indeed derives from human perception rather than data augmentation.

We further conduct human subject experiments using two classification tasks: \begin{enumerate*}[label=(\roman*)]
  \item butterfly vs. moth classification from ImageNet \citep{krizhevsky2012imagenet}, and
  \item pneumonia classification based on chest X-rays \citep{kermany2018chestxray}.
\end{enumerate*}
\added{Our results show} that \mtl provide more effective decision support than \resn.
In particular, \mtl allow laypeople to achieve an accuracy of 79.1\% in pneumonia classification, 15.3\% higher than \resn. A similar improvement has been observed on the butterfly vs. moth classification task (34.8\% over \resn and 17.8\% over random).

To summarize, our main contributions include:
\begin{itemize}[leftmargin=*,itemsep=-2pt, topsep=-4pt]
  \item We highlight the importance of alignment in learning human-compatible representations for case-based decision support.
  \item We propose a multi-task learning framework that combines supervised learning and metric learning to simultaneously learn classification and human visual similarity.
  \item We design a novel evaluation framework for comparing representations in decision support.
  \item Empirical results with synthetic data and human subject experiments demonstrate the effectiveness of our approach.
\end{itemize}

\section{Case-Based Decision Support}
\label{sec:formulation}

Consider the problem of using a classification model $h: \inputspace \rightarrow \outputspace$ as decision support for humans.
Simply showing the predicted label from the model provides limited information.
Explanations are commonly hypothesized to improve human performance by providing additional information \citep{doshi2017towards}.
We focus on information presented in the form of examples from the training data, also known as case-based decision support \cite{kolodneer1991improving,begum2009case,liao2000case,lai+tan:19}.
Case-based decision support can have diverse use cases and goals.
Given a test example ($x$) and its predicted label ($\hat{y}$), two common use cases are:
\begin{itemize}[topsep=-2pt, leftmargin=*, itemsep=-4pt]
  \item Presenting the nearest neighbor of $x$ along with label $\hat{y}$ as a justification of the predicted label. We refer to this scenario as {\em justification} \cite{kolodneer1991improving}.
  \item Presenting the nearest neighbor in each class without presenting $\hat{y}$. This approach makes a best-effort attempt to provide evidence and leaves the final decision to humans, without biasing humans with the predicted label. We refer to this scenario as {\em \nino} \cite{lai+tan:19}.
\end{itemize}

\paragraph{Formulation.}
Building on \citet{kolodneer1991improving}, we formalize the problem of case-based decision support in the context of representation learning. 
The goal is to assist humans on a classification problem with groundtruth $\f: \inputspace \rightarrow \outputspace$.
We assume access to a representation model $g$, which takes an input $x \in \inputspace$ and generates an $m$-dimensional representation $g(x) \in \reals^m$.
For each test instance $x$, an example selection policy $\pi$ chooses $k$ labeled examples from the training set $D^\text{train}$ and shows them to the human (optionally along with the labels); the human then makes a prediction by choosing a label from $\outputspace$.
As discussed in the two common use cases, %
we consider nearest-neighbor-based selection policies in this work.
The focus of this work is thus on the effectiveness of $g$ for case-based decision support.

Given a neural classification model $h: \inputspace \rightarrow \outputspace$, the representation model is the last layer before the classification head, which is a byproduct derived from $h$.
We refer to this model as $e(h)$.\footnote{In general, we can use the representation in any layer, but in preliminary experiments, we find representation from the last layer is most effective.}
In justification, the example selection policy is $\pi=\NN(x, e(h), D^\text{train}_{\hat{y}})$, 
where $\hat{y}=h(x)$, $D^\text{train}_{\hat{y}}$ 
refers to the subset of training data with label $\hat{y}$ (i.e., $\{(x, y) \in D^\text{train}\ |\ y=\hat{y}\}$), and $\NN$ finds the nearest neighbor of $x$ using representations from $e(h)$ among the subset of examples with label $\hat{y}$.
In decision support, the example selection policy is $\{\NN(x, e(h), D^\text{train}_{y}),\ \forall y \in \outputspace \}$.

\paragraph{Misalignment with human similarity metric is detrimental.}
We argue that aligning model representations with human similarity metric is crucial for case-based decision support; we refer to it as the \textit{metric alignment problem}.
To illustrate the importance of alignment, we need to reason about the goal of case-based decision support.
Let us start with justification, which is a relatively easy case. 
To justify a predicted label, the chosen example should ideally {\em appear similar} to the test image.
Crucially, this similarity is perceived by humans (i.e., interpretable), and the example selection policy identifies the nearest neighbor based on model representation (i.e., faithful).
The gap between human representation and model representation (\figref{fig:model_vs_human}) leads to undesirable justification.

\NINO, however, represents a more complicated scenario.
We start by emphasizing that the goal is not simply to maximize human decision accuracy, because one may use policies that intentionally show distant examples to nudge or manipulate humans towards making a particular decision.\footnote{We will consider one such policy for the sake of evaluating the quality of representations in \secref{sec:exp}.}
Choosing the nearest neighbors in each class is thus an attempt to present \emph{faithful} and \emph{neutral} evidence from the representation space so that humans can make their own decisions, hence preserving their agency.
Therefore, the chosen nearest neighbors should be visually similar to the test instance by human perception, again highlighting the potential gap between model representation and human representation.
Assuming that humans follow the natural strategy by picking the presented instance that's most {\em similar} to the test instance and answering with the corresponding label, then ideally, nearest neighbors in each class retain key information useful for classification so that they can reveal the separation learned in the model.

It is unlikely that we get high alignment by solely optimizing classification even when the model's classification accuracy is comparable to the human's.
Models trained with supervised learning almost always exploit patterns in the training data that are
\begin{enumerate*}[label=(\roman*)]
  \item not robust to distribution shifts, and
  \item counterintuitive or even unobservable for humans~\citep{ilyas2019adversarial,xiao2020noise}
\end{enumerate*}.

\paragraph{Combining metric learning on human triplets with supervised classification.}
We propose to address the metric alignment problem with additional supervision on the human similarity metric.
We collect data in the form of human similarity judgment triplets (or \textit{triplets} for short).
Each triplet is an ordered tuple: $(x^r, x^+, x^-)$, which indicates $x^+$ is judged by humans as being closer to the reference $x^r$ than $x^-$~\citep{balntas2016learning}.
Given a triplet dataset $T$ and labeled classification dataset $D$, we learn a model $\theta$ using triplet margin loss~\citep{balntas2016learning} in conjunction with cross-entropy loss, controlled by a hyperparameter $\lambda$:
\begin{equation}
  \small
\lambda\underbrace{\left[- \sum_{(x,y)\sim D} \log\left(p_\theta(y|x)\right)\right]}_\textrm{Cross-entropy loss} + (1-\lambda)\underbrace{\left[\sum_{(x^r,x^+,x^-)\sim T} \max\left(d_\theta(x^r,x^+)-d_\theta(x^r,x^-)+1,0\right)\right]}_\textrm{Triplet margin loss}
\label{eq:mtl_loss}
\end{equation}
where $d_\theta(\cdot,\cdot)$ is the similarity metric based on model representations; we use Euclidean distance. 
In this work, we initialize $\theta$ with a pretrained \resnmodel~\citep{he2016deep}.
When $\lambda=1$ and the triplet margin loss is turned off, the model reduces to a finetuned \resnmodel.
When $\lambda=0$ and the cross-entropy loss is turned off, the model reduces to the triplet based-learning model of \citet{balntas2016learning}; we call it \tn and will use it to simulate humans in some synthetic experiments in the appendix.
Our work is concerned with the representations learned by these models.
Our approach uses the representations learned with $\lambda=0.5$ (henceforth {\em \mtl} and \shortmtl for short).
We refer to the representations fine-tuning \resnmodel with the cross-entropy loss as {\em \resn} (\shortresn for short) and the representations from \tn as \shorttn.

\section{Experimental Setup}
\label{sec:exp}

In this section, we provide the specific model instantiation and detailed experiment setup.

\para{Models.}
All models and baselines use \resnmodel-18~\citep{he2016deep} pretrained on ImageNet as the backbone image encoder. 
Following \citet{chen2020simple}, we take the output of the average pooling layer and feed it into an MLP projection head with desired embedding dimension. We use the output of the projection head as our final embeddings (i.e., representations), where we add task-specific head and loss for training and evaluation.
We use Euclidean distance as the similarity metric for both loss calculation and distance measurement during example selection in decision support.

Our first baseline uses representations from \resnmodel finetuned with classification labels using cross-entropy loss (i.e., \shortresn).
\resnmodel typically achieves high classification accuracy but does not necessarily produce human-aligned representations.
Our second baseline uses representations from the same pretrained model finetuned with human triplets using triplet margin loss~\citep{balntas2016learning} (i.e., \shorttn).
We expect \shorttn to produce more aligned representations but achieve lower classification accuracy than \shortresn and may provide limited effectiveness in decision support.

Our representations, \shortmtl, are learned by combining the two loss terms following Equation~\ref{eq:mtl_loss}.
The hyperparameter $\lambda$ controls the trade-off between metric alignment and classification accuracy:
with higher $\lambda$ we expect \shortmtl to be more similar to \shortresn, while lower $\lambda$ steers \shortmtl towards \shorttn.
Empirically tuning $\lambda$ confirms this hypothesis.
For the main paper, we present results with
$\lambda=0.5$.
More details about model specification and hyperparameter tuning can be found in the appendix.

\para{Filtering classification-inconsistent triplets.}
  Human triplets may not always align with classification: triplet annotators may choose the candidate from the incorrect class over the one from the correct class.
We refer to these data points as {\em classification-inconsistent triplets}.
We consider a variant of \mtl where we isolate human intuition that's compatible with classification and remove these classification-inconsistent triplets from the training set; we refer to this condition as \mtlfiltered. 
Filtering is yet another way to strike a balance between human intuition and classification.
We leave further details on filtering in the appendix.

\para{Evaluation metrics.}
Our method is designed to align representations with human similarity metrics and at the same time retain the representations' predictive power for classification.
We can evaluate these representations with classification and triplet accuracy using existing data, but 
our main evaluation is designed to simulate case-based decision support scenarios.

\begin{itemize}[itemsep=-2pt, topsep=-4pt, leftmargin=*]
  \item \textbf{Head-to-head comparisons} (``\textbf{H2H}''). 
  To evaluate justification, we set up head-to-head comparisons between two representations ($R_1$ vs. $R_2$) and ask: given a test instance and two justifications retrieved by $R_1$ and $R_2$,
  which justification do
   humans consider as closer to the test instance? We report the fraction of rounds that $R_1$ is preferable.
  In addition to the typical justification for the predicted label, we also examine that for 
    classes other than the predicted class,
    as those examples will be used in decision support 
    for users to examine the plausibility of each class.
  We refer to the nearest example 
    in the \emph{predicted} class
   as {\em NI}, and the nearest example 
  in the other class
    as {\em NO}.

  \item \textbf{\NINO}. Following \secref{sec:formulation}, we retrieve the nearest neighbors from each class. We use the accuracy of 
   humans as the measure of effective decision support.

  \item \textbf{\NIFO}. We retrieve the  nearest example with the predicted label
   and the furthest example from the other class.
  If the representation is aligned with human similarity metric, this approach encourages people to follow the predicted label, which likely leads to over-reliance and may be unethical in practice.
  Here, we use this scenario as a surrogate to evaluate the quality of the learned representations.

\end{itemize}

  Note that we do not show model predictions so that humans focus on the similarity between examples.

\section{Synthetic Experiment}
\label{sec:synthetic}

\begin{figure}[t]
  \centering
  \includegraphics[width=\textwidth]{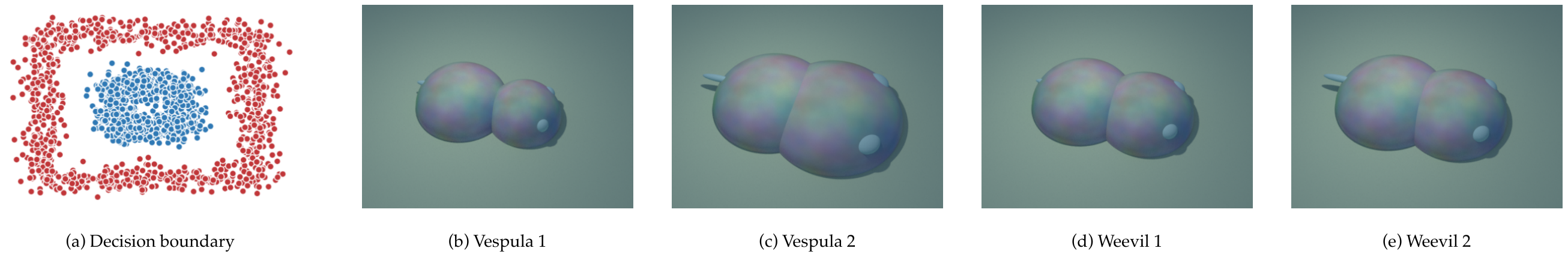}
   \caption{VW dataset. (a) shows the dataset where labels are determined (non-linearly) by two features: the head and the body size of the fictional insects. (b)-(d) show samples of the two classes; the Weevil has a mid-sized body and mid-sized head, while the Vespula does not. Tail length and texture are two non-informative features.}
   \label{fig:vw}
  \end{figure}

To understand the strengths and limitations of our method, we first experiment with synthetic datasets.
Using simulated human similarity metrics, we control and vary the level of disagreement between the classification groundtruth and the synthetic human's knowledge.

\subsection{Synthetic dataset and simulated human similarity metrics}

We use the synthetic dataset ``Vespula vs Weevil'' (VW) from \citet{chen2018near}.
It is a binary image classification dataset of two fictional species of insects.
Each example contains four features, two of them---head and body size---are predictive of the label, and the other two---tail length and texture---are completely non-predictive.
We generate 2000 images and randomly split the dataset into training, validation, and testing sets in a 60\%:20\%:20\% ratio.
The labels are determined by various synthetic decision boundaries, such as the one shown in \figref{fig:vw}a. 

  To generate triplets data, we define simulated human similarity metrics as a weighted Euclidean distance over the visual features: for any instance $a$ and $b$, $d(a,b)=\sqrt{\sum_{i} w_i(a_i-b_i)^2}$, where $i$ refers to the $i$-th feature.
By changing the weight of each feature, we can control the level of disagreement between a synthetic human and the groundtruth.
All procedures that involve humans (i.e., triplet data collection and evaluation) are done by the synthetic human in this section.

To quantify the disagreement, we use 1-NN classification accuracy following the synthetic human similarity metric; we refer to it as the {\em task alignment score}. Note that this is different from our main alignment problem, which is about the representations.
The task alignment score ranges from 50\% (setting the informative features' weights to 0 and distractor weights to 1) to 100\%.
See the appendix for more details on how we generate these weights.
In each setting, we generate 40,000 triplets.

\subsection{Results}

We compare \shortmtl, \shortresn, \shorttn on classification accuracy, triplet accuracy, and decision support performance for the synthetic human. We train all three representations with a large dimension of 512 and a small dimension of 50 and observe that the 512-dimension representation is preferable based on most metrics. We also train \shortmtl on filtered vs. unfiltered triplets as well as with different values $\lambda$. 
For our main results, we report the performance with $\lambda=0.5$ and filtered triplets for the decision boundary in \figref{fig:vw}a.
We will discuss the effect of filtering later in this section.
$\lambda$'s role is relatively limited and we will discuss its effect and other decision boundaries in the appendix.

In synthetic experiments, \shortmtl achieves the same perfect classification accuracy as \shortresn (100\%), and a triplet accuracy of 96.8\%, which is comparable to \shorttn (97.3\%).
This shows that \shortmtl indeed learns both the classification task and human similarity prediction task.
We next present the evaluation of case-based decision support with the synthetic human, which is the key goal of this work.

\para{\shortmtl significantly outperforms \shortresn in H2H.} 
If there is no difference between \shortmtl and \shortresn, the synthetic human should prefer \shortmtl about 50\% of times.
However, as shown in \tabref{tab:main-results}, our synthetic human prefer \shortmtl over \shortresn by a large margin (about 90\% of times) as justifications for both nearest in-class examples and nearest out-of-class examples, indicating the NIs and NOs selected based on the \shortmtl representations are more aligned with the synthetic human than \shortresn. %
For NI H2H, the preference towards \shortmtl declines as the task alignment improves, because if alignment between human similarity and classification increases, \shortresn can capture human similarity as a byproduct of classification.

\begin{table}[]
  \small
  \centering
  \caption{Experiment results on VW with H2H comparison and decision support evaluations.
  }
  \begin{tabular}{@{}lrrrrrr@{}}
  \toprule
  Task alignment   & 50\%   & 80\%   & 83\%  & 92\%  & 92.5\% & 100\%     \\ \midrule
  Weights  &  [0,0,1,1]  & [1,0,1,1] &  [0,1,1,1] &  [1,256,256,256] &  [256,1,256,256] &  [1,1,1,1]  \\ \midrule

  \multicolumn{7}{c}{\textbf{NI-H2H}} \\ \midrule
  \shortmtl vs. \shortresn     & 0.917 & 0.914 & 0.903 & 0.880 & 0.872 & 0.808 \\ \midrule
  \multicolumn{7}{c}{\textbf{NO-H2H}} \\ \midrule
  \shortmtl vs. \shortresn     & 0.916 & 0.968 & 0.946 & 0.958 & 0.962 & 0.970 \\ \midrule
  \multicolumn{7}{c}{\textbf{\NINO}}       \\ \midrule
  \shortresn     & 0.753 & 0.899 & 0.896 & 0.897 & 0.901 & 0.929 \\
  \shorttn       & 0.568 & 0.775 & 0.807 & 0.868 & 0.877 & \textbf{1.000} \\ 
  \shortmtl      & \textbf{0.759} & \textbf{0.901} & \textbf{0.928} & \textbf{0.949} & \textbf{0.955} & \textbf{1.000} \\ \midrule
  \multicolumn{7}{c}{\textbf{\NIFO}}       \\ \midrule
  \shortresn     & 0.704 & 0.900 & 0.903 & 0.903 & 0.901 & 0.919 \\
  \shorttn       & 0.906 & 0.881 & 0.863 & 0.876 & 0.877 & \textbf{1.000} \\ 
  \shortmtl      & \textbf{1.000} & \textbf{1.000} & \textbf{1.000} & \textbf{1.000} & \textbf{1.000} & \textbf{1.000} \\ \midrule
  \end{tabular}
  \label{tab:main-results}
  \end{table}

\begin{wrapfigure}{r}{0.35\textwidth}
\vspace{-0.2in}
  \centering
    \includegraphics[width=0.32\textwidth]{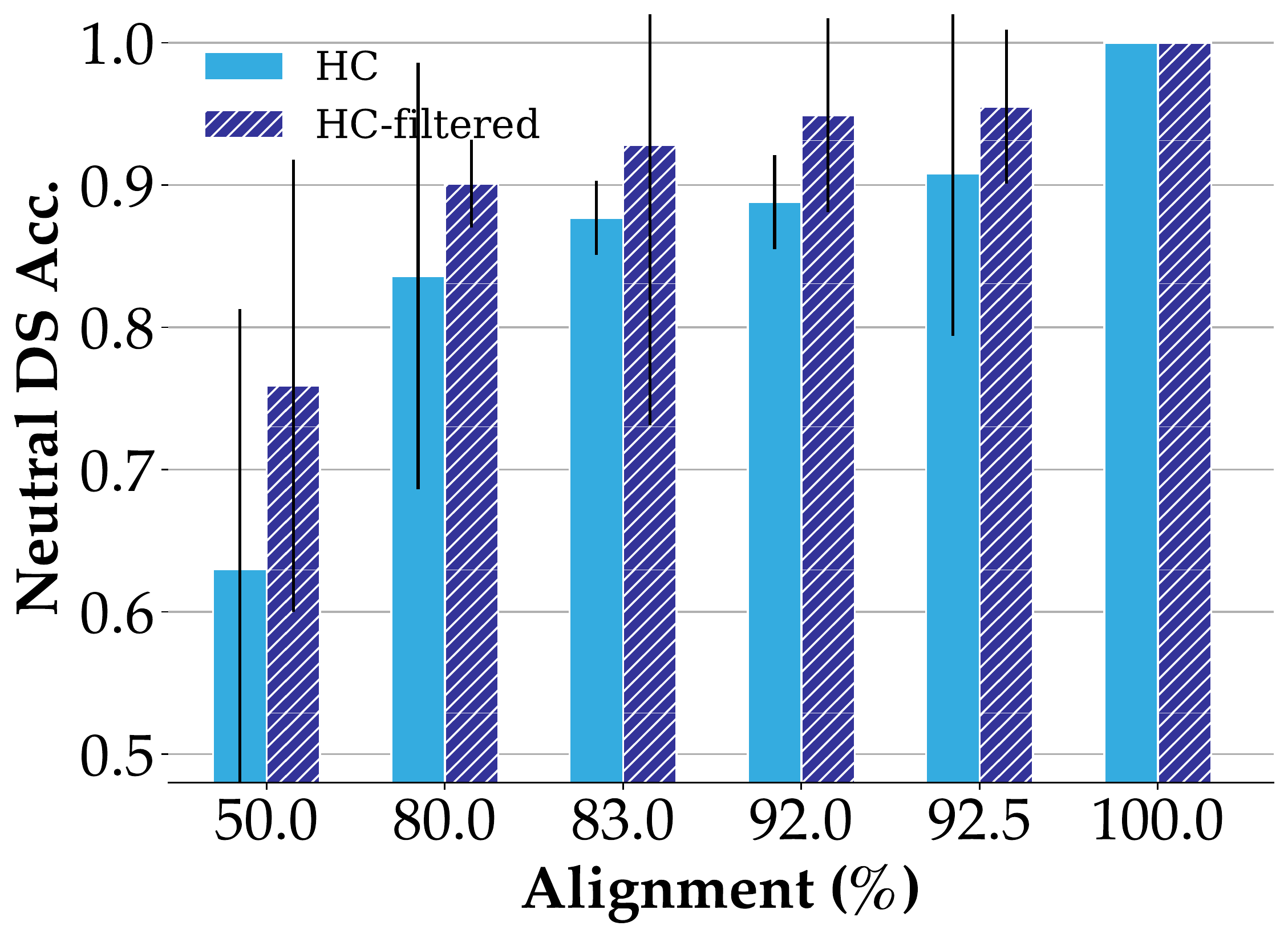}
    \caption{\NINO with \shortmtl and \mtlfiltered. \mtlfiltered leads to improved performance.}
    \label{fig:filter}
\end{wrapfigure}

\para{\shortmtl provides the best decision support.} 
Table \ref{tab:main-results} shows that \shortmtl achieves the highest neutral and persuasive decision support accuracies in all task alignments. %
In \nino, \shortresn consistently outperforms \shorttn, highlighting that representation solely learned for metric learning is ineffective for decision support.
For all models, the decision support performance improves as the task alignment increases, suggesting that decision support is easier when human similarity judgment is aligned with the classification task.
\shortresn and \shorttn are more comparable in \nifo, while \shortmtl consistently achieves 100\%.
The fact that \shortresn shows comparable performance between neutral and \nifo further confirms that \shortresn does not capture human similarity for examples from different classes.

\para{Filtering triplets leads to better decision support}. \figref{fig:filter} shows that filtering class-inconsistent triplets improves \shortmtl's decision support performance across all alignments. Further details in the appendix show that filtering slightly hurts H2H performance. 
This suggests that in terms of decision support, the benefit of filtering out human noise may outweigh the loss of some similarity judgment.

\para{The importance of human perception.} 
One may question whether filtering class-inconsistent triplets essentially provides additional label supervision in the form of triplets.
We show this is not the case by experimenting with \shortmtl trained on \textit{\labelderivedtriplets}. 
Assuming that an instance is more similar to another instance with the same label than one with a different label, we derive \labelderivedtriplets  %
directly from groundtruth labels ($x^+$ from the same class as $x^r$ and $x^-$ from the other class), containing no human perception information.
Table~\ref{tab:wv_triplet_type} shows decision support results for this setting: 
\shortmtl \labelderivedtriplets show worse performance than \mtlfiltered. In fact, \shortmtl \labelderivedtriplets show even worse neutral and \nifo than \shortresn, which may be due to \labelderivedtriplets causing overfitting. This suggests that triplets without human perception do not lead to \mtl.

We also experiment with \shortmtl trained on \sameclasstriplets, human-triplets but only those where the non-reference cases ($x^+, x^-$) are from the same class; that is, the triplets cannot provide any label supervision. We observe from Table~\ref{tab:wv_triplet_type} that \shortmtl trained on these triplets show similar results to \mtlfiltered across all decision support evaluations. This suggests that human perception is the main factor in driving \mtl' high decision support performance.

\begin{table}[]
    \centering
    \caption{Experiment results on VW using synthetic human with 92\% alignment. Comparing \resn and \mtlfiltered with \shortmtl trained on \labelderivedtriplets and \shortmtl trained on \sameclasstriplets.
    40,000 new triplets were generated for each condition. }
    \begin{tabular}{@{}lrrrr@{}}
    \toprule
    Evaluations           & \shortresn & \shortmtl \labelderivedtriplets & \shortmtl \sameclasstriplets  & \mtlfiltered  \\ \midrule
    {NI-H2H with \shortresn}    & N/A    & 0.509           & 0.890   & 0.889                           \\
    {NO-H2H with \shortresn}    & N/A    & 0.607            & 0.970  & 0.958                           \\ 
    {Neutral DS} & 0.897  & 0.723            & 0.960   & 0.949                         \\
    {Persuasive DS} & 0.903  & 0.803            & 0.998    & 1.000                        \\ \bottomrule
    \end{tabular}
    \label{tab:wv_triplet_type}
\end{table}

\section{Human Subject Experiments}

We conduct human subject experiments on two image classification datasets: a natural image dataset, Butterflies v.s. Moths (BM) and a medical image dataset of chest X-rays (CXR). For BM, we followed \citet{singla2014near} and acquired 200 images from ImageNet \cite{krizhevsky2012imagenet}. BM is a binary classification problem and each class contains two species. CXR is a balanced binary classification subset taken from \citet{kermany2018chestxray} with 3,166 chest X-ray images that are labeled with either normal or pneumonia. We randomly split the datasets following 60\%:20\%:20\% ratio. 
The classification accuracy with our base supervised learning models are 97.5\% for BM and 97.3\% for CXR.
We only present results with human subjects in the main paper, but results from simulation experiments with \shorttn as a synthetic agent, such as filtering triplets providing better results, are qualitatively consistent. 
See \secref{sec:supp_human_bm} and \secref{sec:supp_human_cxr} in the appendix for more details.

\subsection{Triplet annotation}

We recruit crowdworkers on Prolific to acquire visual similarity triplets. In each question,
we show a reference image on top and two candidate images below, and ask a 2-Alternative-Forced-Choice (2AFC) question: which candidate image looks more similar to the reference image? A screenshot of the interface can be found in the appendix.
To generate triplets for annotation, we first sample the reference image from either the training, the validation, or the test set. Then for each reference image, we sample two candidates from the training set. We sample the candidates only from the training set because in decision support, the selected examples should always come from the training set, and thus we only need to validate and test triplet accuracies with candidates from the training set.

For BM we recruit 80 crowdworkers, each completing 50 questions, giving us 4000 triplets. For CXR we recruit 100 crowdworkers, each answering 20 questions, yielding 2000 triplets. Our pilot study suggests that visual similarity judgment on chest X-rays is a more mentally demanding task, so we decrease the number of questions for each CXR survey.

\subsection{Results on Butterflies v.s. Moths}

We recruit crowdworkers on Prolific to evaluate representations produced by our models by doing decision support tasks.
We acquire examples with different example selection policies from \shortmtl and \shortresn. We choose the dimension and training triplets of the representation based on the models' classification accuracy, triplet accuracy, and decision support simulation results based on synthetic agents. See more details in the appendix.
We do not include \shorttn in human studies, because in practice, \shorttn models cannot make predictions on class labels, therefore are unable to distinguish and select in-class and out-of-class examples and thus cannot be used for decision support.

\para{H2H comparison results show \shortmtl NI examples are slightly but significantly preferred over \shortresn NI examples according to human visual similarity.}
We recruit 30 Prolific workers to make H2H comparisons between \shortmtl NI examples and \shortresn NI examples over the entire test set. The mean preference for \shortmtl over \shortresn is 0.5316 with a 95\% confidence interval of $\pm0.0302$ ($p=0.0413$ with one-sample t-test). 
This means the \shortmtl NI examples are closer to the test images than \shortresn NI examples with statistical significance according to human visual similarity. 

\begin{figure}[t]
    \centering
    \begin{subfigure}[b]{0.4\textwidth}
        \includegraphics[width=\textwidth]{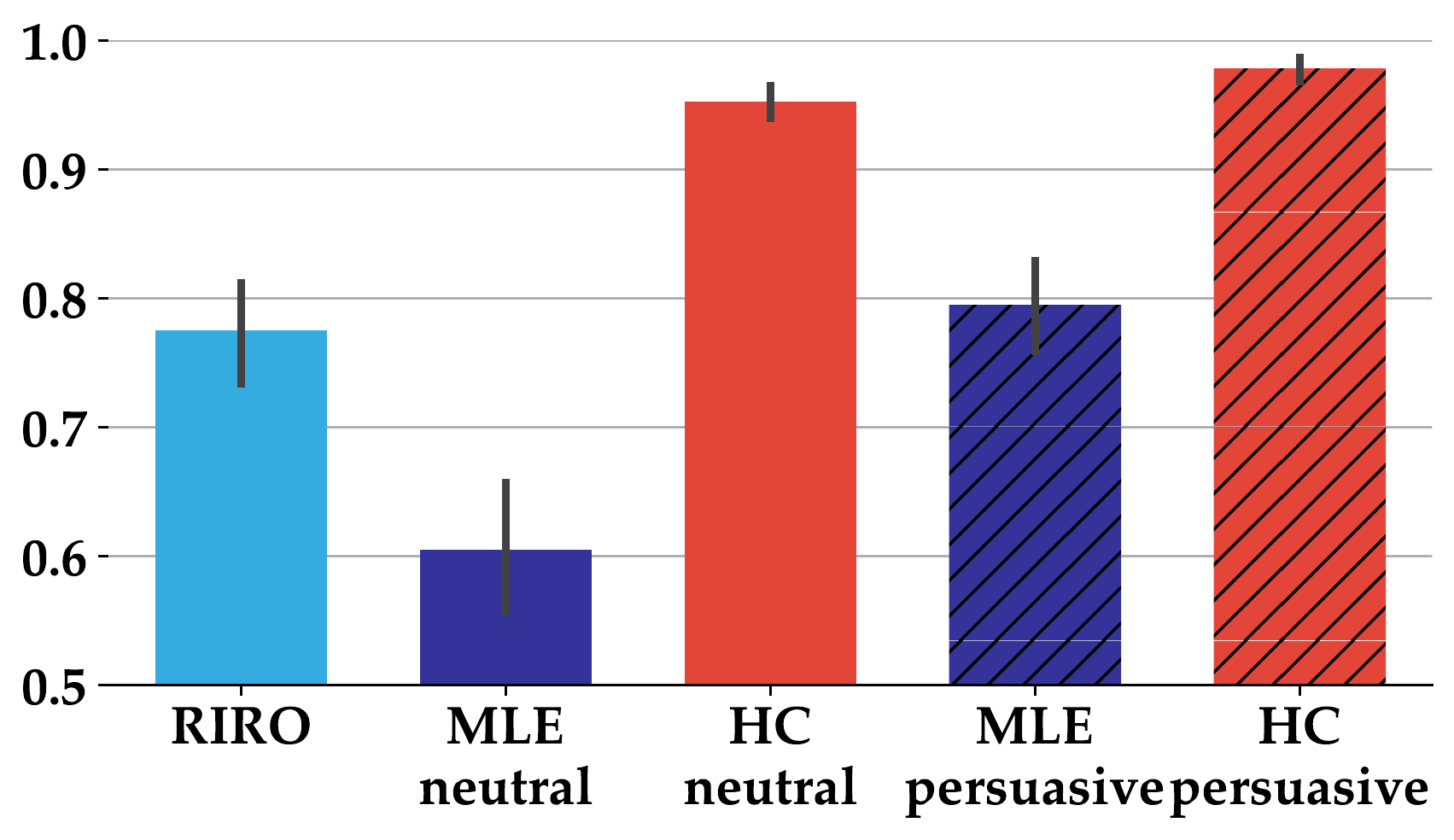}
        \caption{Butterfly vs. Moth.}
        \label{fig:bm_human_decision}
    \end{subfigure}
    \begin{subfigure}[b]{0.4\textwidth}
        \includegraphics[width=\textwidth]{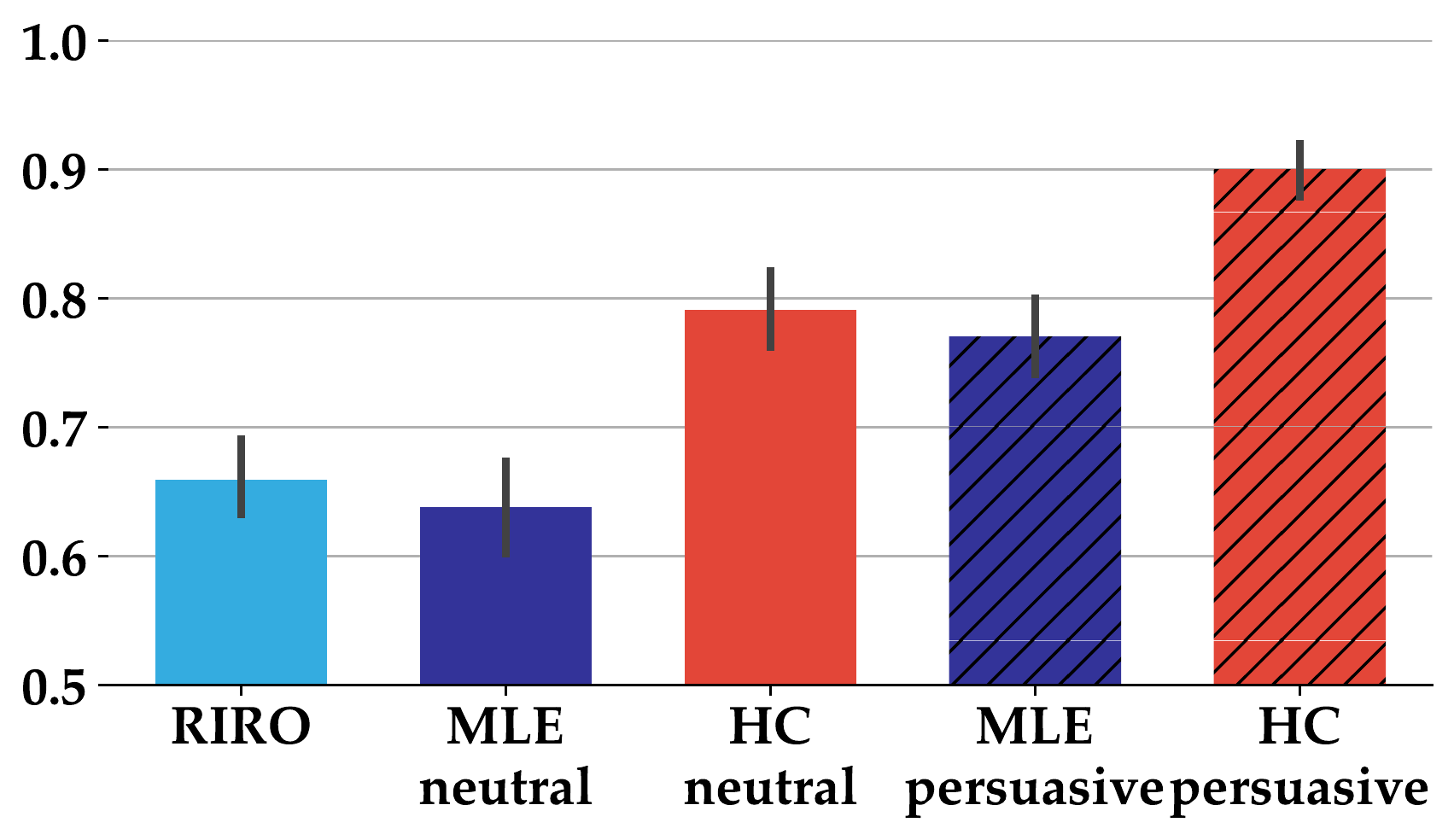}
        \caption{Pneumonia classification.}
        \label{fig:cxr_human_decision}
    \end{subfigure}
    \caption{Decision support accuracy with human subject studies. Error bars show 95\% confidence intervals. \shortmtl dominates \shortresn in both neutral and \nifo. }
\end{figure}

\para{Decision support results show \shortmtl is significantly better than \shortresn both in neutral and \nifo.}
Combining two example selection policies with two representations, we have four conditions: \shortmtl neutral, \shortmtl persuasive, \shortresn neutral, \shortresn persuasive. We also add a baseline condition with random supporting examples, which we call random in-class random out-of-class (RIRO).
We recruit 30 Prolific workers for each condition and ask them to go through %
the images in the test set with supporting examples from each class in the training set. Both the order of the test images and the order of the supporting images within each test question are randomly shuffled.

Figure \ref{fig:bm_human_decision} shows the human classification accuracies with different decision support scenarios and different representations. 
In \nino, we observe that \shortmtl achieves much higher accuracy than \shortresn (95.3\% vs. 60.5\%, $p=4\mathrm{e}{-19}$ with two-sample t-test).
In fact, even RIRO provides better decision support than \resn, suggesting that the supporting images based on \shortresn are confusing and hurt human decision making (77.5\% vs. 60.5\%, $p=3\mathrm{e}{-6}$).
As expected, the accuracies are generally higher in \nifo.
\shortmtl enables an accuracy of 97.8\%, which is much better than \shortresn at 79.5\% ($p=2\mathrm{e}{-13}$).
\shortmtl in \nino already outperforms \shortresn in \nifo.
These findings confirm our results with VW synthetic experiments that \mtl provide much better decision support than \resn.

\subsection{Results on Chest X-rays}
We use the same experimental setup as BM to evaluate \shortmtl and \shortresn representations in CXR.

\para{H2H comparison results show \shortmtl NI examples are slightly preferred over \shortresn NI examples but the difference is not statistically significant.}
We recruit 50 Prolific workers to each make 20 H2H comparisons between \shortmtl NI examples and \shortresn NI examples. The mean preference for \shortmtl over \shortresn is 0.516 with a 95\% confidence interval of $\pm0.0725$ ($p=0.379$ with one-sample t-test). H2H comparison in CXR is especially challenging as laypeople need to differentiate between two chest X-rays in the same class,
hence the slightly worse performance in H2H compared to BM.

\para{Similar to BM, \shortmtl outperforms \shortresn in both neutral and \nifo in CXR.}
As expected,
\figref{fig:cxr_human_decision} shows that pneumonia classification is a much harder task than butterfly vs. moth classification, indicated by the lower accuracies across all conditions. 
In \nino, \shortmtl enables much better accuracy than \shortresn (79.1\% vs. 63.8\%, $p=2\mathrm{e}{-8}$ with two-sample t-test).
In fact, similar to the BM setting, \shortresn provides similar performance with RIRO (63.8\% vs. 65.9\%, $p=0.390$), suggesting that \resn are no different from random representations for selecting nearest neighbors within a class.
To contextualize our results, we would like to highlight that our crowdworkers are laypeople and have no medical training.
It is thus impressive that \mtl enable an accuracy of almost 80\% in \nino, which demonstrates the potential of \mtl.

In \nifo, \shortmtl provides the highest decision support accuracy at 90.0\%, also much higher than \shortresn at 77.0\% ($p=2\mathrm{e}{-10}$). 
Again, while we do not recommend \nifo as a policy for decision support in practice, these results show that our \mtl are indeed more compatible with humans than \resn.

\section{Related Work}

\para{Ordinal embedding.}
The ordinal embedding problem \citep{ghosh2019landmark,van2012stochastic,kleindessner2017kernel,kleindessner2014uniqueness,terada2014local,park2015preference} seeks to find low-dimensional representations that respect ordinal feedback. Currently, there exist several techniques for learning ordinal embeddings. Generalized Non-metric Multidimensional Scaling 
\citep{agarwal2007generalized} takes a max-margin approach by minimizing hinge loss. Stochastic Triplet Embedding 
\citep{van2012stochastic} assumes the Bradley-Terry-Luce 
noise model \citep{bradley1952rank,luce1959individual} and minimizes logistic loss. The Crowd Kernel \citep{tamuz2011adaptively} and t-STE \citep{van2012stochastic} propose alternative non-convex loss measures based on probabilistic generative models. These results are primarily empirical and focus on minimizing prediction error on unobserved triplets. In principle, one can plugin these approaches in our framework
as alternatives to the triplet margin loss in Eq.~\ref{eq:mtl_loss}.

\para{AI explanations and AI-assisted decision making.} 
Various explanation methods have been developed to explain black-box AI models \citep{guidotti2018survey}, such as feature importance \citep{ribeiro2016should, shrikumar2017learning}, saliency map \citep{zhou2016learning, selvaraju2017grad}, and decision rules \citep{ribeiro2018anchors}. 
Example-based explanations are also a type of common explanation methods that use examples to explain AI models. 
Nearest-neighbor examples can explain a model's local decision \citep{wang2021explanations, nguyen2021feature, nguyen2022explanations, lai+tan:19}.
To the best of our knowledge, there has been no prior work that examines the role of representations in choosing the nearest neighbors in the context of AI explanations.
Meanwhile, global example-based explanations such as prototypes can explain a model's global behavior or a model's understanding of the data distribution \citep{kim2016examples, chen2018protopnet, cai2019effects,lai+liu+tan:20}. 
Explaining a model's global behavior is also closely related to machine teaching~\citep{zhu2018overview}.

    Many of these explanation methods have been used in AI-assisted decision making to explain AI predictions or inform users about the AI model or training data \citep{lai2021towards}.
Among them, example-based explanations have shown be useful in many high-stake domains where full AI automation is often not desired, such as recidivism prediction \citep{hayashi2017recidivism} and medical diagnosis \citep{cai2019medical, rajpurkar2020medical, tschandl2020medical}. 
While many of the current literature in AI-assisted decision making focus on generating explanations of AI without considering human feedback, our decision support methods offer assistance by learning from human perceptions and provide examples from human-compatible representations.

\section{Conclusion}
\label{sec:conclusion}

Our work formulates the novel problem of learning human-compatible representations for case-based decision support.
As we identify in this paper, the key to providing effective case-based support with a model is the alignment between the model and the human in terms of similarity metrics: two examples that appear similar to the model should also appear similar to the human.
But models trained to perform classification do not automatically produce representations that satisfy this property.
To address this issue, we propose a multi-task learning method to combine two sources of supervision: labeled examples for classification and triplets of human similarity judgments.
With synthetic experiments and user studies, we validate that \mtl
\begin{enumerate*}[label=(\roman*)]
  \item consistently get the best of both worlds in 
  classification accuracy and triplet accuracy,
  \item select visually more similar examples in head-to-head comparisons,
  \item and provide better decision support%
\end{enumerate*}.

\clearpage

\section*{Acknowledgments}

We thank the anonymous reviewers for their insightful comments.
We also thank members of the Chicago Human+AI lab for their thoughtful feedback.
This work was supported in part by a CDAC discovery grant at the University of Chicago and an NSF grant, IIS-2040989.

\section*{Ethics Statement}

Although coming from a genuine goal to improve human-AI collaboration by aligning AI models with human intuition, our work may have potential negative impacts for the society. We discuss these negative impacts from two perspectives: the multi-task learning framework and the decision support policies.

\subsection*{Multi-task learning framework}
Our \mtl models are trained with two sources of data. The first source of data is classification annotations where groundtruth maybe be derived from scientific evidence or crowdsourcing with objective rules or guidelines. The second source of data is human judgment annotations where groundtruth is probably always acquired from crowdworkers with subjective perceptions. When our data is determined with subjective perceptions, the model that learns from it may inevitably develop bias based on the sampled population. If not carefully designed, the human judgment dataset may contain bias against certain minority group depending on the domain and the task of the dataset. For example, similarity judgment based on chest X-rays of patients in one gender group or racial group may affect the generalizability of the representations learned from it, and may lead to fairness problems in downstream tasks. It is important for researchers to audit the data collection process and make efforts to avoid such potential problems.

\subsection*{Decision support policies}
Among a wide variety of example selection policies, our policies to choose the decision support examples are only attempts at leveraging AI model representations to increase human performance. 
We believe that they are reasonable strategies for evaluating representations learned by a model, but future work is required to establish their use in practice.

The \nino policy aims to select the nearest examples in each class, therefore limiting the decision problem to a small region around the test example. We hope this policy allow human users to zoom in the local neighborhood and scrutinize the difference between the relatively close examples. In other words, \nino help human users develop a local decision boundary with the smallest possible margin. This could be useful for confusing test cases that usually require careful examinations. However, the \nino policy adopts an intervention to present a small region in the dataset and may downplay the importance of global distribution in human users' decision making process. 

The \nifo policy aims to select the nearest in-class examples but the furthest out-of-class examples. It aims to maximize the visual difference between examples in opposite class, thus require less effort for human users to adopt case-based reasoning for classification. It also helps human users to develop a local decision boundary with the largest possible margin. However, when model prediction is incorrect, the policy end up selecting the furthest in-class examples with the nearest out-of-class examples, completely contrary to what it is design to do, may lead to even over-reliance or even adversarial supports.

In general, decision support policies aim to choose a number of supporting examples without considering some global properties such as representativeness and diversity. While aiming to reduce humans' effort required in task by encouraging them to make decision in a local region, the decision support examples do not serve as a representative view of the whole dataset, and may bias human users to have a distorted impression of the data distribution. It remains an open question that how to ameliorate these negative influence when designing decision support interactions with case-based reasoning.

\section*{Reproducibility Statement}
 
Implementation details and computing resources are documented in \secref{sec:implementation} in the appendix. Hyperparameters and model configuration are reported in both the main paper and the appendix along with each experiments.
Our code and data are available at \url{https://github.com/ChicagoHAI/learning-human-compatible-representations}.

\bibliography{main}

\begin{thebibliography}{54}
\providecommand{\natexlab}[1]{#1}
\providecommand{\url}[1]{\texttt{#1}}
\expandafter\ifx\csname urlstyle\endcsname\relax
  \providecommand{\doi}[1]{doi: #1}\else
  \providecommand{\doi}{doi: \begingroup \urlstyle{rm}\Url}\fi

\bibitem[Agarwal et~al.(2007)Agarwal, Wills, Cayton, Lanckriet, Kriegman, and
  Belongie]{agarwal2007generalized}
Sameer Agarwal, Josh Wills, Lawrence Cayton, Gert Lanckriet, David Kriegman,
  and Serge Belongie.
\newblock Generalized non-metric multidimensional scaling.
\newblock In \emph{Artificial Intelligence and Statistics}, pp.\  11--18, 2007.

\bibitem[Balntas et~al.(2016)Balntas, Riba, Ponsa, and
  Mikolajczyk]{balntas2016learning}
Vassileios Balntas, Edgar Riba, Daniel Ponsa, and Krystian Mikolajczyk.
\newblock Learning local feature descriptors with triplets and shallow
  convolutional neural networks.
\newblock In \emph{Bmvc}, volume~1, pp.\ ~3, 2016.

\bibitem[Begum et~al.(2009)Begum, Ahmed, Funk, Xiong, and
  Von~Sch{\'e}ele]{begum2009case}
Shahina Begum, Mobyen~Uddin Ahmed, Peter Funk, Ning Xiong, and
  Bo~Von~Sch{\'e}ele.
\newblock A case-based decision support system for individual stress diagnosis
  using fuzzy similarity matching.
\newblock \emph{Computational Intelligence}, 25\penalty0 (3):\penalty0
  180--195, 2009.

\bibitem[Bradley \& Terry(1952)Bradley and Terry]{bradley1952rank}
Ralph~Allan Bradley and Milton~E Terry.
\newblock Rank analysis of incomplete block designs: I. the method of paired
  comparisons.
\newblock \emph{Biometrika}, 39\penalty0 (3/4):\penalty0 324--345, 1952.
\newblock URL \url{https://bit.ly/2QsOf4P}.

\bibitem[Cai et~al.(2019{\natexlab{a}})Cai, Jongejan, and
  Holbrook]{cai2019effects}
Carrie~J Cai, Jonas Jongejan, and Jess Holbrook.
\newblock The effects of example-based explanations in a machine learning
  interface.
\newblock In \emph{Proceedings of the 24th international conference on
  intelligent user interfaces}, pp.\  258--262, 2019{\natexlab{a}}.

\bibitem[Cai et~al.(2019{\natexlab{b}})Cai, Winter, Steiner, Wilcox, and
  Terry]{cai2019medical}
Carrie~J. Cai, Samantha Winter, David Steiner, Lauren Wilcox, and Michael
  Terry.
\newblock "hello ai": Uncovering the onboarding needs of medical practitioners
  for human-ai collaborative decision-making.
\newblock \emph{Proc. ACM Hum.-Comput. Interact.}, 3\penalty0 (CSCW), nov
  2019{\natexlab{b}}.
\newblock \doi{10.1145/3359206}.
\newblock URL \url{https://doi.org/10.1145/3359206}.

\bibitem[Chen et~al.(2018{\natexlab{a}})Chen, Li, Barnett, Su, and
  Rudin]{chen2018protopnet}
Chaofan Chen, Oscar Li, Alina Barnett, Jonathan Su, and Cynthia Rudin.
\newblock This looks like that: deep learning for interpretable image
  recognition.
\newblock \emph{CoRR}, abs/1806.10574, 2018{\natexlab{a}}.
\newblock URL \url{http://arxiv.org/abs/1806.10574}.

\bibitem[Chen et~al.(2020)Chen, Kornblith, Norouzi, and Hinton]{chen2020simple}
Ting Chen, Simon Kornblith, Mohammad Norouzi, and Geoffrey Hinton.
\newblock A simple framework for contrastive learning of visual
  representations.
\newblock In \emph{International conference on machine learning}, pp.\
  1597--1607. PMLR, 2020.

\bibitem[Chen et~al.(2018{\natexlab{b}})Chen, Mac~Aodha, Su, Perona, and
  Yue]{chen2018near}
Yuxin Chen, Oisin Mac~Aodha, Shihan Su, Pietro Perona, and Yisong Yue.
\newblock Near-optimal machine teaching via explanatory teaching sets.
\newblock In \emph{International Conference on Artificial Intelligence and
  Statistics}, pp.\  1970--1978. PMLR, 2018{\natexlab{b}}.

\bibitem[Doshi-Velez \& Kim(2017)Doshi-Velez and Kim]{doshi2017towards}
Finale Doshi-Velez and Been Kim.
\newblock Towards a rigorous science of interpretable machine learning.
\newblock \emph{arXiv preprint arXiv:1702.08608}, 2017.

\bibitem[Dosovitskiy et~al.(2021)Dosovitskiy, Beyer, Kolesnikov, Weissenborn,
  Zhai, Unterthiner, Dehghani, Minderer, Heigold, Gelly, Uszkoreit, and
  Houlsby]{dosovitskiy2021an}
Alexey Dosovitskiy, Lucas Beyer, Alexander Kolesnikov, Dirk Weissenborn,
  Xiaohua Zhai, Thomas Unterthiner, Mostafa Dehghani, Matthias Minderer, Georg
  Heigold, Sylvain Gelly, Jakob Uszkoreit, and Neil Houlsby.
\newblock An image is worth 16x16 words: Transformers for image recognition at
  scale.
\newblock In \emph{International Conference on Learning Representations}, 2021.
\newblock URL \url{https://openreview.net/forum?id=YicbFdNTTy}.

\bibitem[{Falcon et al.}(2019)]{pytorch-lightning}
William {Falcon et al.}
\newblock Pytorch lightning, 2019.

\bibitem[Ghosh et~al.(2019)Ghosh, Chen, and Yue]{ghosh2019landmark}
Nikhil Ghosh, Yuxin Chen, and Yisong Yue.
\newblock Landmark ordinal embedding.
\newblock In \emph{Advances in Neural Information Processing Systems}, pp.\
  11502--11511, 2019.

\bibitem[Green \& Chen(2019)Green and Chen]{green2019principles}
Ben Green and Yiling Chen.
\newblock The principles and limits of algorithm-in-the-loop decision making.
\newblock \emph{Proceedings of the ACM on Human-Computer Interaction},
  3\penalty0 (CSCW):\penalty0 1--24, 2019.

\bibitem[Guidotti et~al.(2018)Guidotti, Monreale, Ruggieri, Turini, Giannotti,
  and Pedreschi]{guidotti2018survey}
Riccardo Guidotti, Anna Monreale, Salvatore Ruggieri, Franco Turini, Fosca
  Giannotti, and Dino Pedreschi.
\newblock A survey of methods for explaining black box models.
\newblock \emph{ACM computing surveys (CSUR)}, 51\penalty0 (5):\penalty0 1--42,
  2018.

\bibitem[Hayashi \& Wakabayashi(2017)Hayashi and
  Wakabayashi]{hayashi2017recidivism}
Yugo Hayashi and Kosuke Wakabayashi.
\newblock Can ai become reliable source to support human decision making in a
  court scene?
\newblock In \emph{Companion of the 2017 ACM Conference on Computer Supported
  Cooperative Work and Social Computing}, CSCW '17 Companion, pp.\  195--198,
  New York, NY, USA, 2017. Association for Computing Machinery.
\newblock \doi{10.1145/3022198.3026338}.
\newblock URL \url{https://doi.org/10.1145/3022198.3026338}.

\bibitem[He et~al.(2016)He, Zhang, Ren, and Sun]{he2016deep}
Kaiming He, Xiangyu Zhang, Shaoqing Ren, and Jian Sun.
\newblock Deep residual learning for image recognition.
\newblock In \emph{Proceedings of the IEEE conference on computer vision and
  pattern recognition}, pp.\  770--778, 2016.

\bibitem[Huang et~al.(2017)Huang, Liu, Van Der~Maaten, and
  Weinberger]{huang2017densely}
Gao Huang, Zhuang Liu, Laurens Van Der~Maaten, and Kilian~Q Weinberger.
\newblock Densely connected convolutional networks.
\newblock In \emph{Proceedings of the IEEE conference on computer vision and
  pattern recognition}, pp.\  4700--4708, 2017.

\bibitem[Ilyas et~al.(2019)Ilyas, Santurkar, Tsipras, Engstrom, Tran, and
  Madry]{ilyas2019adversarial}
Andrew Ilyas, Shibani Santurkar, Dimitris Tsipras, Logan Engstrom, Brandon
  Tran, and Aleksander Madry.
\newblock Adversarial examples are not bugs, they are features.
\newblock \emph{Advances in neural information processing systems}, 32, 2019.

\bibitem[Kermany et~al.(2018)Kermany, Goldbaum, Cai, Valentim, Liang, Baxter,
  McKeown, Yang, Wu, Yan, Dong, Prasadha, Pei, Ting, Zhu, Li, Hewett, Dong,
  Ziyar, Shi, Zhang, Zheng, Hou, Shi, Fu, Duan, Huu, Wen, Zhang, Zhang, Li,
  Wang, Singer, Sun, Xu, Tafreshi, Lewis, Xia, and Zhang]{kermany2018chestxray}
Daniel~S. Kermany, Michael Goldbaum, Wenjia Cai, Carolina~C.S. Valentim,
  Huiying Liang, Sally~L. Baxter, Alex McKeown, Ge~Yang, Xiaokang Wu, Fangbing
  Yan, Justin Dong, Made~K. Prasadha, Jacqueline Pei, Magdalene~Y.L. Ting, Jie
  Zhu, Christina Li, Sierra Hewett, Jason Dong, Ian Ziyar, Alexander Shi, Runze
  Zhang, Lianghong Zheng, Rui Hou, William Shi, Xin Fu, Yaou Duan, Viet~A.N.
  Huu, Cindy Wen, Edward~D. Zhang, Charlotte~L. Zhang, Oulan Li, Xiaobo Wang,
  Michael~A. Singer, Xiaodong Sun, Jie Xu, Ali Tafreshi, M.~Anthony Lewis,
  Huimin Xia, and Kang Zhang.
\newblock Identifying medical diagnoses and treatable diseases by image-based
  deep learning.
\newblock \emph{Cell}, 172\penalty0 (5):\penalty0 1122--1131.e9, 2018.
\newblock ISSN 0092-8674.
\newblock \doi{https://doi.org/10.1016/j.cell.2018.02.010}.
\newblock URL
  \url{https://www.sciencedirect.com/science/article/pii/S0092867418301545}.

\bibitem[Kim et~al.(2016)Kim, Khanna, and Koyejo]{kim2016examples}
Been Kim, Rajiv Khanna, and Oluwasanmi~O Koyejo.
\newblock Examples are not enough, learn to criticize! criticism for
  interpretability.
\newblock \emph{Advances in neural information processing systems}, 29, 2016.

\bibitem[Kingma \& Ba(2014)Kingma and Ba]{kingma2014adam}
Diederik~P Kingma and Jimmy Ba.
\newblock Adam: A method for stochastic optimization.
\newblock \emph{arXiv preprint arXiv:1412.6980}, 2014.

\bibitem[Kleindessner \& Luxburg(2014)Kleindessner and
  Luxburg]{kleindessner2014uniqueness}
Matth{\"a}us Kleindessner and Ulrike Luxburg.
\newblock Uniqueness of ordinal embedding.
\newblock In \emph{Conference on Learning Theory}, pp.\  40--67, 2014.

\bibitem[Kleindessner \& von Luxburg(2017)Kleindessner and von
  Luxburg]{kleindessner2017kernel}
Matth{\"a}us Kleindessner and Ulrike von Luxburg.
\newblock Kernel functions based on triplet comparisons.
\newblock In \emph{Advances in Neural Information Processing Systems}, pp.\
  6807--6817, 2017.

\bibitem[Kolodneer(1991)]{kolodneer1991improving}
Janet~L Kolodneer.
\newblock Improving human decision making through case-based decision aiding.
\newblock \emph{AI magazine}, 12\penalty0 (2):\penalty0 52--52, 1991.

\bibitem[Krizhevsky et~al.(2012)Krizhevsky, Sutskever, and
  Hinton]{krizhevsky2012imagenet}
Alex Krizhevsky, Ilya Sutskever, and Geoffrey~E Hinton.
\newblock Imagenet classification with deep convolutional neural networks.
\newblock \emph{Advances in neural information processing systems}, 25, 2012.

\bibitem[Lai \& Tan(2019)Lai and Tan]{lai+tan:19}
Vivian Lai and Chenhao Tan.
\newblock On human predictions with explanations and predictions of machine
  learning models: A case study on deception detection.
\newblock In \emph{Proceedings of FAT*}, 2019.

\bibitem[Lai et~al.(2020)Lai, Liu, and Tan]{lai+liu+tan:20}
Vivian Lai, Han Liu, and Chenhao Tan.
\newblock ``why is `chicago' deceptive?'' towards building model-driven
  tutorials for humans.
\newblock In \emph{Proceedings of CHI}, 2020.

\bibitem[Lai et~al.(2021)Lai, Chen, Liao, Smith-Renner, and
  Tan]{lai2021towards}
Vivian Lai, Chacha Chen, Q~Vera Liao, Alison Smith-Renner, and Chenhao Tan.
\newblock Towards a science of human-ai decision making: a survey of empirical
  studies.
\newblock \emph{arXiv preprint arXiv:2112.11471}, 2021.

\bibitem[Liao(2000)]{liao2000case}
Shu-hsien Liao.
\newblock Case-based decision support system: Architecture for simulating
  military command and control.
\newblock \emph{European Journal of Operational Research}, 123\penalty0
  (3):\penalty0 558--567, 2000.

\bibitem[Luce(1959)]{luce1959individual}
R~Duncan Luce.
\newblock Individual choice behavior.
\newblock 1959.

\bibitem[Metz et~al.(2016)Metz, Metz, Tiku, Lapowsky, Finley, Thompson,
  Griffith, and Spector]{metz2016two}
C~Metz, C~Metz, N~Tiku, I~Lapowsky, K~Finley, C~Thompson, E~Griffith, and
  M~Spector.
\newblock In two moves, alphago and lee sedol redefined the future. wired,
  2016.

\bibitem[Nguyen et~al.(2021)Nguyen, Kim, and Nguyen]{nguyen2021feature}
Giang Nguyen, Daeyoung Kim, and Anh Nguyen.
\newblock The effectiveness of feature attribution methods and its correlation
  with automatic evaluation scores.
\newblock \emph{Advances in Neural Information Processing Systems},
  34:\penalty0 26422--26436, 2021.

\bibitem[Park et~al.(2015)Park, Neeman, Zhang, Sanghavi, and
  Dhillon]{park2015preference}
Dohyung Park, Joe Neeman, Jin Zhang, Sujay Sanghavi, and Inderjit Dhillon.
\newblock Preference completion: Large-scale collaborative ranking from
  pairwise comparisons.
\newblock In \emph{International Conference on Machine Learning}, pp.\
  1907--1916, 2015.
\newblock URL \url{https://bit.ly/2ObMA1J}.

\bibitem[Paszke et~al.(2019)Paszke, Gross, Massa, Lerer, Bradbury, Chanan,
  Killeen, Lin, Gimelshein, Antiga, Desmaison, Kopf, Yang, DeVito, Raison,
  Tejani, Chilamkurthy, Steiner, Fang, Bai, and Chintala]{pytorch}
Adam Paszke, Sam Gross, Francisco Massa, Adam Lerer, James Bradbury, Gregory
  Chanan, Trevor Killeen, Zeming Lin, Natalia Gimelshein, Luca Antiga, Alban
  Desmaison, Andreas Kopf, Edward Yang, Zachary DeVito, Martin Raison, Alykhan
  Tejani, Sasank Chilamkurthy, Benoit Steiner, Lu~Fang, Junjie Bai, and Soumith
  Chintala.
\newblock Pytorch: An imperative style, high-performance deep learning library.
\newblock In H.~Wallach, H.~Larochelle, A.~Beygelzimer, F.~d\textquotesingle
  Alch\'{e}-Buc, E.~Fox, and R.~Garnett (eds.), \emph{Advances in Neural
  Information Processing Systems 32}, pp.\  8024--8035. Curran Associates,
  Inc., 2019.
\newblock URL
  \url{http://papers.neurips.cc/paper/9015-pytorch-an-imperative-style-high-performance-deep-learning-library.pdf}.

\bibitem[Rajpurkar et~al.(2020)Rajpurkar, O'Connell, Schechter, Asnani, Li,
  Kiani, Ball, Mendelson, Maartens, van Hoving, Griesel, Ng, Boyles, and
  Lungren]{rajpurkar2020medical}
Pranav Rajpurkar, Chloe O'Connell, Amit Schechter, Nishit Asnani, Jason Li,
  Amirhossein Kiani, Robyn~L. Ball, Marc Mendelson, Gary Maartens, Daniël~J.
  van Hoving, Rulan Griesel, Andrew~Y. Ng, Tom~H. Boyles, and Matthew~P.
  Lungren.
\newblock {CheXaid}: deep learning assistance for physician diagnosis of
  tuberculosis using chest x-rays in patients with {HIV}.
\newblock \emph{npj Digital Medicine}, 3\penalty0 (1):\penalty0 1--8, September
  2020.
\newblock ISSN 2398-6352.
\newblock \doi{10.1038/s41746-020-00322-2}.
\newblock URL \url{https://www.nature.com/articles/s41746-020-00322-2}.

\bibitem[Ribeiro et~al.(2016)Ribeiro, Singh, and Guestrin]{ribeiro2016should}
Marco~Tulio Ribeiro, Sameer Singh, and Carlos Guestrin.
\newblock " why should i trust you?" explaining the predictions of any
  classifier.
\newblock In \emph{Proceedings of the 22nd ACM SIGKDD international conference
  on knowledge discovery and data mining}, pp.\  1135--1144, 2016.

\bibitem[Ribeiro et~al.(2018)Ribeiro, Singh, and Guestrin]{ribeiro2018anchors}
Marco~Tulio Ribeiro, Sameer Singh, and Carlos Guestrin.
\newblock Anchors: High-precision model-agnostic explanations.
\newblock In \emph{Proceedings of the AAAI conference on artificial
  intelligence}, volume~32, 2018.

\bibitem[Selvaraju et~al.(2017)Selvaraju, Cogswell, Das, Vedantam, Parikh, and
  Batra]{selvaraju2017grad}
Ramprasaath~R Selvaraju, Michael Cogswell, Abhishek Das, Ramakrishna Vedantam,
  Devi Parikh, and Dhruv Batra.
\newblock Grad-cam: Visual explanations from deep networks via gradient-based
  localization.
\newblock In \emph{Proceedings of the IEEE international conference on computer
  vision}, pp.\  618--626, 2017.

\bibitem[Shrikumar et~al.(2017)Shrikumar, Greenside, and
  Kundaje]{shrikumar2017learning}
Avanti Shrikumar, Peyton Greenside, and Anshul Kundaje.
\newblock Learning important features through propagating activation
  differences.
\newblock In \emph{International conference on machine learning}, pp.\
  3145--3153. PMLR, 2017.

\bibitem[Silver et~al.(2016)Silver, Huang, Maddison, Guez, Sifre, Van
  Den~Driessche, Schrittwieser, Antonoglou, Panneershelvam, Lanctot,
  et~al.]{silver2016mastering}
David Silver, Aja Huang, Chris~J Maddison, Arthur Guez, Laurent Sifre, George
  Van Den~Driessche, Julian Schrittwieser, Ioannis Antonoglou, Veda
  Panneershelvam, Marc Lanctot, et~al.
\newblock Mastering the game of go with deep neural networks and tree search.
\newblock \emph{nature}, 529\penalty0 (7587):\penalty0 484--489, 2016.

\bibitem[Silver et~al.(2017)Silver, Schrittwieser, Simonyan, Antonoglou, Huang,
  Guez, Hubert, Baker, Lai, Bolton, et~al.]{silver2017mastering}
David Silver, Julian Schrittwieser, Karen Simonyan, Ioannis Antonoglou, Aja
  Huang, Arthur Guez, Thomas Hubert, Lucas Baker, Matthew Lai, Adrian Bolton,
  et~al.
\newblock Mastering the game of go without human knowledge.
\newblock \emph{nature}, 550\penalty0 (7676):\penalty0 354--359, 2017.

\bibitem[Singla et~al.(2014)Singla, Bogunovic, Bart{\'o}k, Karbasi, and
  Krause]{singla2014near}
Adish Singla, Ilija Bogunovic, G{\'a}bor Bart{\'o}k, Amin Karbasi, and Andreas
  Krause.
\newblock Near-optimally teaching the crowd to classify.
\newblock In \emph{International Conference on Machine Learning}, pp.\
  154--162. PMLR, 2014.

\bibitem[Taesiri et~al.(2022)Taesiri, Nguyen, and
  Nguyen]{nguyen2022explanations}
Mohammad~Reza Taesiri, Giang Nguyen, and Anh Nguyen.
\newblock Visual correspondence-based explanations improve ai robustness and
  human-ai team accuracy.
\newblock In \emph{Advances in Neural Information Processing Systems}, 2022.

\bibitem[Tamuz et~al.(2011)Tamuz, Liu, Belongie, Shamir, and
  Kalai]{tamuz2011adaptively}
Omer Tamuz, Ce~Liu, Serge Belongie, Ohad Shamir, and Adam~Tauman Kalai.
\newblock Adaptively learning the crowd kernel.
\newblock In \emph{Proceedings of the 28th International Conference on
  International Conference on Machine Learning}, pp.\  673--680, 2011.
\newblock URL \url{https://bit.ly/2xAshnJ}.

\bibitem[Terada \& Luxburg(2014)Terada and Luxburg]{terada2014local}
Yoshikazu Terada and Ulrike Luxburg.
\newblock Local ordinal embedding.
\newblock In \emph{International Conference on Machine Learning}, pp.\
  847--855, 2014.

\bibitem[Tschandl et~al.(2020)Tschandl, Rinner, Apalla, Argenziano, Codella,
  Halpern, Janda, Lallas, Longo, Malvehy, Paoli, Puig, Rosendahl, Soyer,
  Zalaudek, and Kittler]{tschandl2020medical}
Philipp Tschandl, Christoph Rinner, Zoe Apalla, Giuseppe Argenziano, Noel
  Codella, Allan Halpern, Monika Janda, Aimilios Lallas, Caterina Longo, Josep
  Malvehy, John Paoli, Susana Puig, Cliff Rosendahl, H.~Peter Soyer, Iris
  Zalaudek, and Harald Kittler.
\newblock Human–computer collaboration for skin cancer recognition.
\newblock \emph{Nature Medicine}, 26\penalty0 (8):\penalty0 1229--1234, August
  2020.
\newblock ISSN 1546-170X.
\newblock \doi{10.1038/s41591-020-0942-0}.
\newblock URL \url{https://www.nature.com/articles/s41591-020-0942-0}.

\bibitem[Van~der Maaten \& Hinton(2008)Van~der Maaten and
  Hinton]{van2008visualizing}
Laurens Van~der Maaten and Geoffrey Hinton.
\newblock Visualizing data using t-sne.
\newblock \emph{Journal of machine learning research}, 9\penalty0 (11), 2008.

\bibitem[Van Der~Maaten \& Weinberger(2012)Van Der~Maaten and
  Weinberger]{van2012stochastic}
Laurens Van Der~Maaten and Kilian Weinberger.
\newblock Stochastic triplet embedding.
\newblock In \emph{Machine Learning for Signal Processing (MLSP), 2012 IEEE
  International Workshop on}, pp.\  1--6, 2012.
\newblock URL \url{https://bit.ly/2O2TF8h}.

\bibitem[Wang \& Yin(2021)Wang and Yin]{wang2021explanations}
Xinru Wang and Ming Yin.
\newblock Are explanations helpful? a comparative study of the effects of
  explanations in ai-assisted decision-making.
\newblock In \emph{26th International Conference on Intelligent User
  Interfaces}, IUI '21, pp.\  318–328, New York, NY, USA, 2021. Association
  for Computing Machinery.
\newblock ISBN 9781450380171.
\newblock \doi{10.1145/3397481.3450650}.
\newblock URL \url{https://doi.org/10.1145/3397481.3450650}.

\bibitem[Xiao et~al.(2020)Xiao, Engstrom, Ilyas, and Madry]{xiao2020noise}
Kai Xiao, Logan Engstrom, Andrew Ilyas, and Aleksander Madry.
\newblock Noise or signal: The role of image backgrounds in object recognition.
\newblock \emph{arXiv preprint arXiv:2006.09994}, 2020.

\bibitem[Zhang et~al.(2018)Zhang, Isola, Efros, Shechtman, and
  Wang]{zhang2018perceptual}
Richard Zhang, Phillip Isola, Alexei~A Efros, Eli Shechtman, and Oliver Wang.
\newblock The unreasonable effectiveness of deep features as a perceptual
  metric.
\newblock In \emph{CVPR}, 2018.

\bibitem[Zhou et~al.(2016)Zhou, Khosla, Lapedriza, Oliva, and
  Torralba]{zhou2016learning}
Bolei Zhou, Aditya Khosla, Agata Lapedriza, Aude Oliva, and Antonio Torralba.
\newblock Learning deep features for discriminative localization.
\newblock In \emph{Proceedings of the IEEE conference on computer vision and
  pattern recognition}, pp.\  2921--2929, 2016.

\bibitem[Zhu et~al.(2018)Zhu, Singla, Zilles, and Rafferty]{zhu2018overview}
Xiaojin Zhu, Adish Singla, Sandra Zilles, and Anna~N Rafferty.
\newblock An overview of machine teaching.
\newblock \emph{arXiv preprint arXiv:1801.05927}, 2018.

\end{thebibliography}
\bibliographystyle{iclr2023_conference}

\clearpage

\appendix

\section{\added{Limitations}}

\added{We discuss some of the limitations in our work.}

\para{\added{Limitations of decision support policies.}}
\added{
Our decision support policies are simple first steps towards a more general example selection policy for decision support. There are certain limitations of our selection policies. For example in this work, we only look at selecting two examples from the two classes in binary image classifcation tasks. We encourage future work to explore more selection methods towards effective decision-support.
}

\added{
In addition to the ethical concerns discussed in the main paper and the ethics statement, our neutral decision support and persuasive decision support policies have different limitations and use cases.
Neutral decision support selects the nearest example from each class. Therefore when a test example lies too close to the decision boundary, the test example, in-class example, and out-of-class example may appear too similar to be distinguished by humans.
This is where we may need to select examples further away with different features so that users are more likely to spot the distinction.
Persuasive decision support selects the most similar example in the predicted class and the least similar example in the other class, 
the latter of which has a risk of being an outlier.
This may invite biases about the data distribution of the other class and degrade effectiveness of decision support.
}

\para{\added{Limitations of experimenting with crowdworkers.}}
\added{
There are several limitations of experimenting with crowdworkers. First, crowdworkers may not invest as much time as domain experts in the tasks. Therefore, collected triplets may come from superficial or the salient features among the images. Second, crowdworkers or in general lay people have limited domain knowledge such as basic anatomy of body parts when working with medical image. Therefore it is less likely for them to notice the most important feature in the images. In our CXR task, we mitigate this limitation by providing an instruction and quiz section before our main study that provides basic information about how to examine chest X-rays. However, in other tasks, we may need to provide more detailed instructions and quizzes to help crowdworkers understand the task and in this way polish collected triplets.
}

\added{
As the expertise level of the end users increases, HC should be able to learn a high-quality representation. The effectiveness of our decision support methods may vary due to experts strong domain knowledge, but we would still expect our human-compatible representation to provide more effective decision support than MLE representations.
}

\added{
Our ultimate goal is to apply our method to domain experts. We start with crowdworkers and the positive results are encouraging. We hope these results could be used to convince and invite more domain experts to get involved and work towards an applicable system together in the future.
}

\para{\added{Limitations of design choices in the algorithm.}}
\added{
A number of decision choices were made in the algorithm. 
For example, we use Euclidean distance as the distance metric to be learned for the representation space. 
Experimenting with different kinds of metrics (e.g., in the psychology literature) and exploring the effectiveness of their respective representations in decision support would be an interesting future direction. 
}

\added{
We used ResNet as the backbone network for feature extraction of images due to its competitiveness and popularity. Although model architecture is not the main concern of this paper, one could also plug in other common backbones such as DenseNet \citep{huang2017densely} and ViT \citep{dosovitskiy2021an} into our representation learning algorithm. We leave the exploration of additional architecture and the effectiveness of their learned representation on decision support to future work.
}

\section{Implementation detail}
\label{sec:implementation}

\begin{figure}[t]
    \centering
    \includegraphics[width=\textwidth]{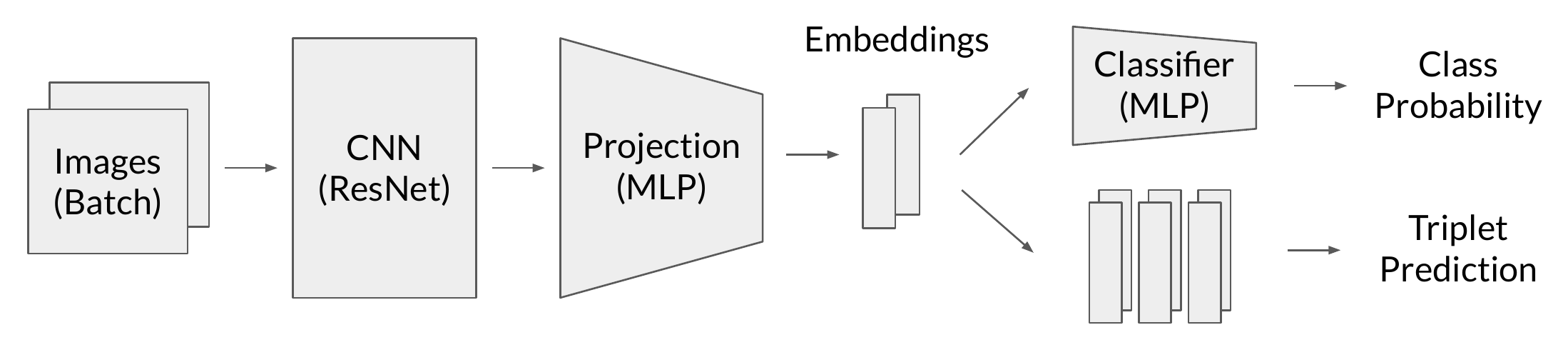}
    \caption{Architecture of the \mtl model.}
    \label{fig:architecture}
\end{figure}

The architecture of our model is presented in \figref{fig:architecture}. 
We first encode image inputs using a Convolutional Neural Network (CNN), and then project the output into an high-dimension representation space with a projection head made of multi-layer perceptron (MLP). 
In our experiments we use one non-linear layer to project the output of the CNN into our representation space. 
For classifcation task we add an MLP classifier head. 
We also use one non-linear layer with softmax activation. 
For triplet prediction, we re-index the representations with the current triplet batch and calculate prediction or loss.
We use the PyTorch framework \cite{pytorch} and the PyTorch Lightning framework \cite{pytorch-lightning} for implementation.
Hyperparameters will be reported in \secref{sec:supp_syn} for models in the synthetic experiments and in \secref{sec:supp_human_bm} and \secref{sec:supp_human_cxr} for models in the human experiments.

\subsection{Computation resources}
We use a computing cluster at our institution. We train our models on nodes with different GPUs including Nvidia GeForce RTX2080Ti, Nvidia GeForce RTX3090, Nvidia Quadro RTX 8000, and Nvidia A40. All models are trained on one allocated node with one GPU access.

\section{Synthetic experiment results}
\label{sec:supp_syn}

\subsection{Hyperparameters}
For our \shortresn backbone we use  We use different controlling strength between classification and human judgment prediction, including $\lambda$s at 0.2, 0.5, and 0.8, and discuss the effect of $\lambda$ in the next section. In contrast to the experiments on BM, we observe that \mtl with 512-dimension embedding shows overall better performance than \mtl with 50-dimension embedding and show results for the latter in the next section.
We use the Adam optimizer \cite{kingma2014adam} with learning rate $1e-4$.
We use a training batch size of $40$ for triplet prediction, and $30$ for classification.

\subsection{Additional results}

\paragraph{Classification and triplet accuracy.} Table~\ref{tab:wv-square-clf-trip} shows how tuning $\lambda$ affects \mtl 's classification and triplet accuracy. Higher $\lambda$ drives \mtl to behave more simlar to \resn while lower \mtl is more similar to \tn.

\begin{table}[t]
    \centering
    \caption{Classification and triplet accuracy of \mtl with different $\lambda$. \tn has no classfication head and no classification accuray.}
    \begin{tabular}{@{}lrr@{}}
    \toprule
    Model  & \textbf{Classification accuracy} & \textbf{Triplet accuracy} \\ \midrule
    \shortresn   & 0.998 $\pm$ 0.003                & 0.673 $\pm$ 0.014         \\
    \shortmtl $\lambda=0.8$ & 0.998 $\pm$ 0.032                & 0.970 $\pm$ 0.024         \\
    \shortmtl $\lambda=0.5$ & 0.995 $\pm$ 0.000                    & 0.972 $\pm$ 0.004         \\
    \shortmtl $\lambda=0.2$ & 0.996 $\pm$ 0.016                & 0.973 $\pm$ 0.039         \\
    \shorttn     & N/A                              & 0.973 $\pm$ 0.016         \\ \bottomrule
    \end{tabular}
    \label{tab:wv-square-clf-trip}
\end{table}

\paragraph{Experiment results on VW with confidence intervals.}
Table~\ref{tab:table1-ci} presents results on VW with \mtl $\lambda=0.5$. This is is simply Table 1 in the main paper with 0.95 confidence intervals.

\begin{table}[t]
  \small
  \centering
  \caption{Experiment results on VW. Models use 512-dimension embeddings; \shortmtl uses $\lambda=0.5$ and filtered triplets. This is the same table as Table~\ref{tab:main-results} and adds confidence intervals.}
  \resizebox{\textwidth}{!}{
  \begin{tabular}{@{}lrrrrrr@{}}
  \toprule
  Alignments   & 50\%   & 80\%   & 83\%  & 92\%  & 92.5\% & 100\%     \\ \midrule
  Weights  &  [0,0,1,1]  & [1,0,1,1] &  [0,1,1,1] &  [1,256,256,256] &  [256,1,256,256] &  [1,1,1,1]  \\ \midrule

  \multicolumn{7}{c}{\textbf{NI-H2H}} \\ \midrule
  \shortmtl vs. \shortresn     & 0.917 $\pm$ 0.064 & 0.914 $\pm$ 0.007 & 0.903 $\pm$ 0.016 & 0.880 $\pm$ 0.022 & 0.872 $\pm$ 0.020 & 0.808 $\pm$ 0.017\\ \midrule

  \multicolumn{7}{c}{\textbf{NO-H2H}} \\ \midrule
  \shortmtl vs. \shortresn     & 0.916 $\pm$ 0.093 & 0.968 $\pm$ 0.011 & 0.946 $\pm$ 0.009 & 0.958 $\pm$ 0.031 & 0.962 $\pm$ 0.008 & 0.970 $\pm$ 0.008\\ \midrule

  \multicolumn{7}{c}{\textbf{\NINO}}       \\ \midrule
  \shortresn     & 0.753 $\pm$ 0.056 & 0.899 $\pm$ 0.025 & 0.896 $\pm $0.044 & 0.897 $\pm$ 0.045 & 0.901 $\pm$ 0.025 & 0.929 $\pm$ 0.028 \\
  \shorttn       & 0.568 $\pm$ 0.049 & 0.775 $\pm$ 0.084 & 0.807 $\pm$ 0.038 & 0.868 $\pm$ 0.012 & 0.877 $\pm$ 0.025 & \textbf{1.000} $\pm$ 0.000\\ 
  \shortmtl      & \textbf{0.759} $\pm$ 0.080 & \textbf{0.901} $\pm$ 0.016& \textbf{0.928} $\pm$ 0.099 & \textbf{0.949} $\pm$ 0.034 & \textbf{0.955} $\pm$ 0.027 & \textbf{1.000} $\pm$ 0.00 \\ \midrule

  \multicolumn{7}{c}{\textbf{\NIFO}}       \\ \midrule
  \shortresn     & 0.704 $\pm$ 0.028 & 0.900 $\pm$ 0.017 & 0.903 $\pm $0.017 & 0.903 $\pm$ 0.017 & 0.901 $\pm$ 0.017 & 0.919 $\pm$ 0.016 \\
  \shorttn       & 0.906 $\pm$ 0.011 & 0.881 $\pm$ 0.043 & 0.863 $\pm$ 0.044 & 0.876 $\pm$ 0.027 & 0.877 $\pm$ 0.076 & \textbf{1.000} $\pm$ 0.000\\ 
  \shortmtl      & \textbf{1.000} $\pm$ 0.000 & \textbf{1.000} $\pm$ 0.000 & \textbf{1.000} $\pm$ 0.000 & \textbf{1.000} $\pm$ 0.000 & \textbf{1.000} $\pm$ 0.000 & \textbf{1.000} $\pm$ 0.000 \\ \midrule
  \end{tabular}}
  \label{tab:table1-ci}
  \end{table}

\paragraph{Results for different $\lambda$.}
In Table~\ref{tab:wv_square_filtered_l=0.2} and Table~\ref{tab:wv_square_filtered_l=0.8} we show experiment results with \mtl using $\lambda=0.2$ and $\lambda=0.8$. We do not observe a clear trend between $\lambda$ and evaluation metric performances. In the main paper we present \mtl with $\lambda=0.5$ as it shows best overall performance.

\begin{table}[t]
  \small
  \centering
  \caption{Experiment results on VW. Models using 512-dimension embeddings; \shortmtl uses $\lambda=0.2$ and filtered triplets.}
  \resizebox{\textwidth}{!}{
  \begin{tabular}{@{}lrrrrrr@{}}
  \toprule
  Alignments   & 50\%   & 80\%   & 83\%  & 92\%  & 92.5\% & 100\%     \\ \midrule
  Weights  &  [0,0,1,1]  & [1,0,1,1] &  [0,1,1,1] &  [1,256,256,256] &  [256,1,256,256] &  [1,1,1,1]  \\ \midrule

  \multicolumn{7}{c}{\textbf{NI-H2H}} \\ \midrule
  \shortmtl vs. \shortresn     & 0.920 $\pm$ 0.005 & 0.890 $\pm$ 0.032 & 0.906 $\pm$ 0.053 & 0.895 $\pm$ 0.016 & 0.862 $\pm$ 0.254 & 0.832 $\pm$ 0.058\\ \midrule

  \multicolumn{7}{c}{\textbf{NO-H2H}} \\ \midrule
  \shortmtl vs. \shortresn     & 0.901 $\pm$ 0.439 & 0.948 $\pm$ 0.095 & 0.970 $\pm$ 0.019 & 0.972 $\pm$ 0.095 & 0.933 $\pm$ 0.154 & 0.981 $\pm$ 0.040\\ \midrule

  \multicolumn{7}{c}{\textbf{\NINO}}       \\ \midrule
  \shortresn     & \textbf{0.753} $\pm$ 0.056 & 0.899 $\pm$ 0.025 & 0.896 $\pm $0.044 & 0.897 $\pm$ 0.045 & 0.901 $\pm$ 0.025 & 0.929 $\pm$ 0.028 \\
  \shorttn       & 0.568 $\pm$ 0.049 & 0.775 $\pm$ 0.084 & 0.807 $\pm$ 0.038 & 0.868 $\pm$ 0.012 & 0.877 $\pm$ 0.025 & \textbf{1.000} $\pm$ 0.000\\
  \shortmtl      & 0.740 $\pm$ 0.540 & \textbf{0.925} $\pm$ 0.127 & \textbf{0.933} $\pm$ 0.064 & \textbf{0.935} $\pm$ 0.000 & \textbf{0.945} $\pm$ 0.349 & \textbf{1.000} $\pm$ 0.000 \\ \midrule

  \multicolumn{7}{c}{\textbf{\NIFO}}       \\ \midrule
  \shortresn     & 0.704 $\pm$ 0.028 & 0.900 $\pm$ 0.017 & 0.903 $\pm $0.017 & 0.903 $\pm$ 0.017 & 0.901 $\pm$ 0.017 & 0.919 $\pm$ 0.016 \\
  \shorttn       & 0.906 $\pm$ 0.011 & 0.881 $\pm$ 0.043 & 0.863 $\pm$ 0.044 & 0.876 $\pm$ 0.027 & 0.877 $\pm$ 0.076 & \textbf{1.000} $\pm$ 0.000\\ 
  \shortmtl      & \textbf{0.996} $\pm$ 0.016 & \textbf{0.995} $\pm$ 0.000 & \textbf{0.998} $\pm$ 0.000 & \textbf{0.996} $\pm$ 0.016 & \textbf{0.995} $\pm$ 0.000 & 0.995 $\pm$ 0.032 \\ \midrule

  \end{tabular}
  }
  \label{tab:wv_square_filtered_l=0.2}
  \end{table}

\begin{table}[t]
  \small
  \centering
  \caption{Experiment results on VW. Models using 512-dimension embeddings; \shortmtl uses $\lambda=0.8$ and filtered triplets.}
  \resizebox{\textwidth}{!}{
  \begin{tabular}{@{}lrrrrrr@{}}
  \toprule
  Alignments   & 50\%   & 80\%   & 83\%  & 92\%  & 92.5\% & 100\%     \\ \midrule
  Weights  &  [0,0,1,1]  & [1,0,1,1] &  [0,1,1,1] &  [1,256,256,256] &  [256,1,256,256] &  [1,1,1,1]  \\ \midrule

  \multicolumn{7}{c}{\textbf{NI-H2H}} \\ \midrule
  \shortmtl vs. \shortresn     & 0.916 $\pm$ 0.082 & 0.869 $\pm$ 0.217 & 0.891 $\pm$ 0.029 & 0.879 $\pm$ 0.066 & 0.853 $\pm$ 0.164 & 0.828 $\pm$ 0.138\\ \midrule

  \multicolumn{7}{c}{\textbf{NO-H2H}} \\ \midrule
  \shortmtl vs. \shortresn     & 0.902 $\pm$ 0.193 & 0.944 $\pm$ 0.093 & 0.959 $\pm$ 0.005 & 0.956 $\pm$ 0.090 & 0.942 $\pm$ 0.026 & 0.969 $\pm$ 0.034\\ \midrule

  \multicolumn{7}{c}{\textbf{\NINO}}       \\ \midrule
  \shortresn     & \textbf{0.753} $\pm$ 0.056 & \textbf{0.899} $\pm$ 0.025 & 0.896 $\pm $0.044 & 0.897 $\pm$ 0.045 & 0.901 $\pm$ 0.025 & 0.929 $\pm$ 0.028 \\
  \shorttn       & 0.568 $\pm$ 0.049 & 0.775 $\pm$ 0.084 & 0.807 $\pm$ 0.038 & 0.868 $\pm$ 0.012 & 0.877 $\pm$ 0.025 & \textbf{1.000} $\pm$ 0.000\\
  \shortmtl      & 0.740 $\pm$ 0.095 & 0.894 $\pm$ 0.111 & \textbf{0.929} $\pm$ 0.079 & \textbf{0.960} $\pm$ 0.032 & \textbf{0.923} $\pm$ 0.127 & \textbf{1.000} $\pm$ 0.000 \\ \midrule

  \multicolumn{7}{c}{\textbf{\NIFO}}       \\ \midrule
  \shortresn     & 0.704 $\pm$ 0.028 & 0.900 $\pm$ 0.017 & 0.903 $\pm $0.017 & 0.903 $\pm$ 0.017 & 0.901 $\pm$ 0.017 & 0.919 $\pm$ 0.016 \\
  \shorttn       & 0.906 $\pm$ 0.011 & 0.881 $\pm$ 0.043 & 0.863 $\pm$ 0.044 & 0.876 $\pm$ 0.027 & 0.877 $\pm$ 0.076 & \textbf{1.000} $\pm$ 0.000\\ 
  \shortmtl      & \textbf{0.998} $\pm$ 0.032 & \textbf{0.995} $\pm$ 0.000 & \textbf{0.998} $\pm$ 0.000 & \textbf{0.998} $\pm$ 0.032 & \textbf{0.995} $\pm$ 0.000 & \textbf{0.999} $\pm$ 0.016 \\ \midrule

  \end{tabular}}
  \label{tab:wv_square_filtered_l=0.8}
  \end{table}

\begin{table}[t]
  \small
  \centering
  \caption{Experiment results on VW. Models use 512-dimension embeddings; \shortmtl uses $\lambda=0.5$ and unfiltered triplets.}
  \resizebox{\textwidth}{!}{
  \begin{tabular}{@{}lrrrrrr@{}}
  \toprule
  Alignments   & 50\%   & 80\%   & 83\%  & 92\%  & 92.5\% & 100\%     \\ \midrule
  Weights  &  [0,0,1,1]  & [1,0,1,1] &  [0,1,1,1] &  [1,256,256,256] &  [256,1,256,256] &  [1,1,1,1]  \\ \midrule

  \multicolumn{7}{c}{\textbf{NI-H2H}} \\ \midrule
  \shortmtl vs. \shortresn     & 0.921 $\pm$ 0.015 & 0.900 $\pm$ 0.035 & 0.920 $\pm$ 0.023 & 0.895 $\pm$ 0.008 & 0.867 $\pm$ 0.034 & 0.846 $\pm$ 0.016\\ \midrule

  \multicolumn{7}{c}{\textbf{NO-H2H}} \\ \midrule
  \shortmtl vs. \shortresn     & 0.951 $\pm$ 0.034 & 0.969 $\pm$ 0.024 & 0.991 $\pm$ 0.002 & 0.991 $\pm$ 0.004 & 0.958 $\pm$ 0.010 & 0.980 $\pm$ 0.023\\ \midrule

  \multicolumn{7}{c}{\textbf{\NINO}}       \\ \midrule
  \shortresn     & \textbf{0.753} $\pm$ 0.056 & \textbf{0.899} $\pm$ 0.025 & \textbf{0.896} $\pm $0.044 & \textbf{0.897} $\pm$ 0.045 & \textbf{0.901} $\pm$ 0.025 & 0.929 $\pm$ 0.028 \\
  \shorttn       & 0.568 $\pm$ 0.049 & 0.775 $\pm$ 0.084 & 0.807 $\pm$ 0.038 & 0.868 $\pm$ 0.012 & 0.877 $\pm$ 0.025 & \textbf{1.000} $\pm$ 0.000\\ 
  \shortmtl      & 0.603 $\pm$ 0.051 & 0.801 $\pm$ 0.025 & 0.848 $\pm$ 0.053 & 0.880 $\pm$ 0.000 & 0.880 $\pm$ 0.081 & \textbf{1.000} $\pm$ 0.000 \\ \midrule

  \multicolumn{7}{c}{\textbf{\NIFO}}       \\ \midrule
  \shortresn     & 0.704 $\pm$ 0.028 & 0.900 $\pm$ 0.017 & 0.903 $\pm $0.017 & 0.903 $\pm$ 0.017 & 0.901 $\pm$ 0.017 & 0.919 $\pm$ 0.016 \\
  \shorttn       & 0.906 $\pm$ 0.011 & 0.881 $\pm$ 0.043 & 0.863 $\pm$ 0.044 & 0.876 $\pm$ 0.027 & 0.877 $\pm$ 0.076 & \textbf{1.000} $\pm$ 0.000\\ 
  \shortmtl      & \textbf{0.996} $\pm$ 0.004 & \textbf{0.999} $\pm$ 0.004 & \textbf{0.996} $\pm$ 0.004 & \textbf{0.996} $\pm$ 0.004 & \textbf{0.996} $\pm$ 0.004 & 0.997 $\pm$ 0.004 \\ \midrule
  \end{tabular}
  }
  \label{tab:wv_square_unfiltered_l=0.5}
  \end{table}

\added{
\para{Number of triplets.} We examine the effect of the number of triplets, showing the results in ~\figref{fig:vw_vary-num}. We decrease number of triplets by powers of 2 and find that H2H preference towards \mtl indeed declines as \shortmtl is less human-compatible with fewer training data.
As for decision support, in \nino \shortmtl performance declines and eventually approaches \resn except an outlier in the end, while in \nifo \shortmtl performance is able to stay 100\% even as the number of triplets declines.
}

\begin{table}[t]
  \begin{minipage}[b]{\textwidth}
    \centering
    \begin{subfigure}[b]{0.32\textwidth}
      \includegraphics[width=\textwidth]{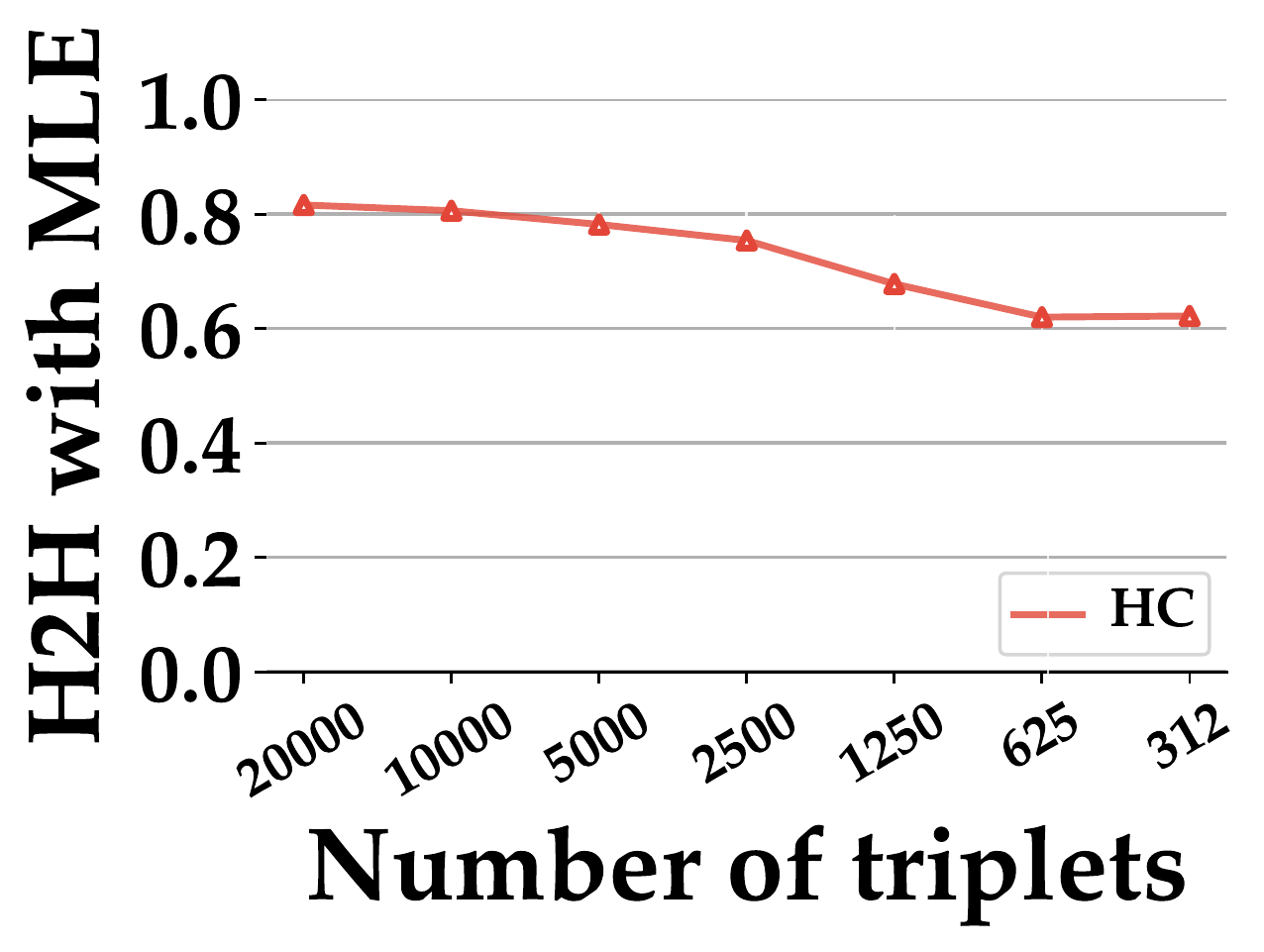}
    \end{subfigure}
    \begin{subfigure}[b]{0.32\textwidth}
    \includegraphics[width=\textwidth]{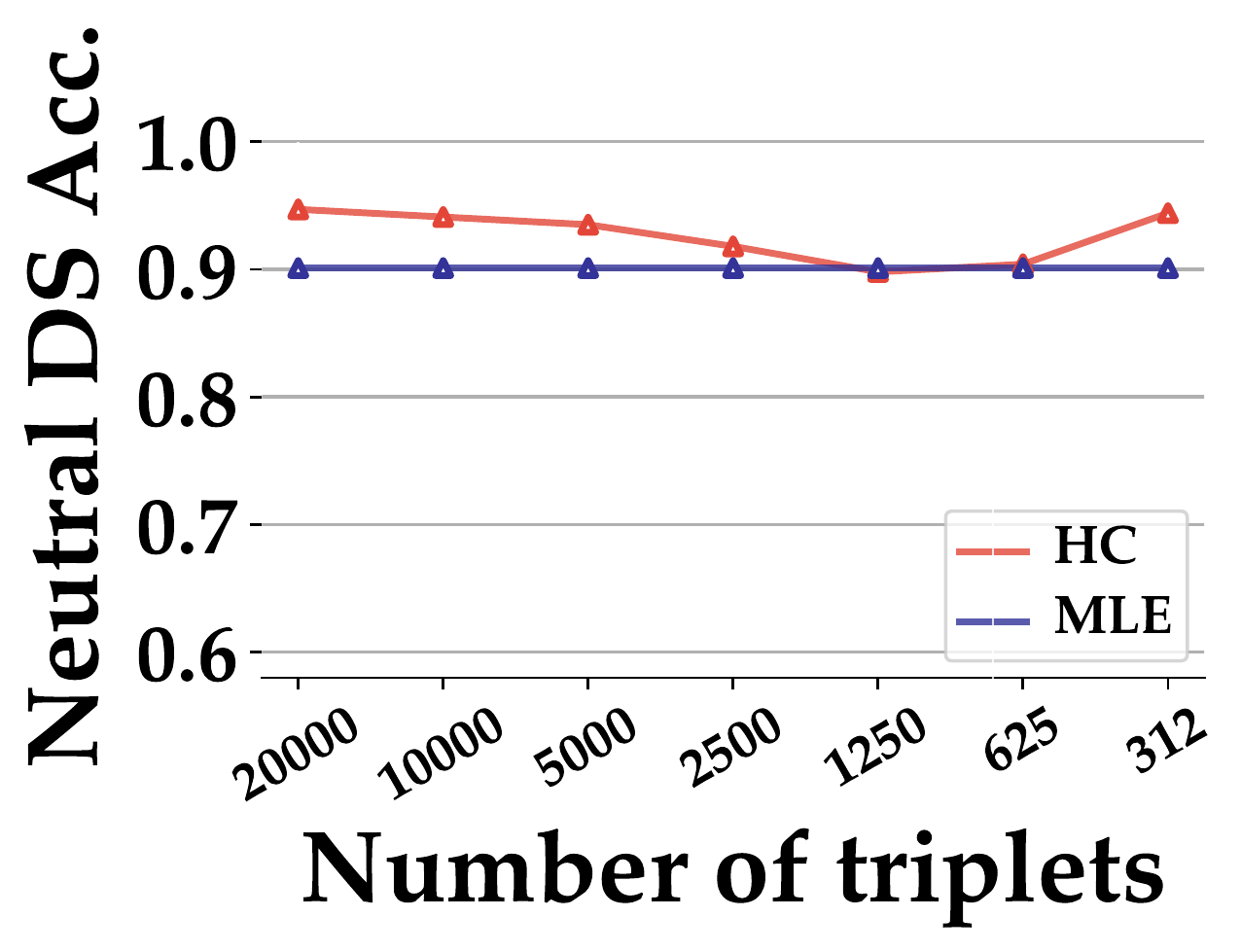}
    \end{subfigure}
    \begin{subfigure}[b]{0.32\textwidth}
    \includegraphics[width=\textwidth]{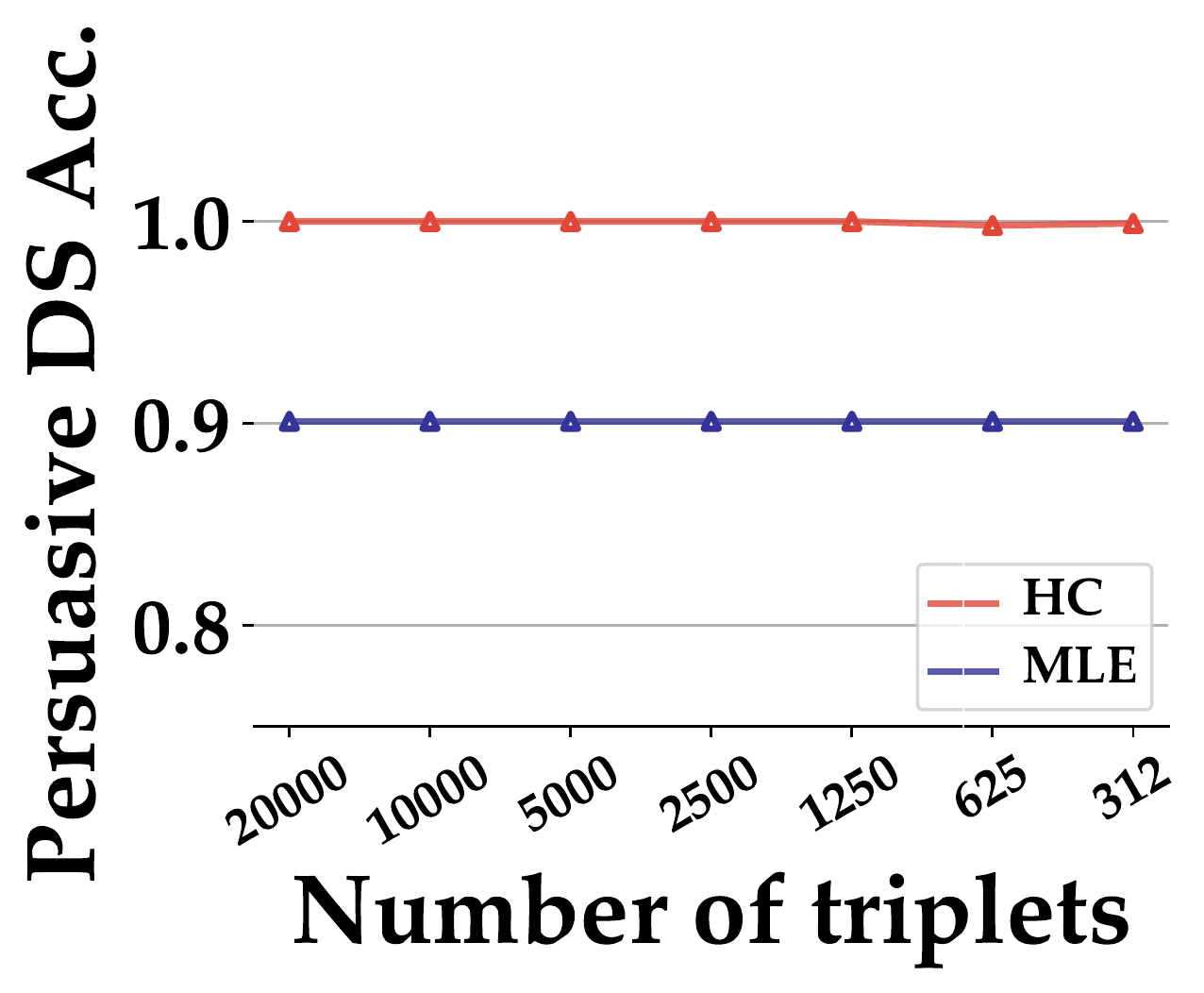}
    \end{subfigure}
    \captionof{figure}{\added{\shortmtl performance declines as the number of triplets decreases, but shows strong \nifo accuracy even with very few triplets.}}
    \label{fig:vw_vary-num}
  \end{minipage}
\end{table}

\begin{figure}[t]
  \centering
  \begin{minipage}[b]{0.49\textwidth}
  \includegraphics[width=\textwidth]{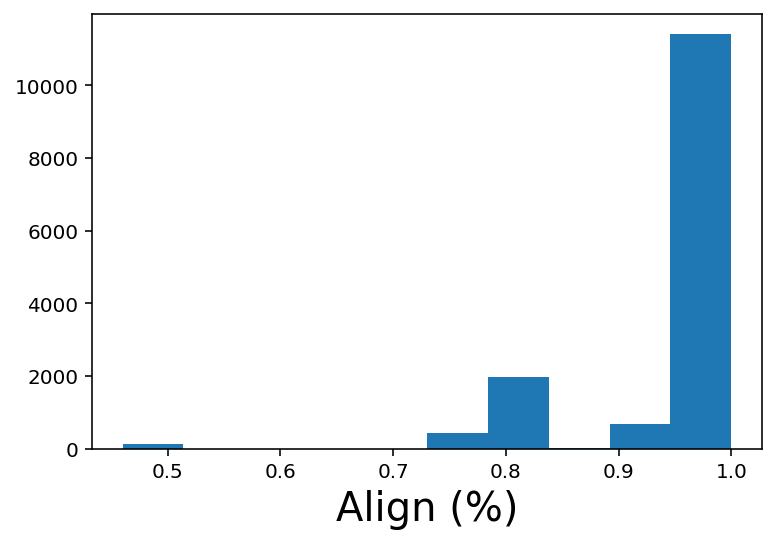}
  \end{minipage}
  \captionof{figure}{Histogram of alignments generated by searching informative weights in powers of 2.}
  \label{fig:align-hist}
\end{figure}

\paragraph{Additional details on weight generation.}
We generate alignment scores by searching through weight combinations of the simulated human visual similarity metrics. We search the weights in powers of 2, from 0 to $2^{10}$, producing a sparse distribution of alignments (\figref{fig:align-hist}). Increasing search range to powers of 10 produces smoother distribution, but the weights are also more extreme and unrealistic. We note that the alignment distribution may vary across different datasets. In our experiments we choose weights and alignments to be as representative to the distribution as possible.

\subsection{Additional decision boundaries}

We create a variant of the VW dataset where the labels are populated by a linear separator. We refer to this dataset as VW-Linear (\figref{fig:linear-dist}).
We find the results are overall similar to the original VW data.

\begin{table}[t]
    \centering
  \begin{minipage}[b]{0.31\textwidth}
    \includegraphics[width=\textwidth]{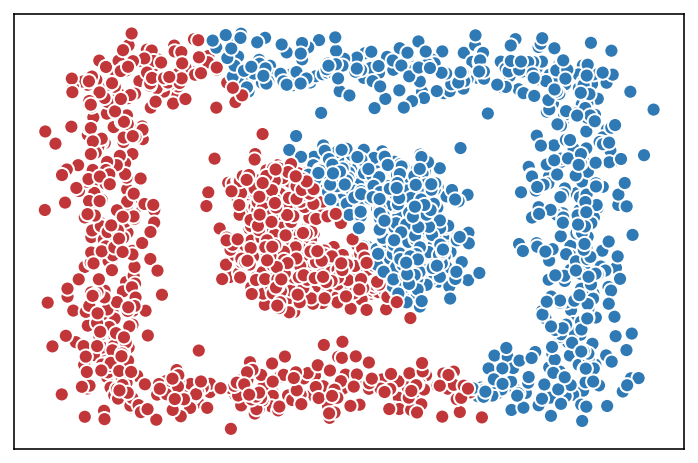}
    \captionof{figure}{VW-Linear
    }
    \label{fig:linear-dist}
  \end{minipage}
  \hfill
  \begin{minipage}[b]{0.63\textwidth}
    
    \caption{\shortmtl performance with different $\lambda$ on VW linear decision boundary data.
    }
    \resizebox{\textwidth}{!}{
    \begin{tabular}{@{}lll@{}}
    \toprule
    Model  & \textbf{Classification accuracy} & \textbf{Triplet accuracy} \\ \midrule
    \shortresn   & 0.993 $\pm$ 0.003                & 0.673 $\pm$ 0.014         \\
    \shortmtl $\lambda=0.8$ & 0.988 $\pm$ 0.032                & 0.968 $\pm$ 0.030         \\
    \shortmtl $\lambda=0.5$ & 0.978 $\pm$ 0.013                    & 0.966 $\pm$ 0.007         \\
    \shortmtl $\lambda=0.2$ & 0.978 $\pm$ 0.032                & 0.970 $\pm$ 0.010         \\
    \shorttn     & N/A                              & 0.976 $\pm$ 0.012         \\ \bottomrule
    \end{tabular}
    \label{tab:lin-clf}
    }

  \end{minipage}
\end{table}

\paragraph{Classification and triplet accuracy.} Table~\ref{tab:lin-clf} shows classification and
triplet accuracy of tuning $\lambda$, showing a similar trend to the previous experiment.

\paragraph{H2H and decision support results}

In Table~\ref{tab:wv_lin_filtered_l=0.5} we present results with the best set of hyperparameter: filtered triplets, 512-dimension embedding, $\lambda=0.5$. We show results for $\lambda=0.2$ in Table~\ref{tab:wv_square_filtered_l=0.2} and $\lambda=0.8$ in Table~\ref{tab:wv_square_filtered_l=0.8}.

Similar to the experiment on VW square decision boundary, we see no clear relation between $\lambda$, embedding dimension and our evaluation metrics.

\begin{table}[t]
  \small
  \centering
  \caption{Experiment results on VW-Linear. %
  Models use 512-dimension embeddings; \shortmtl uses $\lambda=0.5$ and filtered triplets.}
  \begin{tabular}{@{}lrrrr@{}}
  \toprule
  Alignments   & 56\%   & 84\%   & 95\%  & 98.5\%    \\ \midrule
  Weights  &  [0,1,1,1]  & [1,0,1,1] &  [1,1,1,1] &  [32,256,1,1]  \\ \midrule

  \multicolumn{5}{c}{\textbf{NI-H2H}} \\ \midrule
  \shortmtl vs. \shortresn     & 0.913 $\pm$ 0.023 & 0.922 $\pm$ 0.008 & 0.899 $\pm$ 0.020 & 0.848 $\pm$ 0.055\\ \midrule

  \multicolumn{5}{c}{\textbf{NO-H2H}} \\ \midrule
  \shortmtl vs. \shortresn     & 0.932 $\pm$ 0.034 & 0.960 $\pm$ 0.027 & 0.921 $\pm$ 0.013 & 0.928 $\pm$ 0.034\\ \midrule

  \multicolumn{5}{c}{\textbf{\NINO}}       \\ \midrule
  \shortresn     & 0.778 $\pm$ 0.084 & 0.792 $\pm$ 0.144 & 0.839 $\pm$ 0.130 & 0.927 $\pm$ 0.019 \\
  \shorttn       & 0.554 $\pm$ 0.175 & 0.770 $\pm$ 0.318 & 0.950 $\pm$ 0.095 & 0.914 $\pm$ 0.075\\
  \shortmtl      & \textbf{0.841} $\pm$ 0.053 & \textbf{0.911} $\pm$ 0.053 & \textbf{0.967} $\pm$ 0.009 & \textbf{0.961} $\pm$ 0.014 \\ \midrule

  \multicolumn{5}{c}{\textbf{\NIFO}}       \\ \midrule
  \shortresn     & 0.802 $\pm$ 0.249 & 0.815 $\pm$ 0.151 & 0.848 $\pm$ 0.188 & 0.953 $\pm$ 0.051 \\
  \shorttn       & 0.473 $\pm$ 1.016 & 0.653 $\pm$ 1.747 & 0.441 $\pm$ 0.016 & 0.381 $\pm$ 0.474\\
  \shortmtl      & \textbf{0.979} $\pm$ 0.014 & \textbf{0.977} $\pm$ 0.009 & \textbf{0.977} $\pm$ 0.009 & \textbf{0.978} $\pm$ 0.013 \\ \midrule
  \end{tabular}
  \label{tab:wv_lin_filtered_l=0.5}
  \end{table}

\begin{table}[t]
  \small
  \centering
  \caption{Experiment results on VW-Linear. Models use 512-dimension embeddings; \shortmtl uses $\lambda=0.2$ and filtered triplets.}
  \begin{tabular}{@{}lrrrr@{}}
  \toprule
  Alignments   & 56\%   & 84\%   & 95\%  & 98.5\%    \\ \midrule
  Weights  &  [0,1,1,1]  & [1,0,1,1] &  [1,1,1,1] &  [32,256,1,1]  \\ \midrule

  \multicolumn{5}{c}{\textbf{NI-H2H}} \\ \midrule
  \shortmtl vs. \shortresn     & 0.936 $\pm$ 0.024 & 0.921 $\pm$ 0.008 & 0.912 $\pm$ 0.074 & 0.856 $\pm$ 0.034\\ \midrule

  \multicolumn{5}{c}{\textbf{NO-H2H}} \\ \midrule
  \shortmtl vs. \shortresn     & 0.946 $\pm$ 0.032 & 0.974 $\pm$ 0.032 & 0.949 $\pm$ 0.003 & 0.934 $\pm$ 0.029\\ \midrule

  \multicolumn{5}{c}{\textbf{\NINO}}       \\ \midrule
  \shortresn     & 0.778 $\pm$ 0.084 & 0.792 $\pm$ 0.144 & 0.839 $\pm$ 0.130 & 0.927 $\pm$ 0.019 \\
  \shorttn       & 0.554 $\pm$ 0.175 & 0.770 $\pm$ 0.318 & 0.950 $\pm$ 0.095 & 0.914 $\pm$ 0.075\\ 
  \shortmtl      & \textbf{0.845} $\pm$ 0.127 & \textbf{0.880} $\pm$ 0.127 & \textbf{0.956} $\pm$ 0.016 & \textbf{0.956} $\pm$ 0.111 \\ \midrule

  \multicolumn{5}{c}{\textbf{\NIFO}}       \\ \midrule
  \shortresn     & 0.802 $\pm$ 0.249 & 0.815 $\pm$ 0.151 & 0.848 $\pm$ 0.188 & 0.953 $\pm$ 0.051 \\
  \shorttn       & 0.473 $\pm$ 1.016 & 0.653 $\pm$ 1.747 & 0.441 $\pm$ 0.016 & 0.381 $\pm$ 0.474\\ 
  \shortmtl      & \textbf{0.974} $\pm$ 0.016 & \textbf{0.970} $\pm$ 0.064 & \textbf{0.968} $\pm$ 0.064 & \textbf{0.988} $\pm$ 0.032 \\ \midrule
  \end{tabular}
  \label{tab:wv_lin_filtered_l=0.2}
  \end{table}

\begin{table}[t]
  \small
  \centering
  \caption{Experiment results on VW-Linear. Models use 512-dimension embeddings; \shortmtl uses $\lambda=0.8$ and filtered triplets.}
  \begin{tabular}{@{}lrrrr@{}}
  \toprule
  Alignments   & 56\%   & 84\%   & 95\%  & 98.5\%    \\ \midrule
  Weights  &  [0,1,1,1]  & [1,0,1,1] &  [1,1,1,1] &  [32,256,1,1]  \\ \midrule

  \multicolumn{5}{c}{\textbf{NI-H2H}} \\ \midrule
  \shortmtl vs. \shortresn     & 0.906 $\pm$ 0.122 & 0.909 $\pm$ 0.111 & 0.882 $\pm$ 0.135 & 0.848 $\pm$ 0.050\\ \midrule

  \multicolumn{5}{c}{\textbf{NO-H2H}} \\ \midrule
  \shortmtl vs. \shortresn     & 0.926 $\pm$ 0.021 & 0.955 $\pm$ 0.199 & 0.936 $\pm$ 0.053 & 0.912 $\pm$ 0.095\\ \midrule

  \multicolumn{5}{c}{\textbf{\NINO}}       \\ \midrule
  \shortresn     & 0.778 $\pm$ 0.084 & 0.792 $\pm$ 0.144 & 0.839 $\pm$ 0.130 & 0.927 $\pm$ 0.019 \\
  \shorttn       & 0.554 $\pm$ 0.175 & 0.770 $\pm$ 0.318 & \textbf{0.950} $\pm$ 0.095 & 0.914 $\pm$ 0.075\\ 
  \shortmtl      & \textbf{0.824} $\pm$ 0.175 & \textbf{0.895} $\pm$ 0.159 & \textbf{0.950} $\pm$ 0.032 & \textbf{0.969} $\pm$ 0.016 \\ \midrule

  \multicolumn{5}{c}{\textbf{\NIFO}}       \\ \midrule
  \shortresn     & 0.802 $\pm$ 0.249 & 0.815 $\pm$ 0.151 & 0.848 $\pm$ 0.188 & 0.953 $\pm$ 0.051 \\
  \shorttn       & 0.473 $\pm$ 1.016 & 0.653 $\pm$ 1.747 & 0.441 $\pm$ 0.016 & 0.381 $\pm$ 0.474\\ 
  \shortmtl      & \textbf{0.981} $\pm$ 0.048 & \textbf{0.964} $\pm$ 0.206 & \textbf{0.961} $\pm$ 0.175 & \textbf{0.978} $\pm$ 0.064 \\ \midrule
  \end{tabular}
  \label{tab:temp}
  \end{table}

\section{Human subject study on Butterflies v.s. Moths}
\label{sec:supp_human_bm}

\subsection{Dataset}
\begin{figure}[t]
    \centering
    \begin{subfigure}{0.24\textwidth}
      \includegraphics[width=\textwidth]{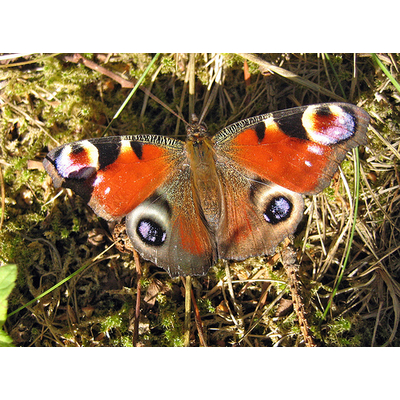}
      \caption{Ringlet Butterfly}
    \end{subfigure}
    \begin{subfigure}{0.24\textwidth}
        \includegraphics[width=\textwidth]{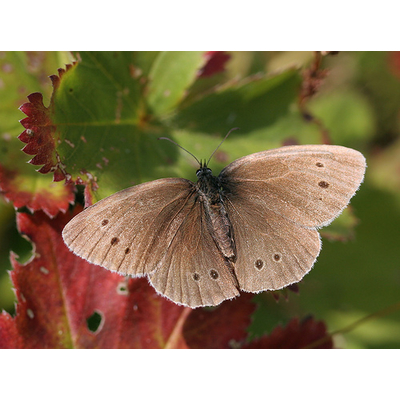}
        \caption{Peacock Butterfly}
    \end{subfigure}
    \begin{subfigure}{0.24\textwidth}
        \includegraphics[width=\textwidth]{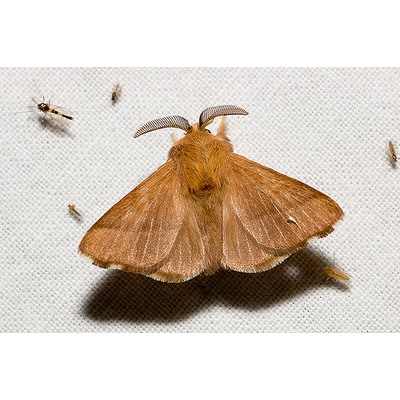}
        \caption{Caterpiller Moth}
    \end{subfigure}
    \begin{subfigure}{0.24\textwidth}
        \includegraphics[width=\textwidth]{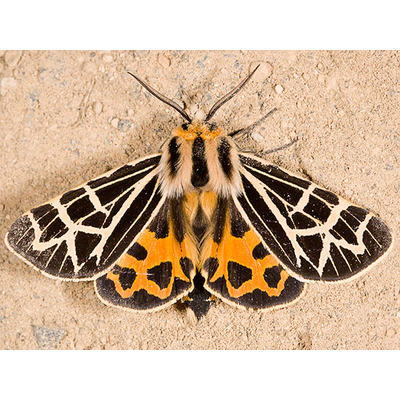}
        \caption{Tiger Moth}
    \end{subfigure}
    \caption{An example of each species in the BM dataset.}
    \label{fig:bm-species}
\end{figure}

Our BM dataset include four species of butterflies and moths including: Peacock Butterfly, Ringlet Butterfly, Caterpiller Moth, and Tiger Moth. An example of each species is shown in Fig \ref{fig:bm-species}. 

\subsection{Hyperparameters}
We use different controlling strength between classification and human judgment prediction, including $\lambda$s at 0.2, 0.5, and 0.8.
We use the Adam optimizer \cite{kingma2014adam} with learning rate $1e-4$.
Our training batch size is $120$ for triplet prediction, and $30$ for classification.
All models are trained for 50 epoches. The checkpoint with the lowest validation total loss in each run is selected for evaluations and applications.

\subsection{Classification and Triplet Accuracy}

\begin{table}[t]
    \centering
    \caption{Classification and triplet accuracy of BM models.
    }
    \begin{tabular}{@{}lcc@{}}
        \toprule
        Model  & \textbf{Classification accuracy} & \textbf{Triplet accuracy} \\ \midrule
        \multicolumn{3}{c}{\textbf{Dimension 50}} \\ \midrule
        \shortresn   & 0.975  & 0.610         \\
        \shortmtl    & 0.975  & 0.762         \\
        \mtlfiltered & 0.975  & 0.707         \\
        \shorttn     & N/A    & 0.759         \\
        \tnfiltered  & N/A    & 0.721         \\ \midrule
        \multicolumn{3}{c}{\textbf{Dimension 512}} \\ \midrule
        \shortresn   & 0.975  & 0.631         \\
        \shortmtl    & 1.000  & 0.741         \\
        \mtlfiltered & 0.975  & 0.709         \\
        \shorttn     & N/A    & 0.748         \\
        \tnfiltered  & N/A    & 0.732         \\ \bottomrule
    \end{tabular}
    \label{tab:bm-models}
\end{table}
We present the test-time classification and triplet accuracy of our models in Table \ref{tab:bm-models}. Both \shortresn and \shortmtl achieve above 97.5\% classification accuracy. \shortmtl in the 512-dimension unfiltered setting achieve 100.0\% classification accuracy. Both \shorttn and \shortmtl achieve above 70.7\% triplet accuracy. Both \shorttn and \shortmtl achieve the highest triplet accuracy in the 50-dimension unfiltered setting with triplet accuracy at 75.9\% and 76.2\% respectively.
\added{Filtering out class-inconsistent triplets removes 15.75\% of the triplet annotations in this dataset.}

We also evaluate the pretrained LPIPS metric \cite{zhang2018perceptual} on our triplet test set as baselines for learning perceptual similarity.
Results with AlexNet backbone and VGG backbone are at 54.5\% and 55.0\% triplet accuracy respectively, suggesting that \shorttn and \shortmtl provides much better triplet accuracy in this task.

\subsection{\added{Effect of triplet amount and type}}
\begin{table}[t]
    \begin{minipage}[b]{\textwidth}
        \centering
        \begin{subfigure}[b]{0.26\textwidth}
        \includegraphics[width=\textwidth]{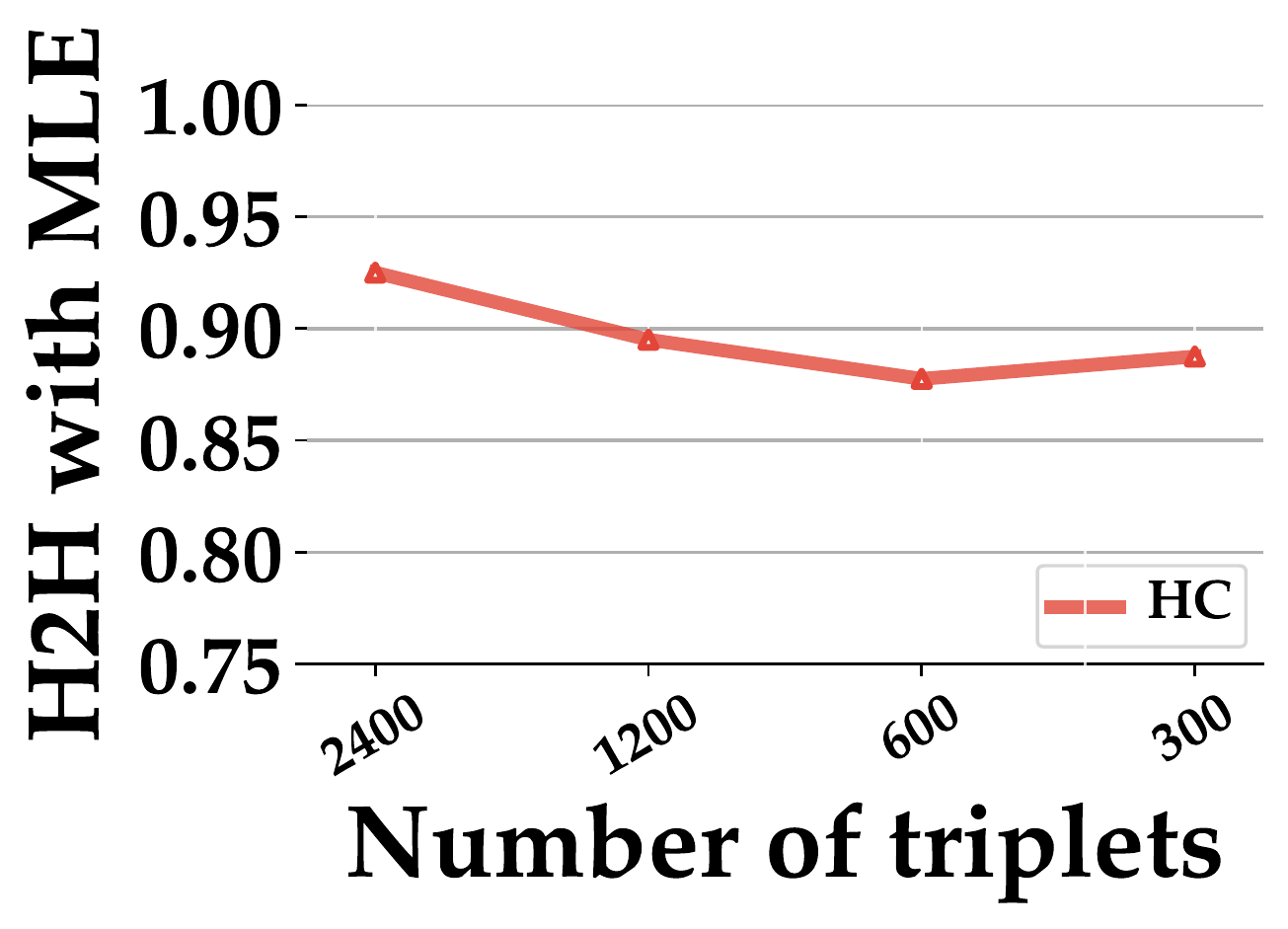}
        \end{subfigure}
        \begin{subfigure}[b]{0.32\textwidth}
        \includegraphics[width=\textwidth]{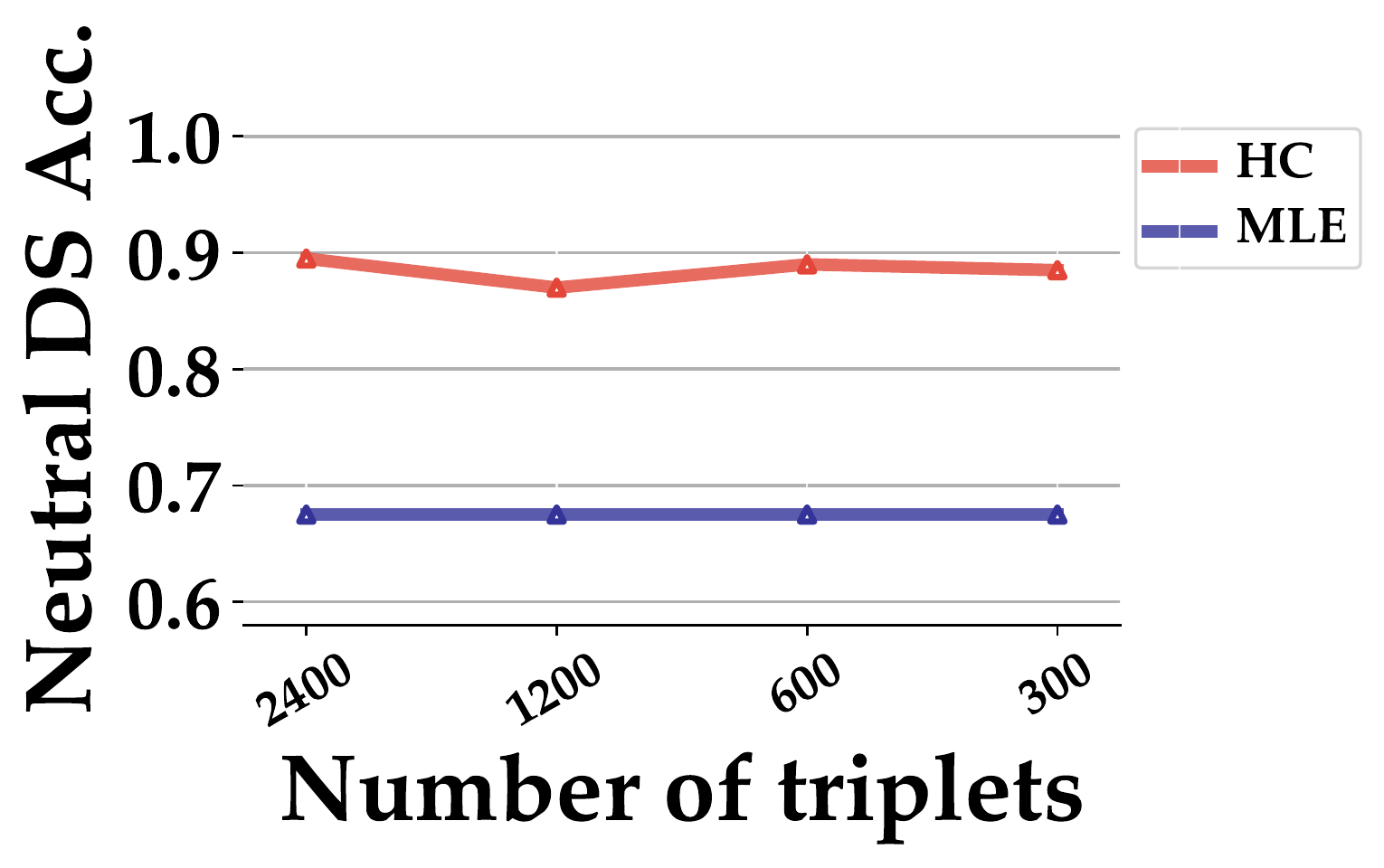}
        \end{subfigure}
        \begin{subfigure}[b]{0.32\textwidth}
        \includegraphics[width=\textwidth]{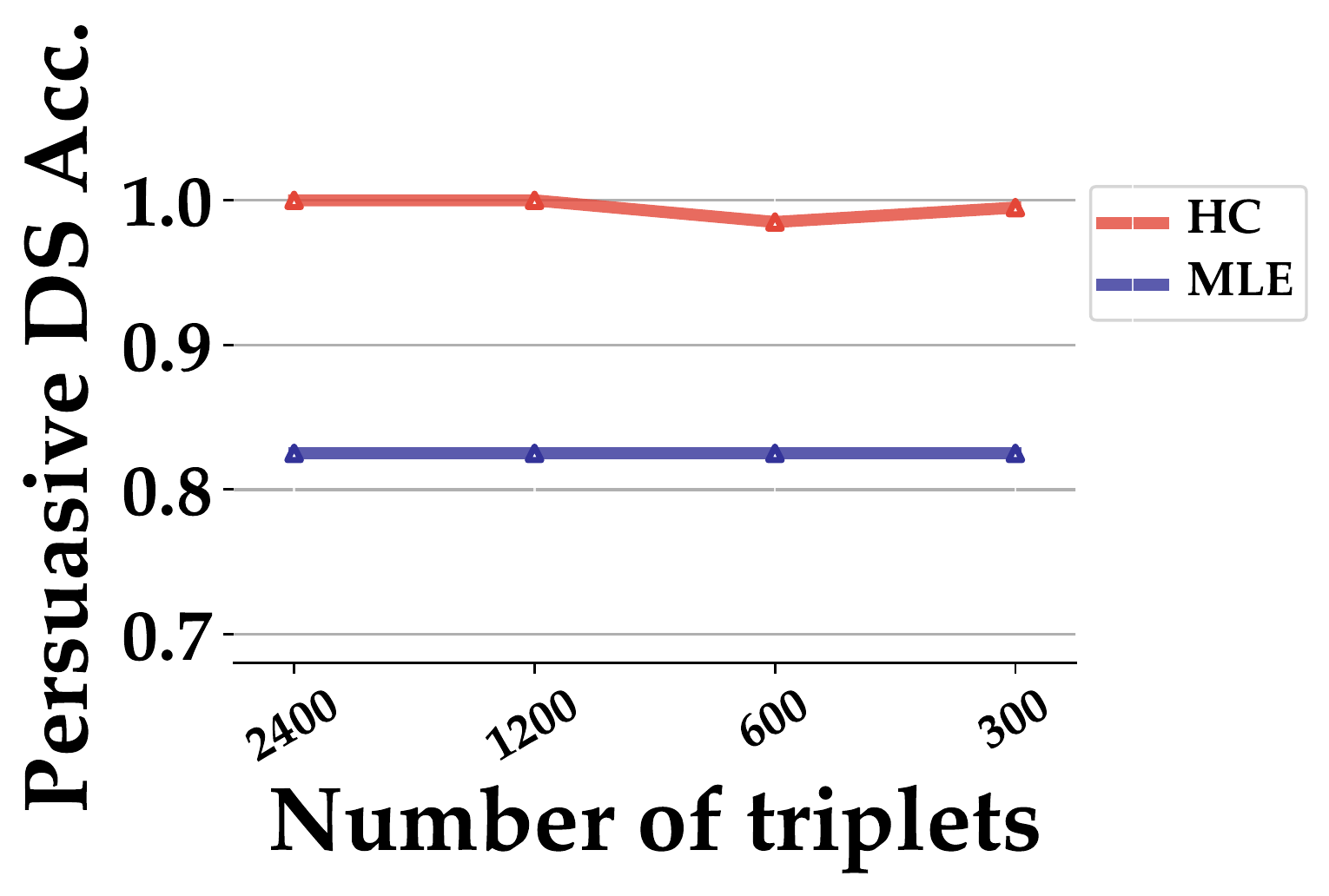}
        \end{subfigure}
        \captionof{figure}{\added{\shortmtl performance declines as the number of triplets decreases, but shows strong \nifo accuracy even with very few triplets.}}
        \label{fig:bm-vary-num}
    \end{minipage}
\end{table}

\added{
We evaluate the effect of the number of triplets on our models in \figref{fig:bm-vary-num}. Similar to the VW experiments, H2H preference towards \mtl and \nino performance decrease as the number of triplets decreases. \Mtl achieve strong \nifo performance even with very few triplets.
}

\subsection{Model Evaluation with Synthetic Agent}

We trained models with different configurations. We mainly discuss two factors: 1) filtering out class-inconsistent triplets or not; 2) a large dimension at 512 vs. a small dimension at 50 for the output representations. We also tried different hyperparameters such as different $\lambda$s that control the strength of the classification loss and triplet margin loss as well as different random seeds.
We select the best \shorttn\ / \shortmtl\ / \shortresn in each filtering-dimension configuration with the highest average of test classification accuracy and test triplet accuracies.

\para{Label accuracy and triplet accuracy.}
As this task is relatively simple, both \shortresn and \shortmtl achieves test accuracy of above 97.5\%. In fact, \shortmtl without filtering out class-inconsistent triplets achieved 100\%.
Note that \shorttn cannot classify alone.
As for triplet accuracy, as expected, both \shortmtl and \shorttn outperform \shortresn.
Dimensionality does not affect triplet accuracy, but
filtering out class-inconsistent triplets decrease triplet accuracy (76.2\% vs. 70.7\% with 50 dimensions, 74.1\% vs. 70.9\% with 512 dimensions).
This is because filtering creates a distribution shift of the triplet annotations, and limits the models' ability to learn general human visual similarity.

To run synthetic experiments for case-based decision support, we select the \shorttn with the best test triplet accuracy as our synthetic agent, and then evaluate the examples produced by all representations.
We do not show results of \shorttn as we use it as the synthetic agent.

\begin{table}[t]
    \centering
    \begin{minipage}[b]{0.48\textwidth}
        \captionof{table}{BM H2H preference results with synthetic agent.}
        \small
\centering
\begin{tabular}{@{}lrr@{}}
    \toprule
    Dimensions  & 50 & 512  \\ \midrule
    \multicolumn{3}{c}{\textbf{NI H2H with \shortresn}} \\ \midrule
    \shortmtl     & 0.838 & 0.575 \\ 
    \shortmtl filtered & 0.863 & 0.725 \\ \midrule
    \multicolumn{3}{c}{\textbf{NO H2H with \shortresn}} \\ \midrule
    \shortmtl     & 0.775 & 0.925 \\
    \shortmtl filtered  & 0.700 & 0.775 \\ \bottomrule
\end{tabular}
        \label{tb:bm_h2h}
    \end{minipage}
    \hfill
    \begin{minipage}[b]{0.48\textwidth}
        \captionof{table}{BM decision support accuracy with synthetic agent.}
        
    \small
    \centering
    \begin{tabular}{@{}lrr@{}}
    \toprule
    Dimensions  & 50 & 512  \\ \midrule
    \multicolumn{3}{c}{\textbf{Neutral Decision Support}}       \\ \midrule
    \shortresn     & 0.675 & 0.875  \\
    \shortmtl      & 0.900 & 0.800 \\ 
    \shortmtl filtered     & 0.875 & 0.900 \\ 
    \midrule
    \multicolumn{3}{c}{\textbf{Persuasive Decision Support}}       \\ \midrule
    \shortresn     & 0.825 & 0.875  \\
    \shortmtl      & 1.000 & 1.000  \\
    \shortmtl  filtered    & 1.000 & 1.000  \\ 
    \bottomrule
    \end{tabular}

        \label{tb:bm_decision}
    \end{minipage}
\end{table}

\para{\Mtl is prefered over \resn in H2H.}
We compare examples selected from different models in different configurations to examples selected by the \shortresn baseline with the same dimensionality.

Table~\ref{tb:bm_h2h} shows how often the synthetic agent prefers the tested model examples to baseline \shortresn examples.
In all settings, the preference towards \shortmtl is above 50\%, but not as high as those in our synthetic experiments with the VW dataset.
Filtering out class-inconsistent triplets improves the preference for the nearest example with the predicted label, while hurting the preference for the nearest out-of-class example.

\para{Decision support simulations shows a large dimension benefits \resn but hurts unfiltered \mtl in \nino.}
We also run simulated decision support with the \shorttn synthetic agent. Table \ref{tb:bm_decision} shows decision support accuracy for different settings.
\shortresn have both higher \nino accuracy and \nifo scores when we use a large dimension at 512. We hypothesize that for \shortresn, reducing dimension may force the network to discard dimensions useful for human judgments but keep dimensions useful for classification. We then use the 512-dimension \shortresn with the highest intrinsic evaluation scores as our \shortresn baseline in later studies.

For \shortmtl, \nino accuracy are in general comparable to 87.5\% score of the 512-dimension \shortresn baseline except unfiltered 512-dimension \shortmtl which has only 80\%.
We hypothesize that representations of large dimension may struggle more with contradicting signals between metric learning and supervised classification in the unfiltered settings.
For \nifo, \shortmtl achieves perfect scores in all settings.

Overall, to proceed with our human-subject experiments, we choose \shortmtl filtered with 50 dimensions as our best \shortmtl as it achieves a good balance between H2H and \nino.
For \shortresn, we choose the representation with 512 dimensions.
We conduct head-to-head comparison between these two representations.
Our synthetic agent prefers \shortmtl in 70\% of the nearest in-class examples and in 97.5\% of the nearest out-of-class examples.

\subsection{Interface}
\begin{figure}
    \centering
    \includegraphics[width=\textwidth]{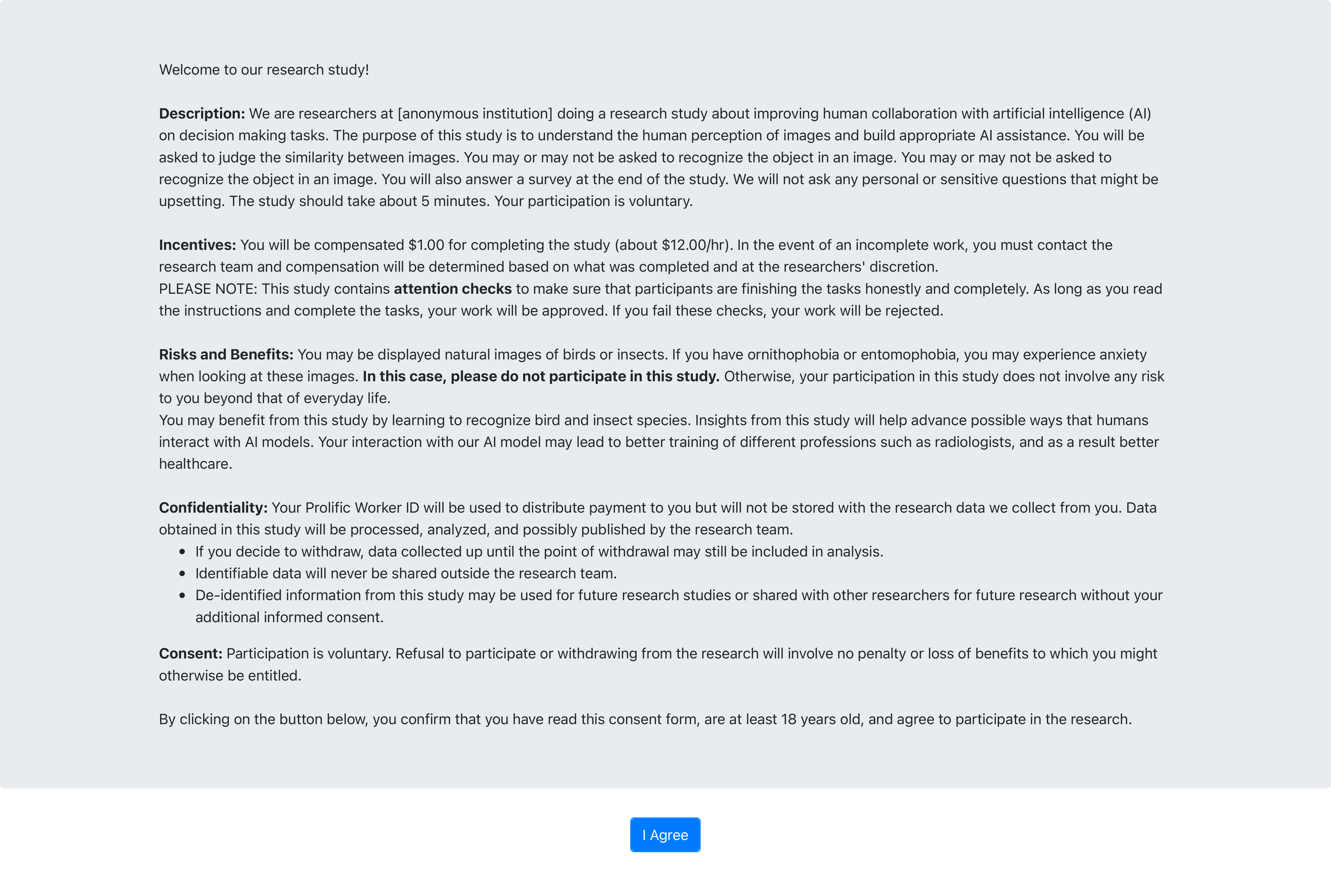}
    \caption{The consent form page on our interface.}
    \label{fig:interface_consent}
\end{figure}

\begin{figure}
    \centering
    \begin{subfigure}[t]{0.48\textwidth}
        \includegraphics[width=\textwidth]{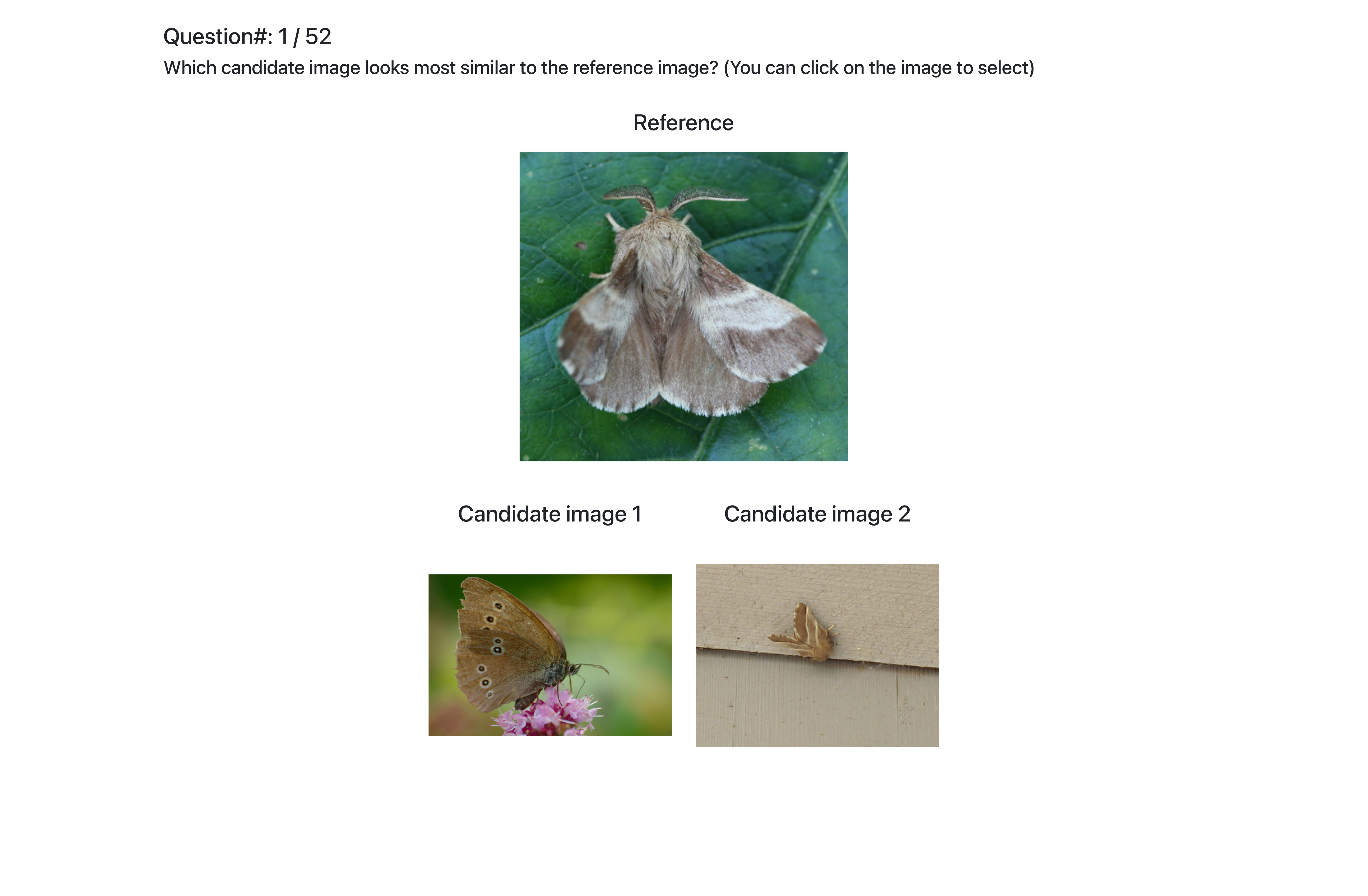}
        \caption{The annotation and head-to-head comparision task questions.}
    \end{subfigure}
    \begin{subfigure}[t]{0.48\textwidth}
        \includegraphics[width=\textwidth]{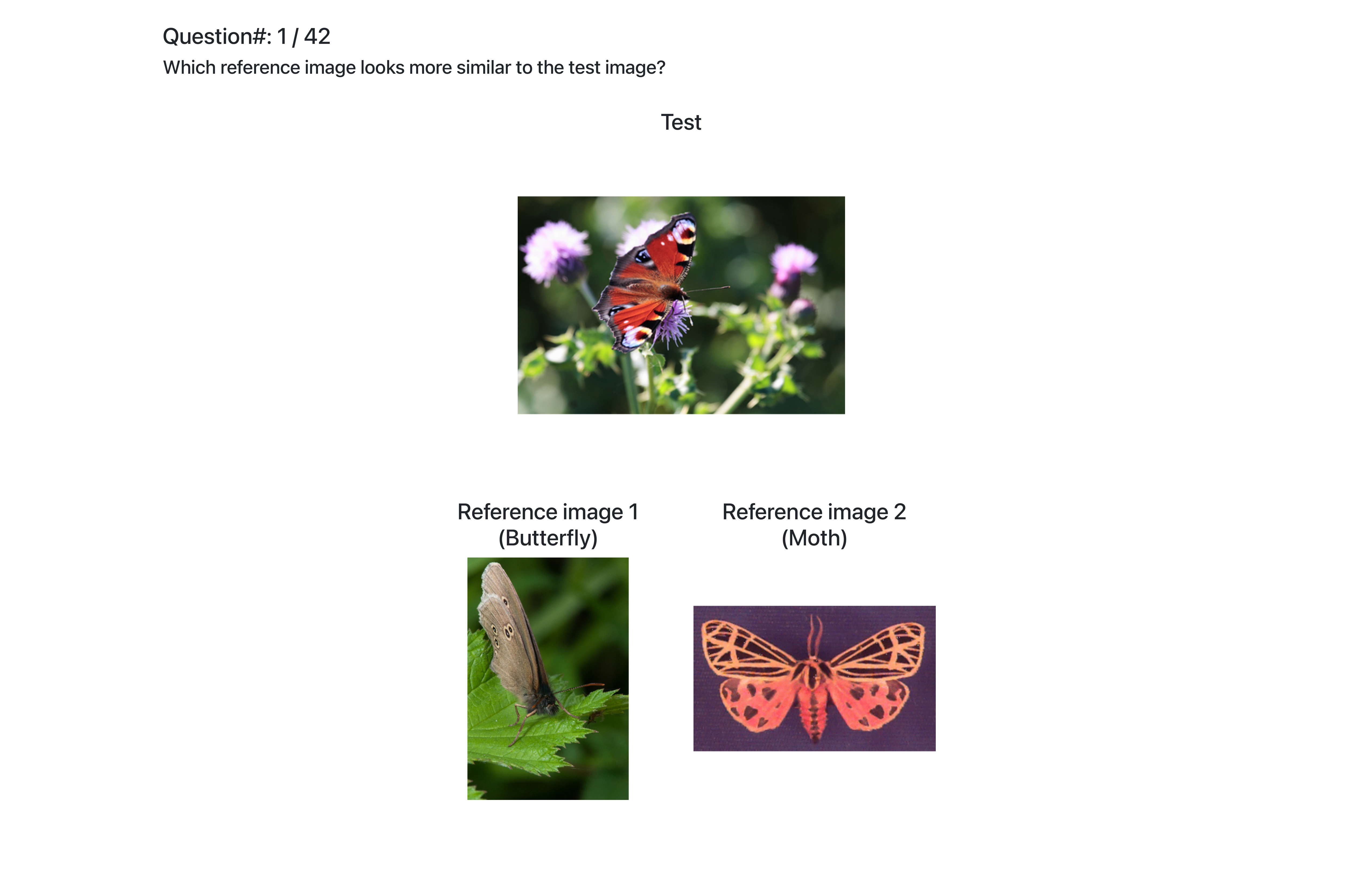}
        \caption{The decision support task questions.}
    \end{subfigure}
    \caption{The task-specific questions for BM.}
    \label{fig:interface_questions}
\end{figure}

\begin{figure}
    \centering
    \begin{subfigure}[t]{0.48\textwidth}
        \includegraphics[width=\textwidth]{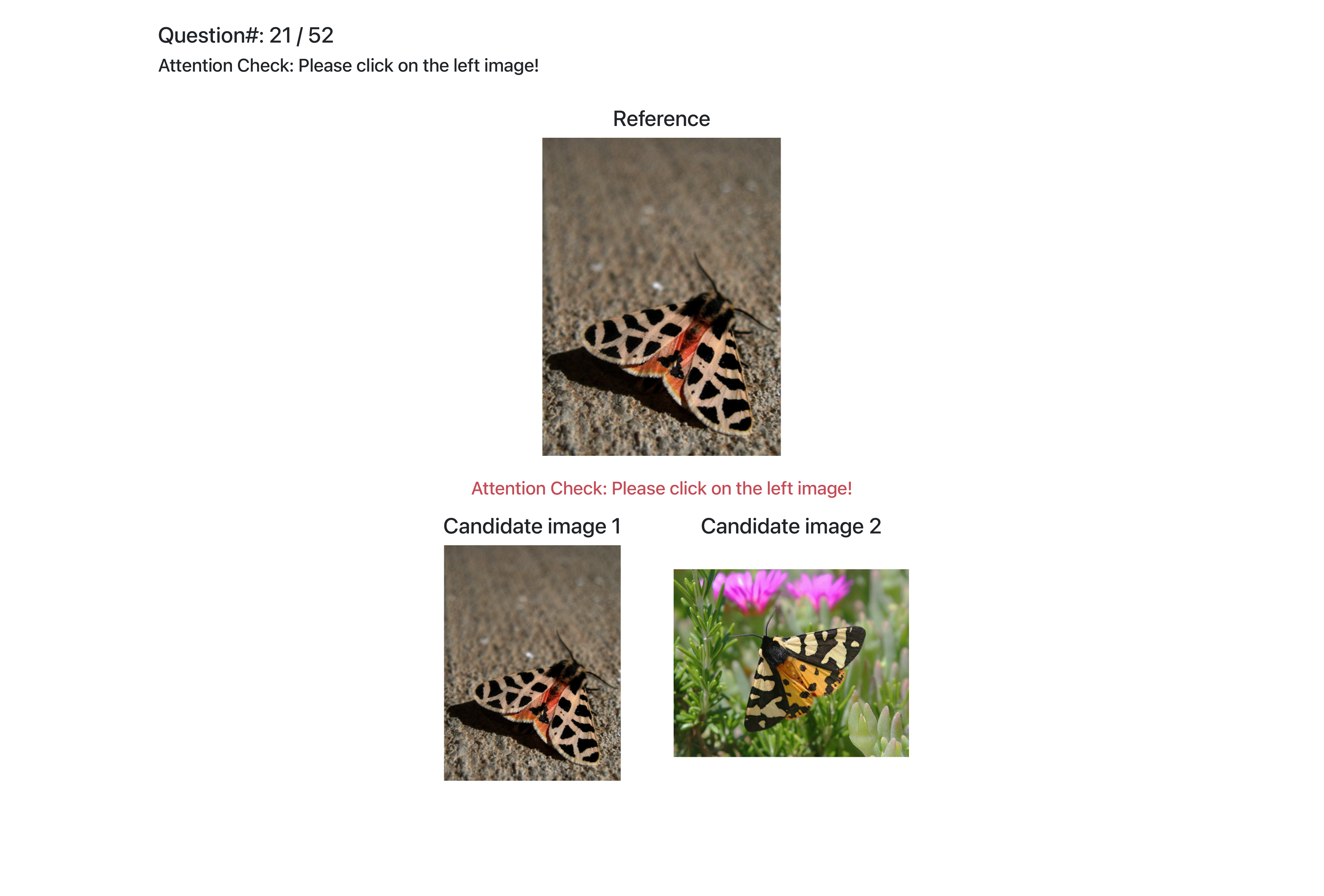}
        \caption{The annotation and head-to-head comparision task attention check questions.}
    \end{subfigure}
    \begin{subfigure}[t]{0.48\textwidth}
        \includegraphics[width=\textwidth]{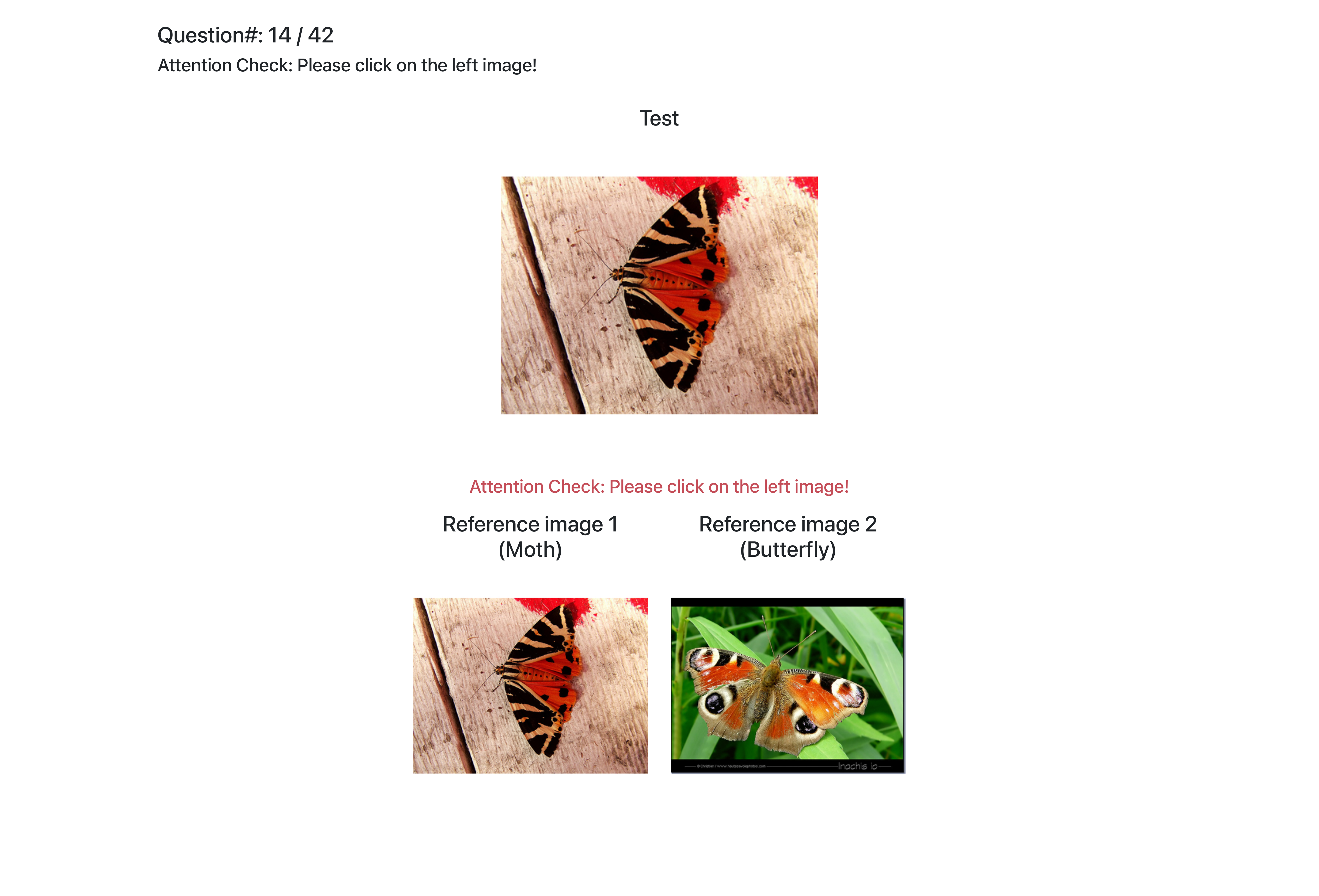}
        \caption{The decision support task attention check questions.}
    \end{subfigure}
    \caption{The task-specific attention check questions for BM.}
    \label{fig:interface_attention}
\end{figure}

\begin{figure}
    \centering
    \begin{subfigure}[t]{0.48\textwidth}
        \includegraphics[width=\textwidth]{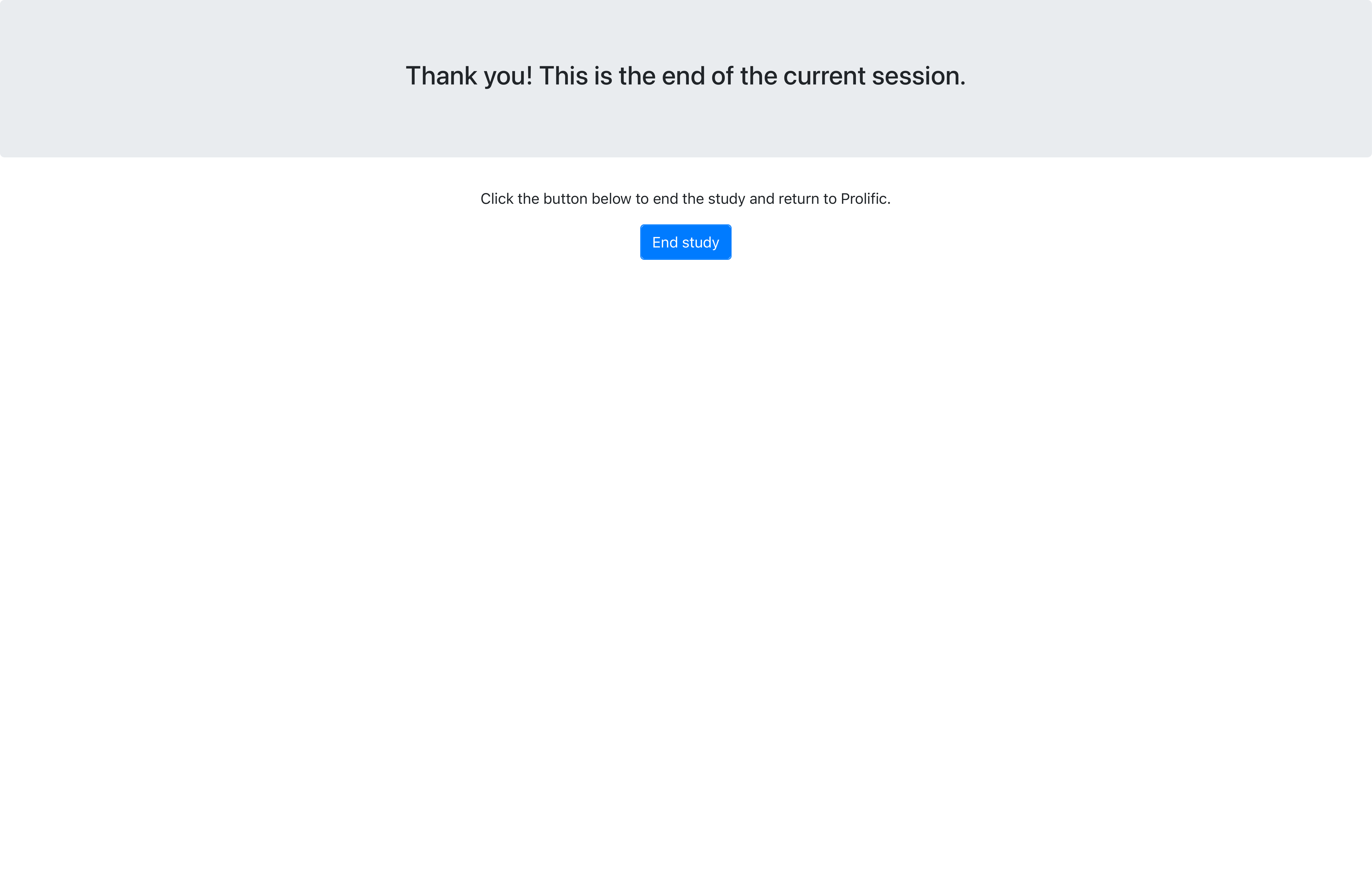}
        \caption{The annotation and head-to-head comparision task end page.}
    \end{subfigure}
    \begin{subfigure}[t]{0.48\textwidth}
        \includegraphics[width=\textwidth]{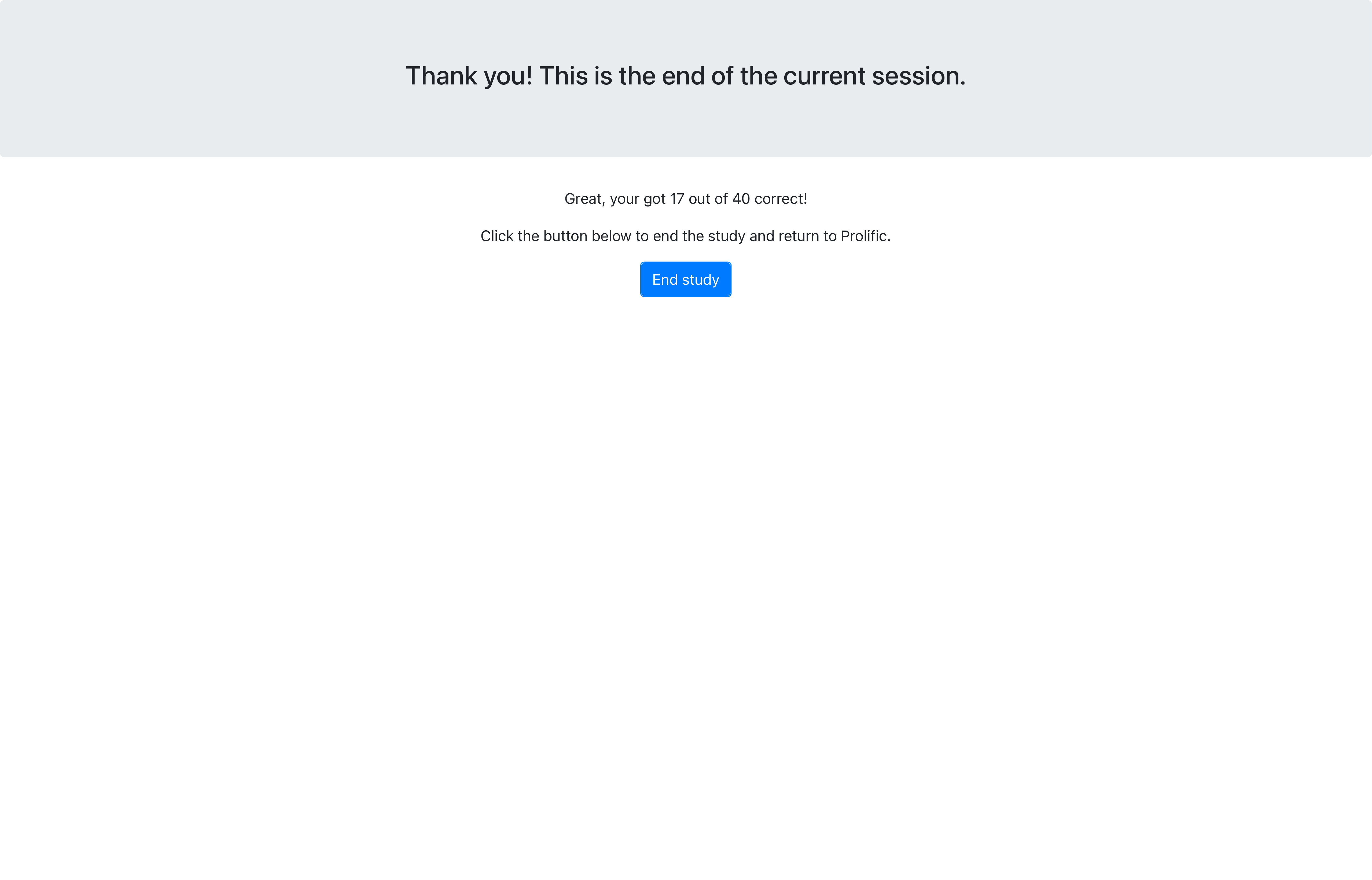}
        \caption{The decision support task end page.\\ }
    \end{subfigure}
    \caption{The task-specific end page on our interface.}
    \label{fig:interface_end}
\end{figure}

We present the screenshots of our interface at the end of the appendix.
Our interface consists of four stages.
Participants will see the consent page at the beginning, as shown in Fig \ref{fig:interface_consent}.
After consent page, participants will see task specific instructions, as shown in Fig \ref{fig:interface_prolific}.
After entering the task, partipants will see the questions, as shown in Fig \ref{fig:interface_questions}.
We also include two attention check questions in all studies to check whether participants are paying attention to the questions.
Following suggestions on Prolific, we design the attention check with explicit instructions, as shown in Fig \ref{fig:interface_attention}.
After finishing all questions, participants will reach the end page and return to Prolific, as shown in Fig \ref{fig:interface_end}.
Our study is reviewed by the Institutional Review Board (IRB) at our institution (IRB22-0388).

\subsection{Crowdsourcing}
We recruit our participants on a crowdsourcing platform: Prolific (www.prolific.co) [April-May 2022].
We conduct three total studies: an annotation study, a decision support study, and a head-to-head comparison study.
We use the default standard sampling on Prolific for participant recruitment.
Eligible participants are limited to those reside in United States.
Participants are not allowed to attempt the same study more than once.

\para{Triplet annotation study}
We recruit 90 participants in total. We conduct a pilot study with 7 participants to test the interface, and recruit 83 participants for the actual collection of annotations. 3 participants fail the attention check questions and their responses are excluded in the results. We spend in total \$76.01 with an average pay at \$10.63 per hour. The median time taken to complete the study is 3'22''.

\para{Decision support study}
We recruit 161 participants in total. 3 participants fail the attention check questions and their responses are excluded in the results. We take the first 30 responses in each conditon to compile the results. We spend in total \$126.40 with an average pay at \$9.32 per hour. The median time taken to complete the study is 3'53''.

\para{Head-to-head comparison study}
We recruit 31 participants in total, where 1 participant fail the attention check questions and their responses are excluded in the results. We spend in total \$24.00 with an average pay at \$9.40 per hour. The median time taken to complete the study is 3'43''.

\section{Human subject study on chest X-rays}
\label{sec:supp_human_cxr}

\subsection{Dataset}
\begin{figure}[t]
    \centering
    \begin{subfigure}{0.32\textwidth}
      \includegraphics[width=\textwidth]{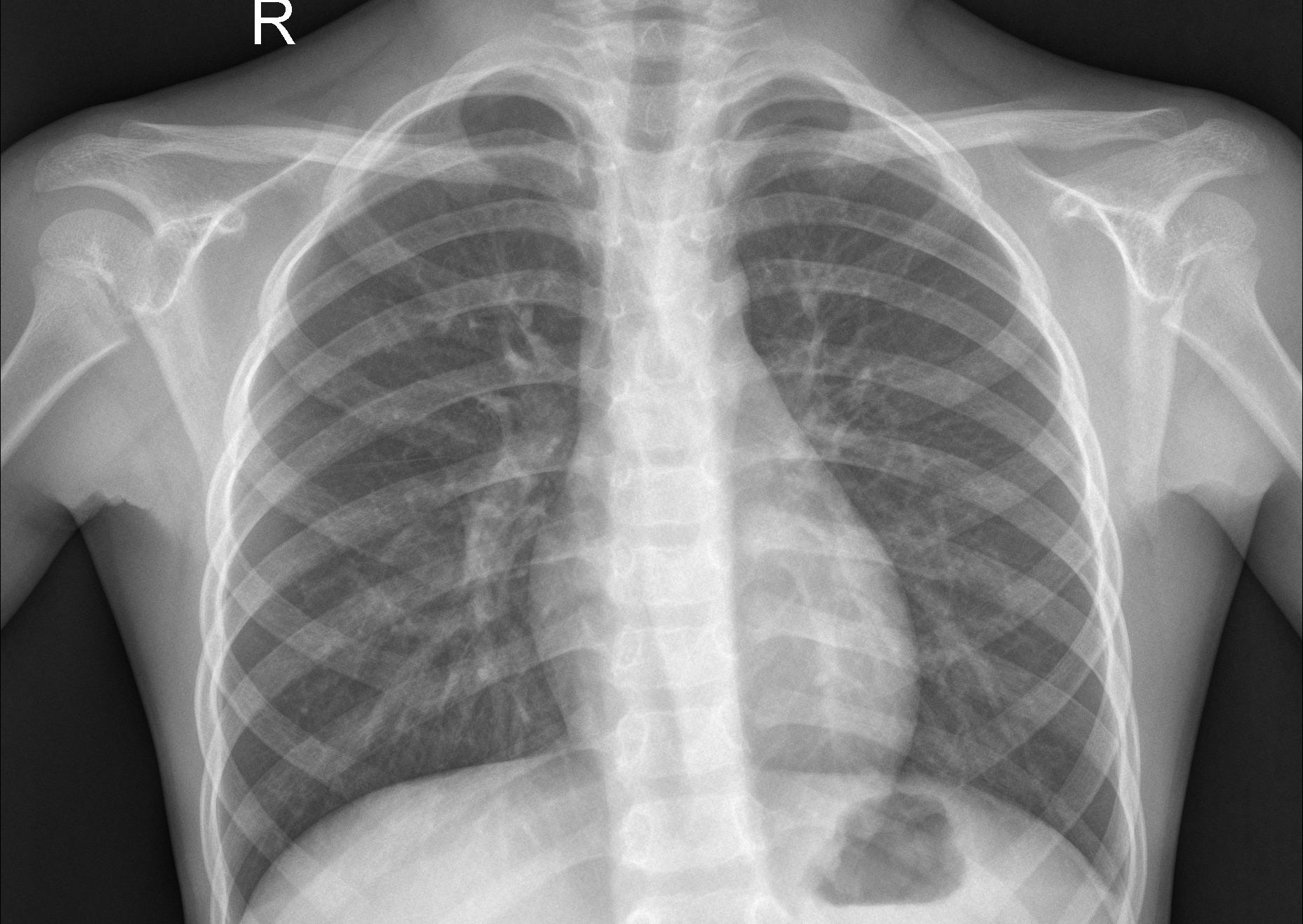}
      \caption{Normal}
    \end{subfigure}
    \hfill
    \begin{subfigure}{0.32\textwidth}
        \includegraphics[width=\textwidth]{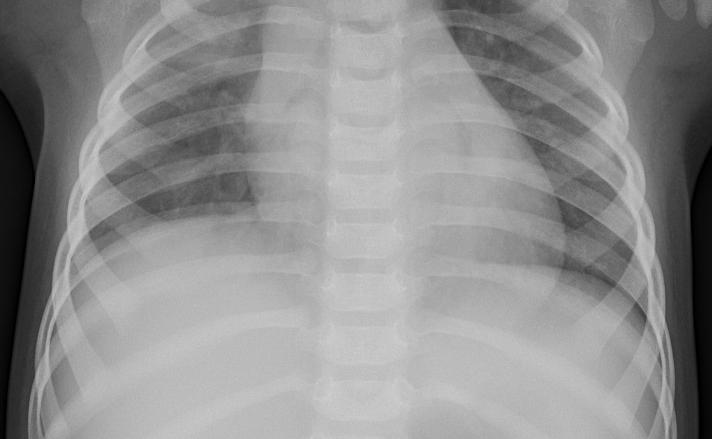}
        \caption{Bacterial pneumonia}
    \end{subfigure}
    \hfill
    \begin{subfigure}{0.32\textwidth}
        \includegraphics[width=\textwidth]{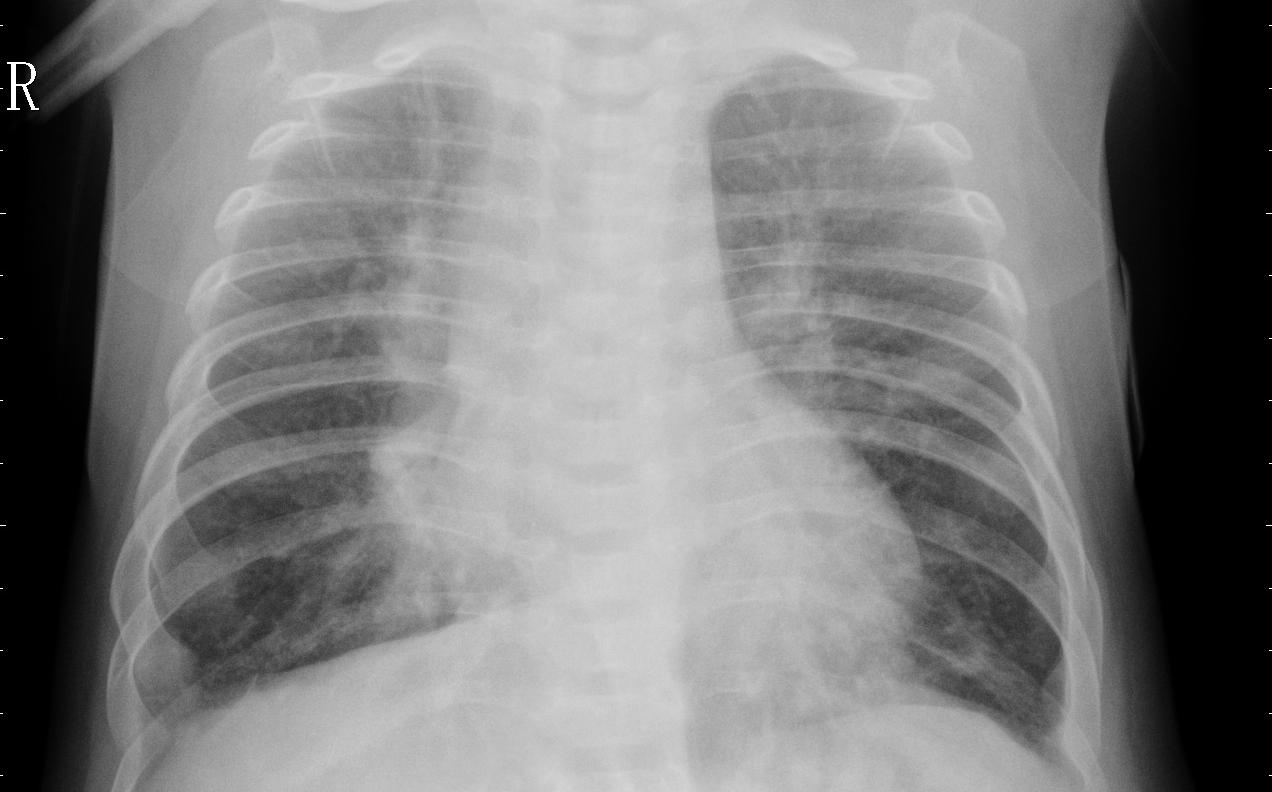}
        \caption{Viral pneumonia}
    \end{subfigure}
    \caption{An example of each image class in the CXR dataset.}
    \label{fig:cxr-species}
\end{figure}

Our CXR dataset is constructed from a subset of the chest X-ray dataset used by \citet{kermany2018chestxray}, which had 5,232 images. We take a balanced subset of 3,166 images, 1,583 characterized as depicting pneumonia and 1,583 normal. The pneumonia class contains bacterial pneumonia and viral pneumoia images, but we do not differentiate them for this study. An example of each image class is shown in Fig \ref{fig:cxr-species}.

\subsection{Hyperparameters}
For CXR experiment, instead of ResNet-18 pretrained from ImageNet, we use a ResNet-18 finetuned on CXR classifcation as our CNN backbone, as we observe it provides better decision support simulation results.
For training our \shortmtl model we use $\lambda$ of 0.5.
We use the Adam optimizer \cite{kingma2014adam} with learning rate $1e-4$.
Our training batch size is $16$ for triplet prediction, and $30$ for classification.
All models are trained for 10 epoches. The checkpoint with the lowest validation total loss in each run is selected for evaluations and applications.

\subsection{Classification and Triplet Accuracy}

\begin{table}[t]
    \centering
    \caption{Classification and triplet accuracy of CXR models.
    }
    \begin{tabular}{@{}lcc@{}}
        \toprule
        Model  & \textbf{Classification accuracy} & \textbf{Triplet accuracy} \\ \midrule
        \multicolumn{3}{c}{\textbf{Dimension 50}} \\ \midrule
        \shortresn   & 0.973  & 0.571         \\
        \shortmtl    & 0.954  & 0.576         \\
        \mtlfiltered & 0.955  & 0.574         \\
        \shorttn     & N/A    & 0.602         \\
        \tnfiltered  & N/A    & 0.587         \\ \midrule
        \multicolumn{3}{c}{\textbf{Dimension 512}} \\ \midrule
        \shortresn   & 0.973  & 0.588         \\
        \shortmtl    & 0.968  & 0.602         \\
        \mtlfiltered & 0.971  & 0.561         \\
        \shorttn     & N/A    & 0.618         \\
        \tnfiltered  & N/A    & 0.591         \\ \bottomrule
    \end{tabular}
    \label{tab:cxr-models}
\end{table}
We present the test-time classification and triplet accuracy of our models in Table \ref{tab:cxr-models}. Both \shortresn and \shortmtl achieve above 95\% classification accuracy. Both \shorttn and \shortmtl achieve above above 65\% triplet accuracy. Both \shorttn model and \shortmtl achieve the highest triplet accuracy in the 512-dimension unfiltered setting with triplet accuracy at 69.1\% and 72.2\% respectively.
\added{Filtering out class-inconsistent triplets removes 20.69\% of the triplet annotations in this dataset.}

\subsection{Model Evaluation with Synthetic Agent}

\begin{table}[t]
    \centering
    \begin{minipage}[b]{0.48\textwidth}
        \captionof{table}{CXR H2H preference results with synthetic agent.}
        \small
\centering
\begin{tabular}{@{}lrr@{}}
    \toprule
    Dimensions              &    50 & 512  \\ \midrule 
    \multicolumn{3}{c}{\textbf{NI H2H with \shortresn}} \\ \midrule
    \shortmtl               & 0.536 & 0.675 \\ 
    \shortmtl filtered      & 0.472 & 0.599 \\ \midrule
    \multicolumn{3}{c}{\textbf{NO H2H with \shortresn}} \\ \midrule
    \shortmtl               & 0.535 & 0.635 \\
    \shortmtl filtered      & 0.487 & 0.494 \\ \bottomrule
\end{tabular}
        \label{tb:cxr_h2h}
    \end{minipage}
    \hfill
    \begin{minipage}[b]{0.48\textwidth}
        \captionof{table}{CXR decision support accuracy with synthetic agent.}
        
    \small
    \centering
    \begin{tabular}{@{}lrr@{}}
    \toprule
    Dimensions              & 50    & 512  \\ \midrule
    \multicolumn{3}{c}{\textbf{Neutral Decision Support}}       \\ \midrule
    \shortresn              & 0.711 & 0.726  \\
    \shortmtl               & 0.742 & 0.779 \\ 
    \mtlfiltered     & 0.732 & 0.804 \\ 
    \midrule
    \multicolumn{3}{c}{\textbf{Persuasive Decision Support}}       \\ \midrule
    \shortresn              & 0.881 & 0.882  \\
    \shortmtl               & 0.949 & 0.966  \\
    \mtlfiltered    & 0.948 & 0.946  \\ 
    \bottomrule
    \end{tabular}
        \label{tb:cxr_decision}
    \end{minipage}
\end{table}

Similar to the BM setting, we select the \shorttn with the best test triplet accuracy as our synthetic agent, and then evaluate the examples produced by all representations. As table~\ref{tb:cxr_h2h} shows, preference for \shortmtl over \shortresn in H2H is less significant compared to BM, likely due to the challenging nature of the CXR dataset. We still observe the patten that filtering improves H2H performance.

Table \ref{tb:cxr_decision} shows decision support accuracy for different settings. All models benefit from a large dimension at 512. We observe consistent patterns such as filtering leading to better decision support.

\subsection{\added{Effect of triplet amount and type}}

\begin{table}[t]
    \begin{minipage}[b]{\textwidth}
        \centering
        \begin{subfigure}[b]{0.30\textwidth}
        \includegraphics[width=\textwidth]{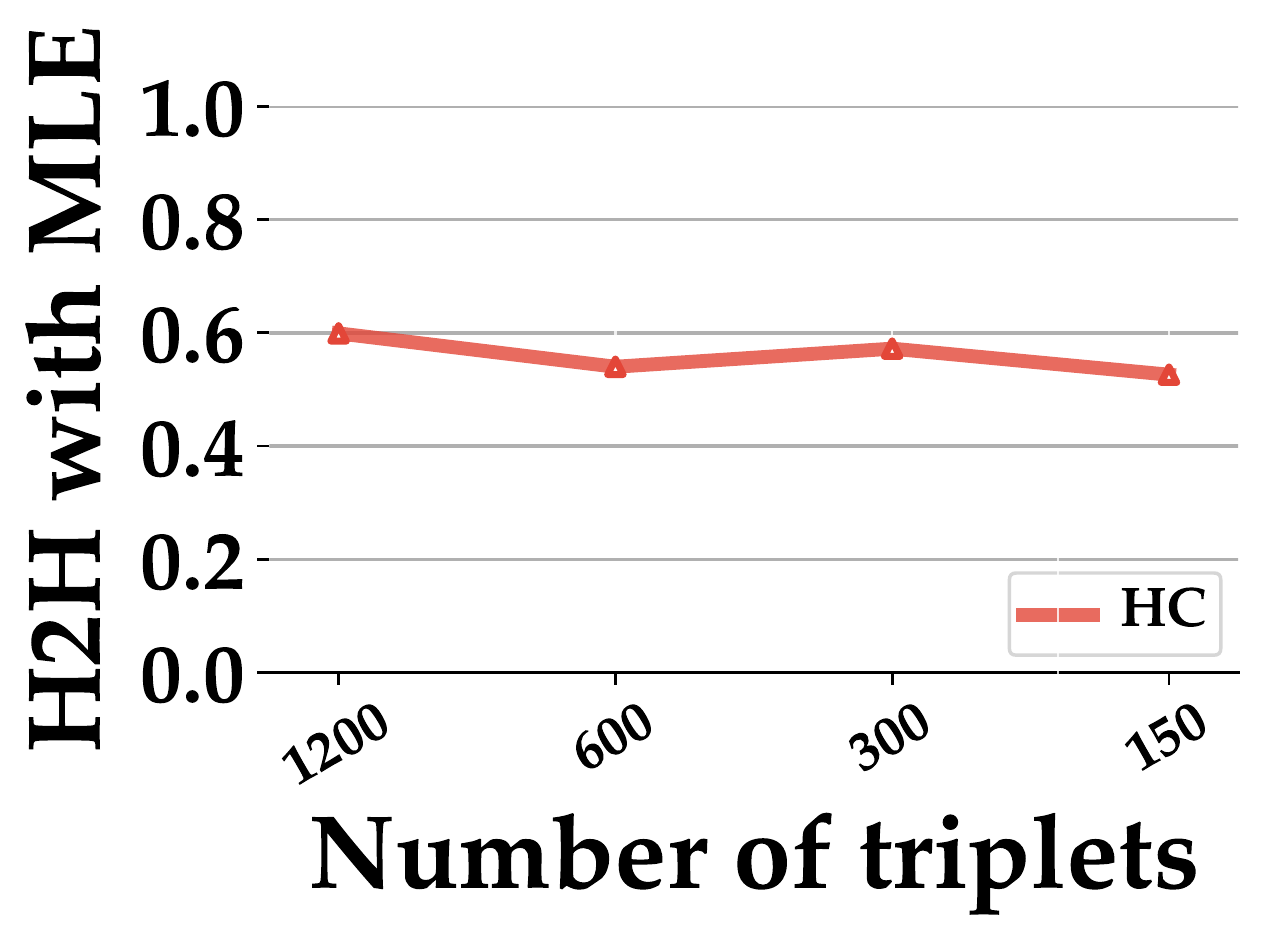}
        \end{subfigure}
        \begin{subfigure}[b]{0.31\textwidth}
        \includegraphics[width=\textwidth]{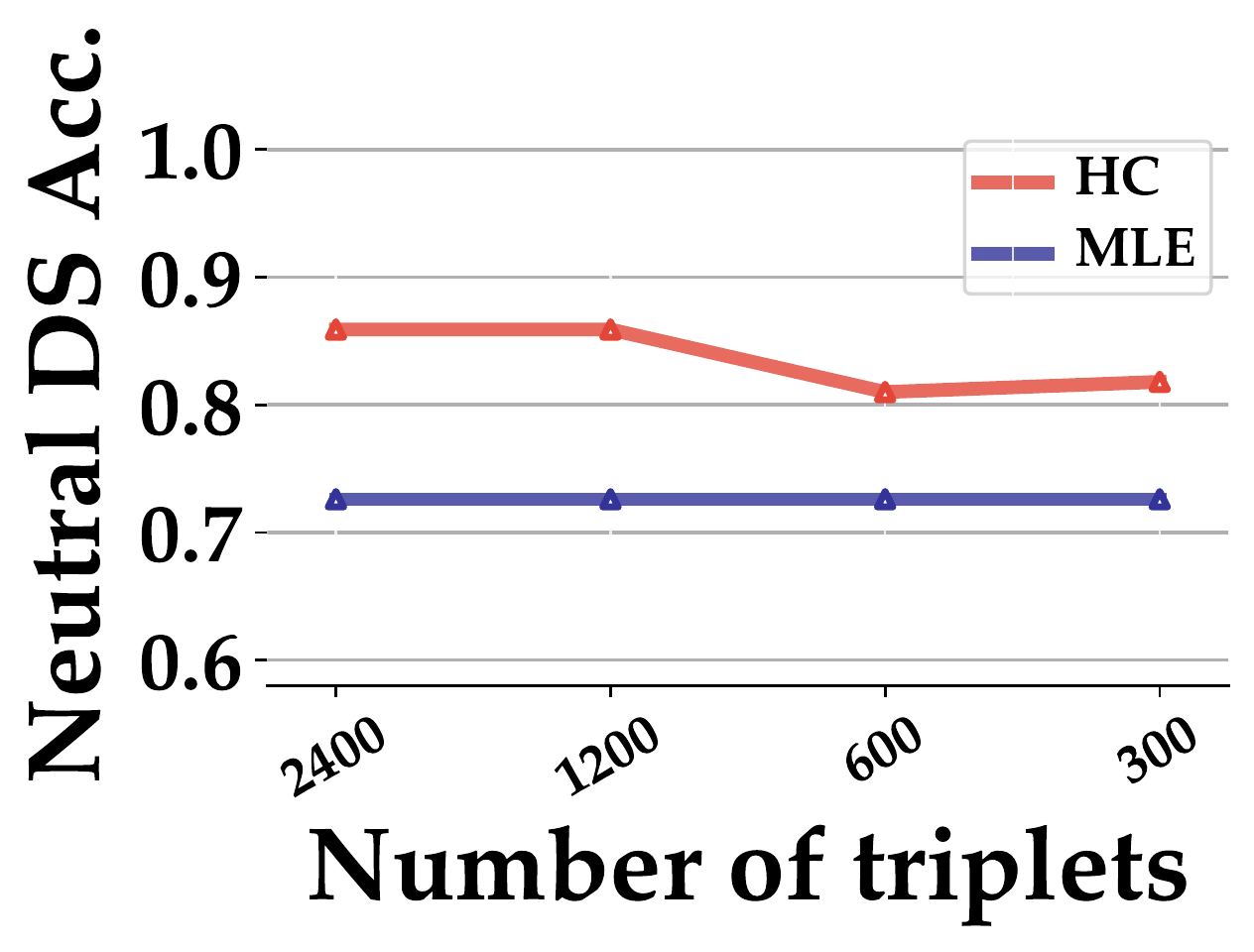}
        \end{subfigure}
        \begin{subfigure}[b]{0.37\textwidth}
        \includegraphics[width=\textwidth]{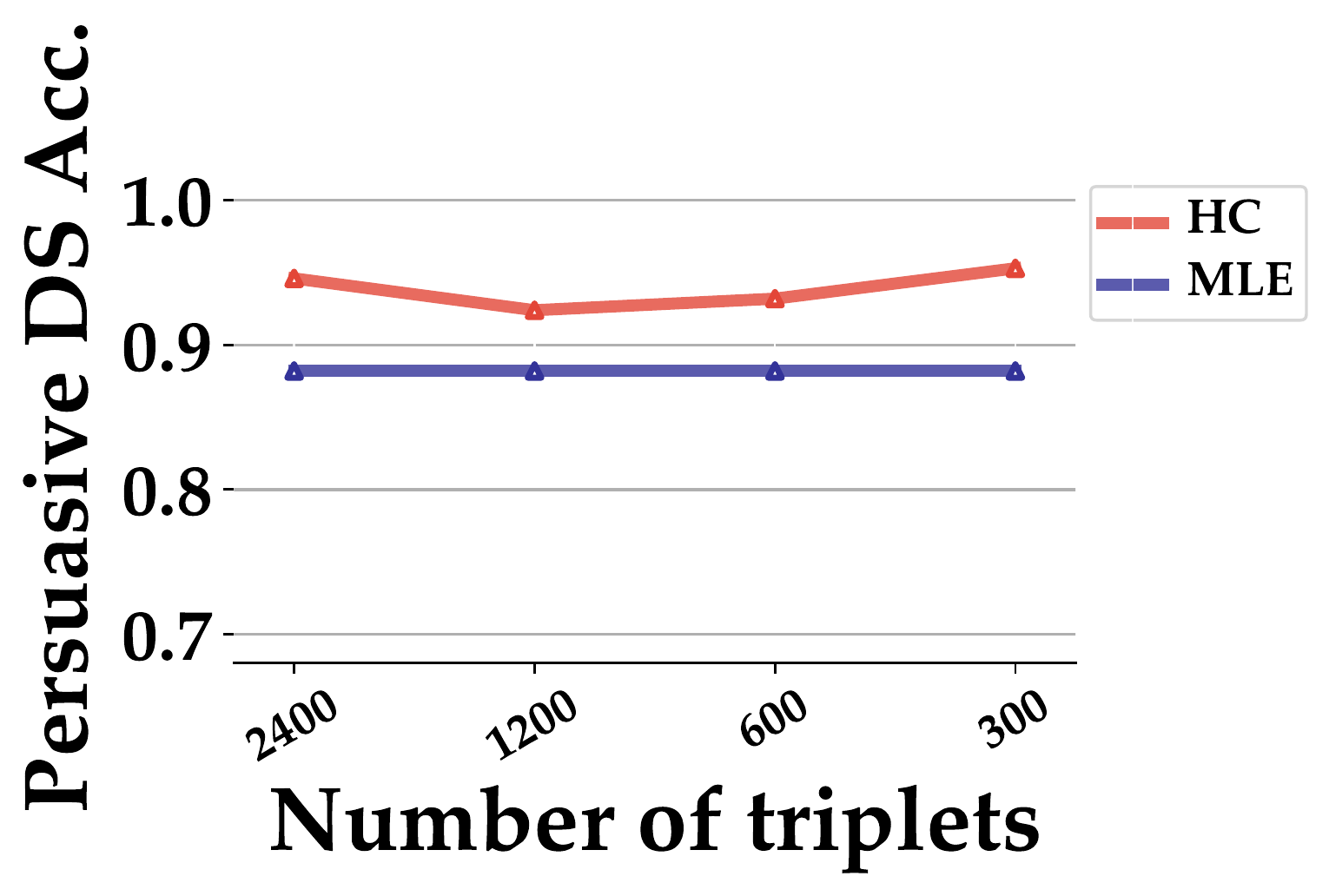}
        \end{subfigure}
        \captionof{figure}{\added{\shortmtl performance declines as the number of triplets decreases, but shows strong \nifo accuracy even with very few triplets.}}
        \label{fig:cxr-vary-num}
    \end{minipage}
\end{table}

\added{
We evaluate the effect of the number of triplets on our models in \figref{fig:cxr-vary-num}. Similar to the BM experiments, H2H preference towards \mtl and \nino performance decrease as the number of triplets decreases. \Mtl achieve strong \nifo performance even with very few triplets.
}

\subsection{Interface}
\begin{figure}
    \centering
    \begin{subfigure}[t]{0.48\textwidth}
        \includegraphics[width=\textwidth]{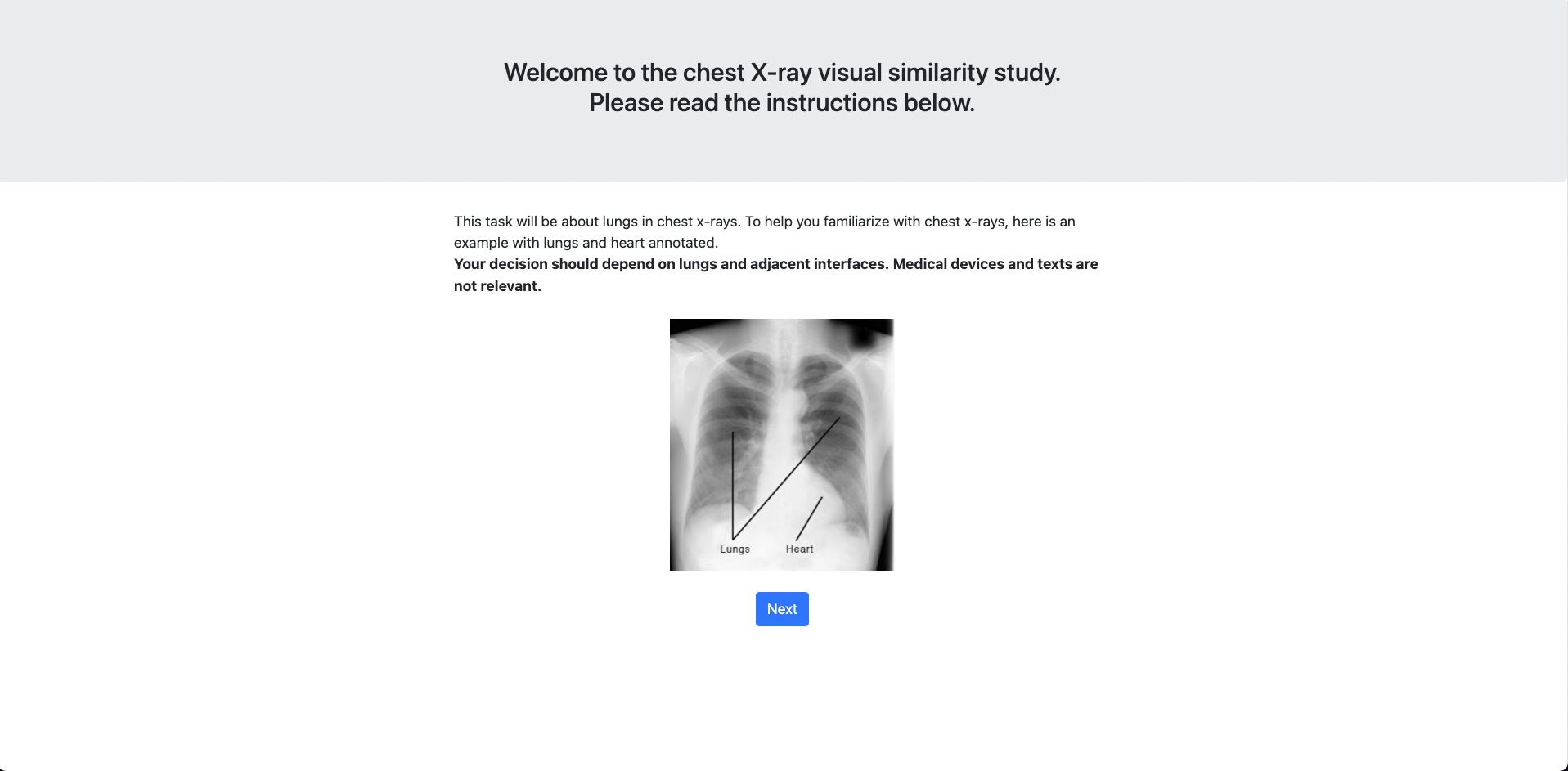}
        \caption{Basic instructions about chest X-rays.}
    \end{subfigure}
    \begin{subfigure}[t]{0.48\textwidth}
        \includegraphics[width=\textwidth]{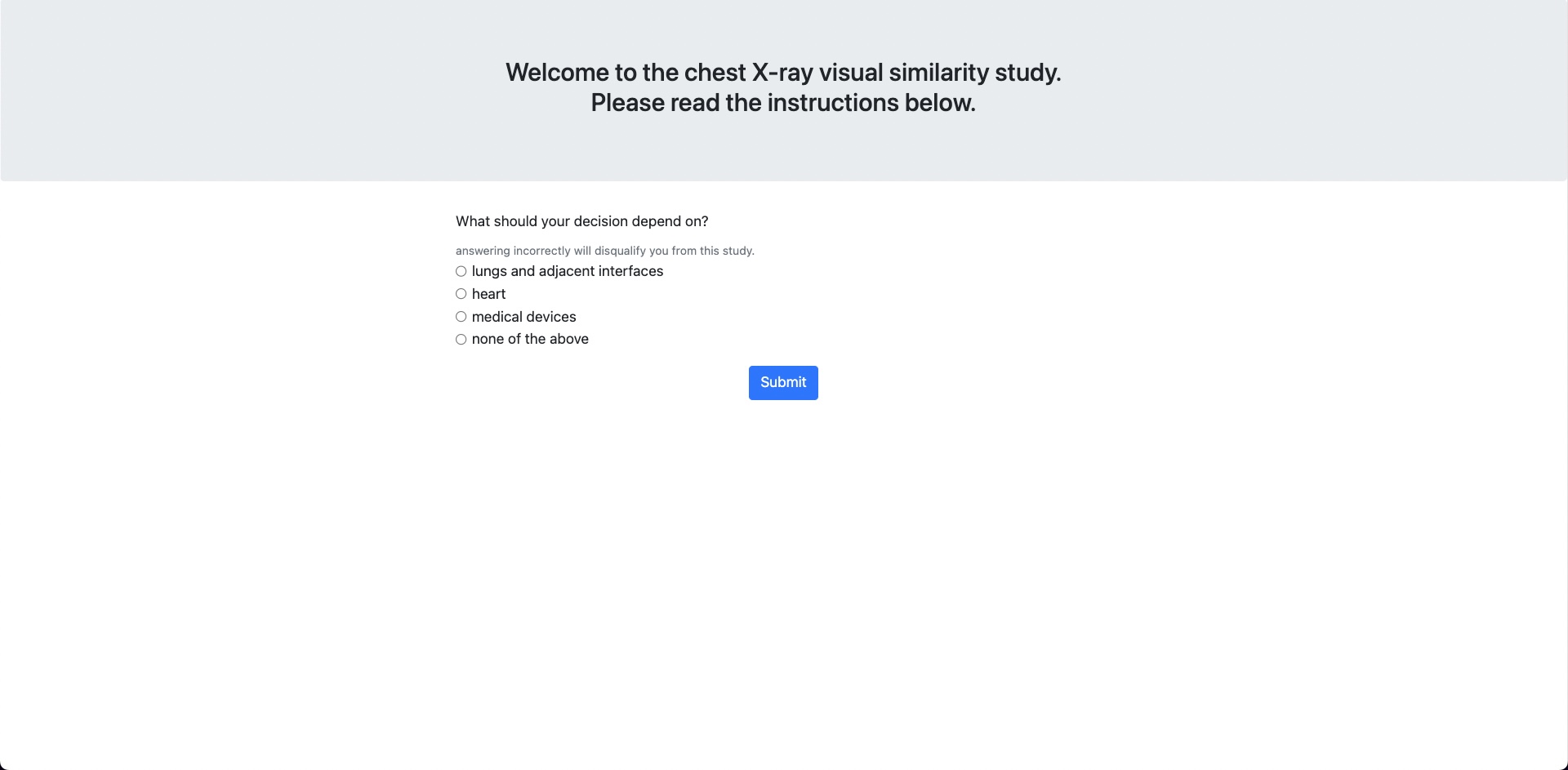}
        \caption{Multiple-choice attention check for CXR tasks. The correct answer is ``lungs and adjacent interfaces''.}
    \end{subfigure}
    \caption{Pre-task instructions and attentions check for CXR tasks}
    \label{fig:interface_cxr_instructions}
\end{figure}

\begin{figure}
    \centering
    \begin{subfigure}[t]{0.48\textwidth}
        \includegraphics[width=\textwidth]{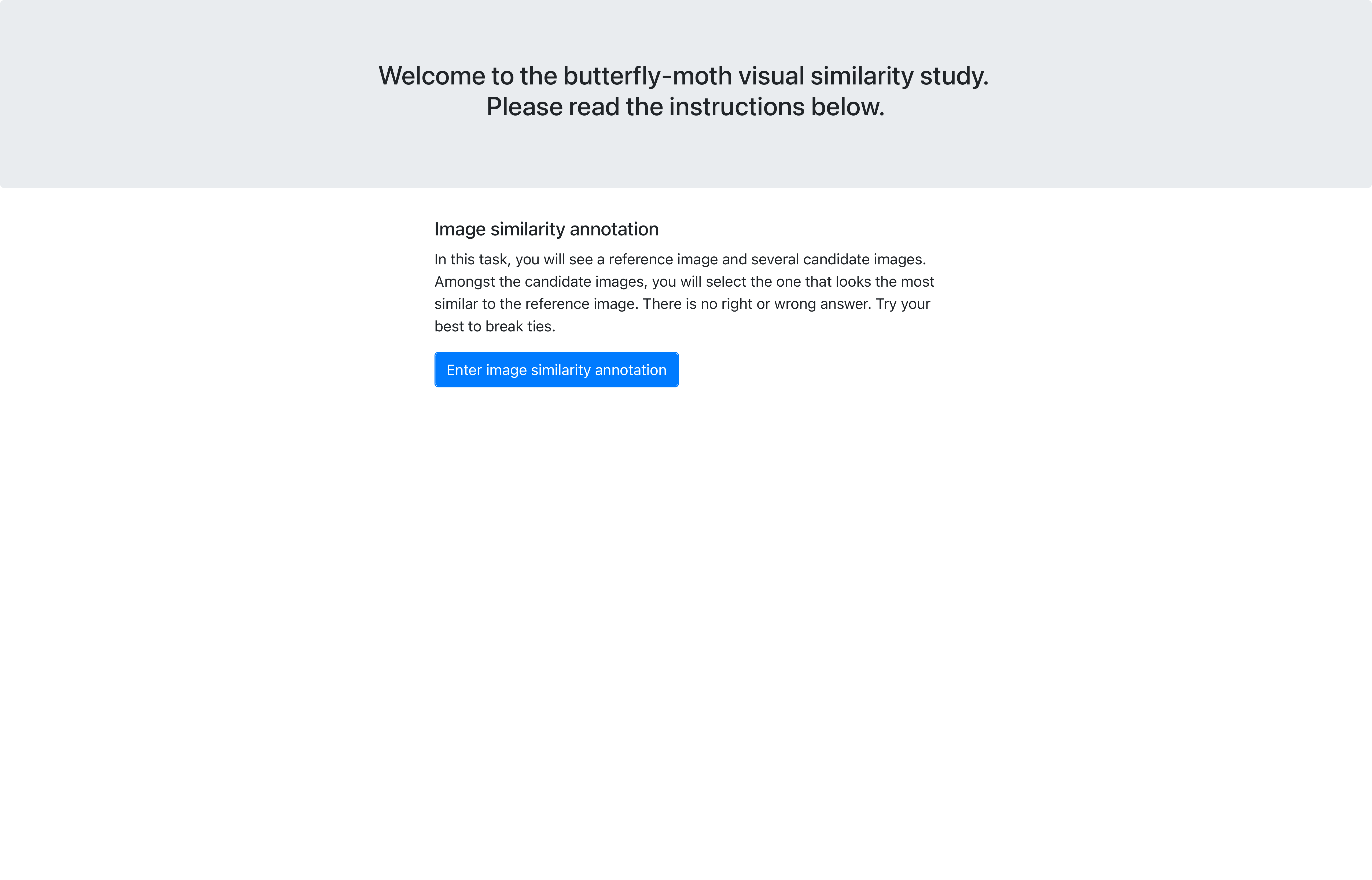}
        \caption{The annotation and head-to-head comparision task instructions.}
    \end{subfigure}
    \begin{subfigure}[t]{0.48\textwidth}
        \includegraphics[width=\textwidth]{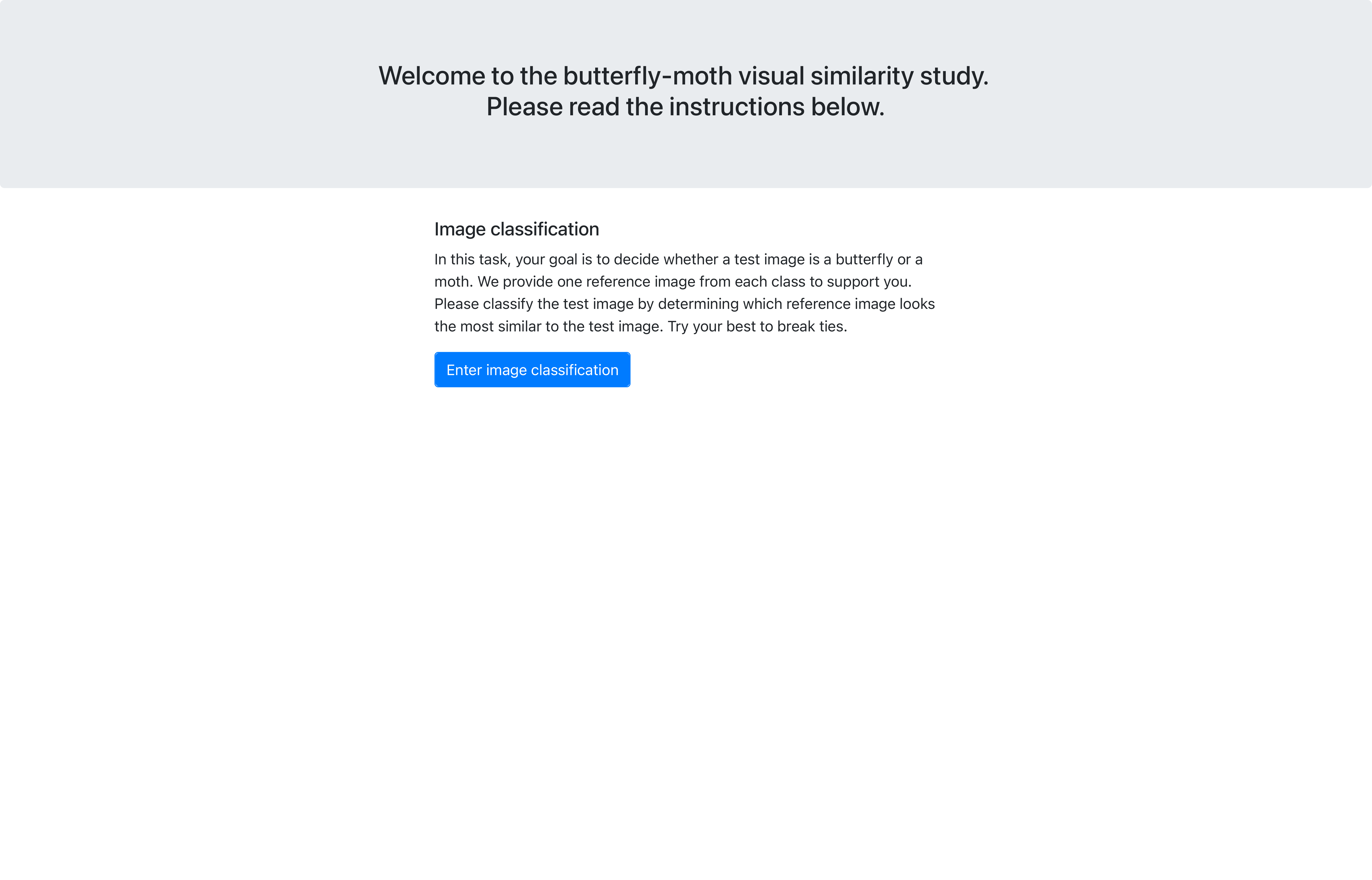}
        \caption{The decision support task instructions.}
    \end{subfigure}
    \caption{The task-specific instruction page on our interface.}
    \label{fig:interface_prolific}
\end{figure}

\begin{figure}
    \centering
    \begin{subfigure}[t]{0.48\textwidth}
        \includegraphics[width=\textwidth]{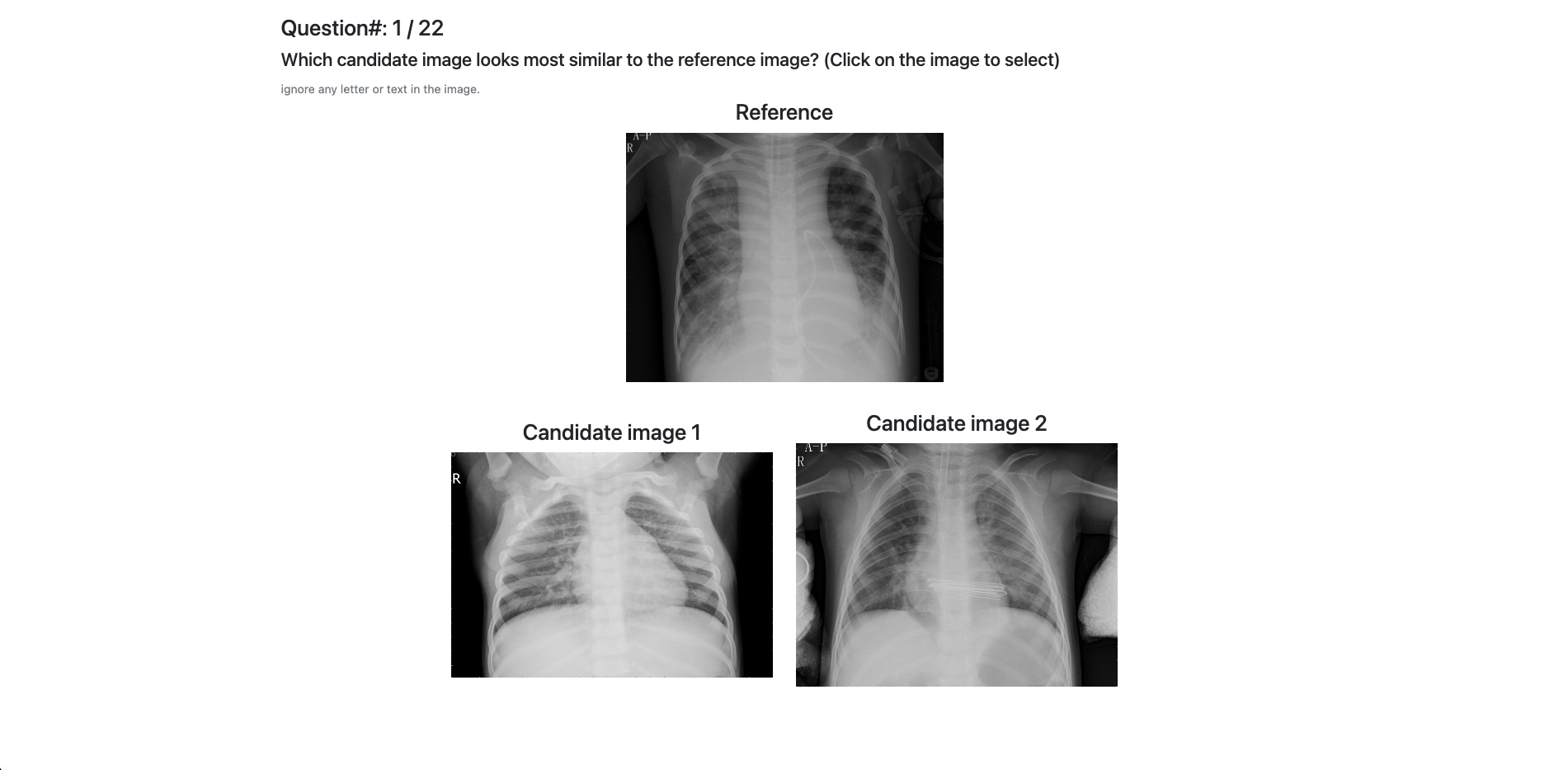}
        \caption{The annotation and head-to-head comparision task questions.}
    \end{subfigure}
    \begin{subfigure}[t]{0.48\textwidth}
        \includegraphics[width=\textwidth]{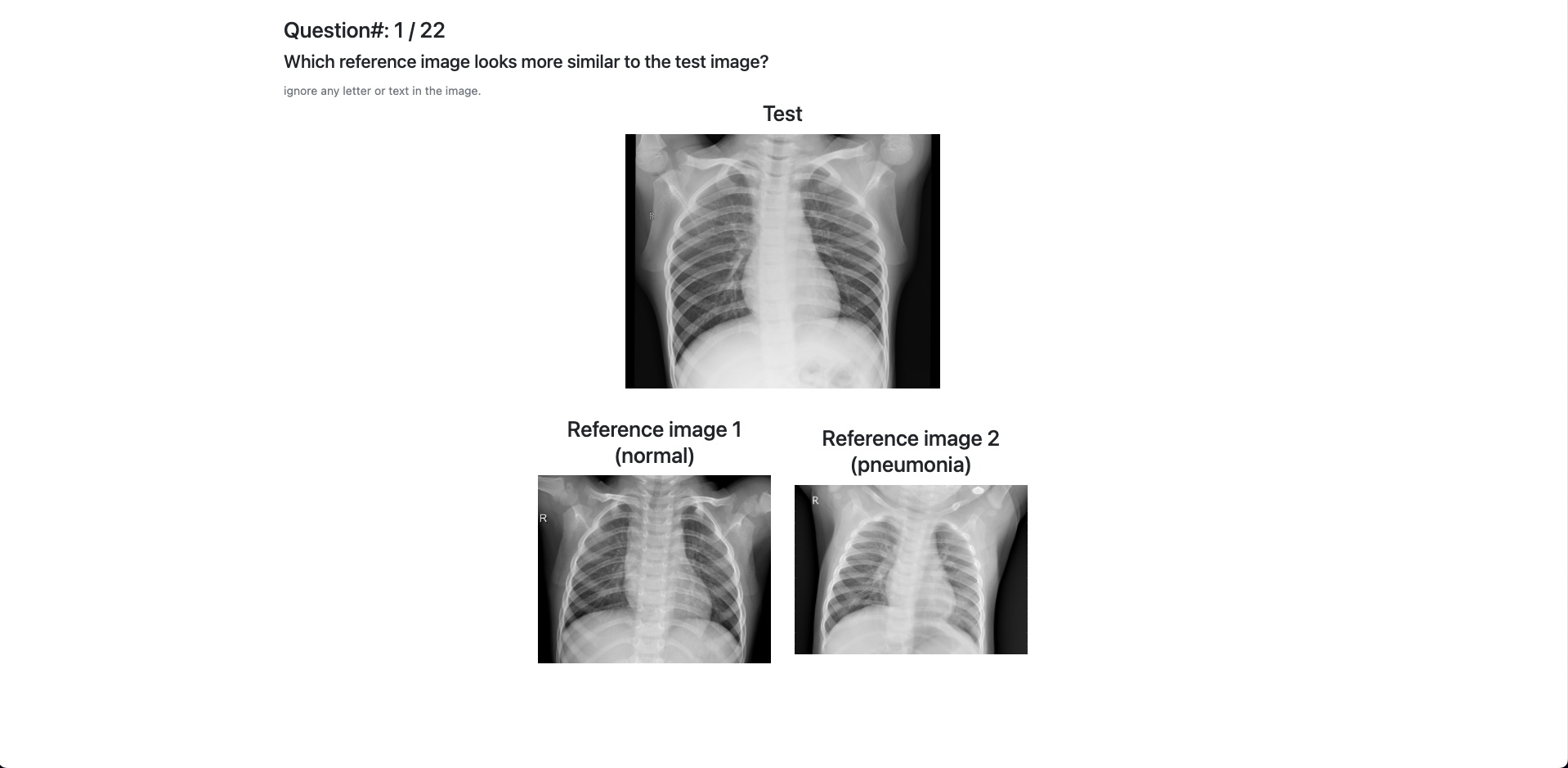}
        \caption{The decision support task questions.}
    \end{subfigure}
    \caption{The task-specific questions for CXR.}
    \label{fig:interface_cxr_questions}
\end{figure}

Our CXR interface is mostly the same as our BM interface, except that we add basic chest X-ray instructions as participants may not be familiar with medical images.
After the consent page at the beginning, participants will see basic chest X-ray instructions showing where the lungs and hearts.
Then, they enter an multiple-choice attention check, as shown in Fig \ref{fig:interface_cxr_instructions}. The correct answer in  ``lungs and adjacent interfaces''. Failing the attention check will disqualify the participant.
After correctly answering the pre-task attention check, participants will see the same task specific instructions as in the BM studies, as shown in Fig \ref{fig:interface_prolific}.
Screenshots of questions are shown in Fig \ref{fig:interface_cxr_questions}.
We also include two in-task attention check questions simlar to the BM study.
Our study is reviewed by the Institutional Review Board (IRB) at our institution with study number that we will release upon acceptance to preserve anonymity.

\subsection{Crowdsourcing}
We recruit our participants on Prolific (www.prolific.co) [September 2022].
We conduct three total studies: an annotation study, a decision support study, and a head-to-head comparison study.
We use the default standard sampling on Prolific for participant recruitment.
Eligible participants are limited to those reside in United States.
Participants are not allowed to attempt the same study more than once.

\para{Triplet annotation study}
We recruit 123 participants in total. 20 partipants fail the pre-task attention check question and 3 participants fail the in-task attention check questions; their responses are excluded in the results. We spend in total \$80.00 with an average pay at \$10.70 per hour. The median time taken to complete the study is 3'22''.

\para{Decision support study}
We recruit 296 participants in total. 34 partipants fail the pre-task attention check question and 10 participants fail the in-task attention check questions; their responses are excluded in the results. We spend in total \$221.67 with an average pay at \$11.00 per hour. The median time taken to complete the study is 3'40''.

\para{Head-to-head comparison study}
We recruit 57 participants in total. 6 partipants fail the pre-task attention check question and 1 participants fail the in-task attention check questions; their responses are excluded in the results. We spend in total \$40.00 with an average pay at \$10.54 per hour. The median time taken to complete the study is 3'25''.

\clearpage
\label{sec:supp_interface}

\begin{figure}
    \centering
    \includegraphics[width=\textwidth]{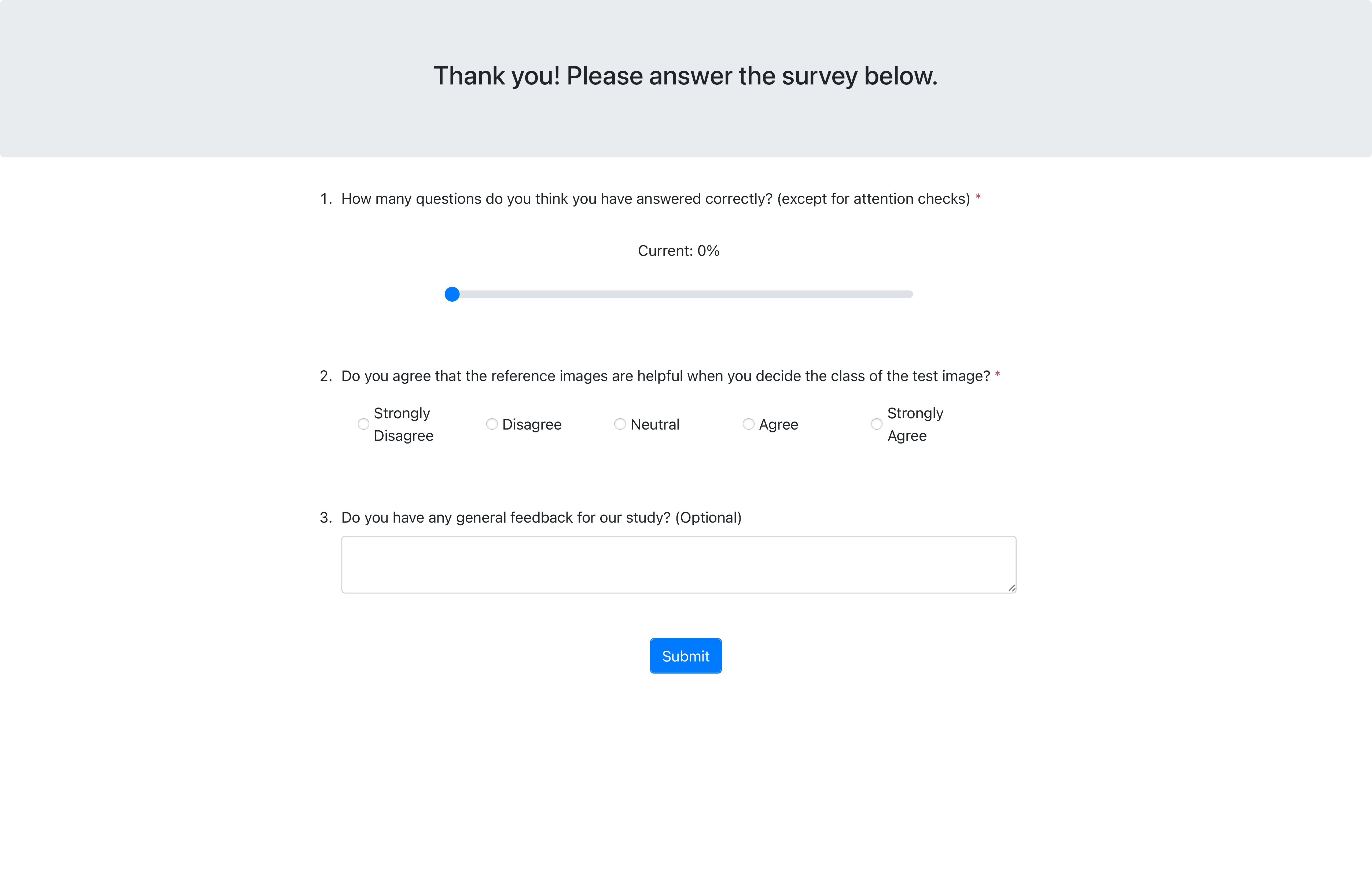}
    \caption{The survey page of the decision support task on our interface.}
    \label{fig:interface_survey_decision}
\end{figure}

\end{document}